\newif\ifconfver
\confverfalse      

\newif\ifcutshort      
\cutshorttrue

\newif\ifcutshortlvltwo  
\cutshortlvltwofalse

\ifconfver
\documentclass[10pt,twocolumn,twoside]{IEEEtran}
\else
\documentclass[11pt, draftclsnofoot, onecolumn]{IEEEtran}
\fi

\usepackage{amsmath,amsfonts,bm,amssymb,epsfig,psfrag,ifthen,color}
\usepackage{booktabs}
\usepackage{multirow}
\usepackage[table]{xcolor}  
\usepackage{array}
\usepackage{calc}
\usepackage{caption}
\usepackage{graphicx}
\usepackage{psfrag}
\usepackage[tight,footnotesize]{subfigure} 
\usepackage{stfloats}
\usepackage{url}
\usepackage{longtable}
\usepackage{amsmath,amsfonts,amssymb,amsbsy,bm,paralist,theorem,ifthen,color}
\usepackage{algorithmic,algorithm,theorem}
\usepackage{mathtools}

\newcommand\Ec{\ensuremath{\mathcal{E}}}
\newcommand\Wc{\ensuremath{\mathcal{W}}}
\newcommand\Hc{\ensuremath{\mathcal{H}}}

\newcommand\Pc{\ensuremath{{\mathcal{P}}}}
\newcommand\Ac{\ensuremath{{\mathcal{A}}}}

\newcommand\Qc{\ensuremath{{\mathcal{Q}}}}

\newcommand\xb{\ensuremath{{\bf x}}}

\newcommand\yb{\ensuremath{{\bm y}}}

\newcommand\Hb{\ensuremath{{\bf H}}}
\newcommand\hb{\ensuremath{{\bf h}}}
\newcommand\Ab{\ensuremath{{\bf A}}}

\newcommand\Bb{\ensuremath{{\bm B}}}

\newcommand\Eb{\ensuremath{{\bf E}}}

\newcommand\Gb{\ensuremath{{\bf G}}}

\newcommand\Ib{\ensuremath{{\bm I}}}

\newcommand\Xb{\ensuremath{{\bf X}}}

\newcommand\Ub{\ensuremath{{\bf U}}}

\newcommand\Vb{\ensuremath{{\bf V}}}

\newcommand\Wb{\ensuremath{{\bf W}}}

\newcommand\zerob{\ensuremath{{\bm 0}}}

\newcommand\E{\ensuremath{{\mathbb{E}}}}

\newcommand\oneb{\ensuremath{{\bf 1}}}

\newcommand{\wt}{\widetilde}

\newcommand{\ol}{\overline}

\newtheorem{Lemma}{Lemma}

\newtheorem{Theorem}{Theorem}

\newtheorem{Corollary}{Corollary}

\newtheorem{Rmk}{Remark}
\newtheorem{assumption}{Assumption}

\ifconfver

\else

\fi

\newcommand{\tabincell}[2]{\begin{tabular}{@{}#1@{}}#2\end{tabular}}

\begin{document}

    \bibliographystyle{IEEEtran}

    \title{ {Federated Matrix Factorization: Algorithm Design and Application to Data Clustering}}

    \ifconfver \else {\linespread{1.1} \rm \fi

        \author{\vspace{0.8cm}Shuai~Wang~and
        	Tsung-Hui~Chang\\
            \thanks{
   Shuai Wang and Tsung-Hui Chang are with the Shenzhen Research Institute of Big Data and School of Science and Engineering, The Chinese University of Hong Kong, Shenzhen 518172, China (e-mail: shuaiwang@link.cuhk.edu.cn,~tsunghui.chang@ieee.org).}
            }

        \maketitle
        
\vspace{-1.5cm}
\begin{center}
\today
\end{center}\vspace{0.5cm}

        \begin{abstract}
	       Recent demands on data privacy have called for federated learning (FL) as a  new distributed learning paradigm in massive and heterogeneous networks. 
	       Although many FL algorithms have been proposed, few of them have considered the matrix factorization (MF) model, which is known to have a vast number of signal processing and machine learning applications.
	       Different from the existing FL algorithms that are designed for smooth problems with single block of variables, 
	       in federated MF (FedMF), one has to deal with challenging non-convex and non-smooth problems (due to constraints or regularization) with two blocks of variables. 
	       In this paper, we address the challenge by proposing two new FedMF algorithms, namely, FedMAvg and FedMGS, based on the model averaging and gradient sharing principles, respectively. 
	       Both FedMAvg and FedMGS adopt multiple steps of local updates per communication round to speed up convergence, and allow only a randomly sampled subset of clients to communicate with the server for reducing the communication cost. 
	       Convergence analyses for the two algorithms are respectively presented, which delineate the impacts of data distribution, local update number, and partial client communication on the algorithm performance. By focusing on a data clustering task, extensive experiment results 
	       are presented to examine the practical performance of both algorithms, as well as demonstrating their efficacy over the existing distributed clustering algorithms.
            \\\\
        \end{abstract}


        \ifconfver \else \IEEEpeerreviewmaketitle} \fi

    \vspace{-0.8cm}

\section{Introduction}

\IEEEPARstart{M}{atrix} factorization (MF) is one of the most fundamental models which has vast applications in signal processing and machine learning, including data clustering, dimension reduction, item recommendation, hyperspectral unmixing, and biological analysis,  to name a few  \cite{Pomplili_2014,MF_RS_2009,NMF_review_2008}. 
Mathematically, a general MF problem is formulated as follows
\begin{subequations}\label{eqn: general prob}
	\begin{align}
	\min_{\substack{\Wb, \Hb}}~&  
	\Phi(\Xb, \Wb\Hb) +R_W(\Wb)+R_H(\Hb)\\
	{\rm s.t.}~&\Wb \in \Wc, \Hb \in \Hc,
	\end{align}
\end{subequations}
where $\Xb\in \mathbb{R}^{M \times N}$ is the observation data, $\Wb\in \mathbb{R}^{M \times K}$ and $\Hb=[\hb_1,\ldots,\hb_N]\in \mathbb{R}^{K \times N}$ are two matrix factors,  and $\Wc$ and $\Hc$ are some constraints, and $R_W(\cdot)$ and $R_H(\cdot)$ are 
regularization functions
for $\Wb$ and $\Hb$, respectively. 
The cost function $\Phi(\Xb, \Wb\Hb)$ measures the quality of the approximation $\Xb \approx \Wb\Hb$; for example,  $\Phi(\Xb, \Wb\Hb) = \frac{1}{N}\|\Xb - \Wb\Hb\|_{F}^2$  is a popular cost function used in many applications. 
In view of the increasing volume of real-life data, distributed MF methods that can process large-scale datasets 
have gained significant interests in the last decade \cite{SDNMF__BWU_2018}. However, recent emphasis on user privacy has called for new distributed schemes that can realize these MF based applications without revealing the users' private data. Specific examples include processing distributed patient medical records stored in multiple hospitals \cite{FML_CA_2019} and daily personal data of mobile users \cite{FL_Ondevice_2016}.

As an emerging distributed learning paradigm, federated learning (FL) has been introduced by Google  to enable collaborative model learning over distributed data owned by massive clients (e.g., mobile devices or institutions). The FL runs under the orchestration of a central server without the need of knowing the clients' raw private data. Compared with traditional distributed learning schemes \cite{ChangSPM2020}, FL faces new challenges.
This includes dealing with massively distributed clients in heterogeneous networks which have unbalanced and non-i.i.d. data distribution, and limited communication resource for message exchanges between the server and clients \cite{FL_Ondevice_2016}.

To train a model under the challenging FL setting, serveral FL algorithms have been proposed \cite{CE_DDNN_2017,FedProx_2018,EDDL_DModelAvg_2018,Parallel_RSGD_2019,FedAvg_noniid_2019}, mostly based on the classical stochastic gradient descent (SGD) method. In particular, \cite{CE_DDNN_2017} proposed a \emph{model averaging} algorithm, called {FedAvg}, where the server coordinates the training by iteratively averaging the local models learned by the clients via SGD. A salient feature of {FedAvg} over the classical \emph{gradient sharing} approach is \emph{local SGD}, where the clients are allowed to perform multiple epochs of SGD locally before sending the local model to the server for averaging. Local SGD has been proven an effective strategy to reduce the required number of communication rounds for producing a good model \cite{CE_DDNN_2017,Parallel_RSGD_2019,FedAvg_noniid_2019}. 
The second feature of FedAvg is \emph{partial client participation} (PCP), where only a small number of clients are sampled and communicate with the server in each communication round. Partial participation can greatly alleviate the network congestion problem especially when the number of clients is large and communication bandwidth is limited.
It also models that a client might become offline randomly due to poor link quality.
We notice that, while many successful efforts have been made for supervised FL tasks, few works have been done for the MF model \eqref{eqn: general prob} and its applications.

\subsection{Related Works}
For example, the recent works  \cite{DRMF_CF_2019} and \cite{SFMF_2019} studied the federated MF (FedMF) problem for recommendation systems. They considered the gradient sharing strategy, where, in each communication round, the server aggregates the gradient information of the cost function from the clients and applies one step of gradient descent for model training. However, as mentioned in \cite{CE_DDNN_2017}, the gradient sharing based methods would require a large number of communication rounds to produce a good model. It is also worth noting that the existing model averaging based FL algorithms \cite{CE_DDNN_2017,FedProx_2018,EDDL_DModelAvg_2018, Parallel_RSGD_2019,FedAvg_noniid_2019} are not directly applicable to problem \eqref{eqn: general prob} because these existing studies have assumed smooth and unconstrained problems while \eqref{eqn: general prob} is constrained, and the existing algorithms are designed to deal with problems with single-block variable while  \eqref{eqn: general prob}  involves two blocks of variables. 

On the other hand, although many distributed MF methods have been proposed, they did not consider the FL scenario and address the associated issues. Specifically,  
a large body of the existing distributed MF methods are parallel implementations of the centralized sequential SGD or alternating least square (ALS) algorithms, either on MapReduce \cite{LSMF_DSGD_2011, DMF_MRBJ_2013, FDSGD_MF_2014} or Parameter server \cite{Factorbird_2014}. 
Analogously, parallel implementations of the multiplicative rule \cite{DNMF_WSDD_2010} and block coordinate descent \cite{DNMF_HALS_2017, SDNMF__BWU_2018} on MapReduce are developed for non-negative MF (NMF) models.
Again, these works usually assume that there is a shared memory that all nodes can access, and careful model/data partition is required for efficient parallelization.
Another category of works considered decentralized MF methods such as 
\cite{wai2015consensus,pmlr-v70-hong17a,PPPIR_DMF_2019,DLMF_EC_2019} with the absence of the central server.
Consensus methods are often used to achieve distributed optimization. However, the critical issues of FL such as  unbalanced/non-i.i.d. data are not considered therein.

In summary, firstly,  the existing FedMF and distributed MF algorithms have not been fully customized to overcome the challenges of FL in massive and heterogeneous networks. Secondly, the current studies have focused on specific applications (such as  item recommendation), there still lacks a systematic algorithm design for the general MF model \eqref{eqn: general prob}  so that the designed algorithms can subsequently be employed in various MF applications. Thirdly, the existing FedMF works do not present theoretical convergence analysis and thus are not able to analytically explain how various factors, such as non-i.i.d. data distribution and partial communication, can affect the algorithm performance. 
\subsection{Contributions}

In this paper, we aim to investigate communication-efficient FedMF algorithms. In particular, we propose two novel FedMF algorithms, based on the principles of model average (MA) and gradient sharing (GS), respectively. We assume that $\Wb$ is a shared variable of all clients whereas $\Hb$ can be (column-wise) partitioned into multiple local variables exclusively owned by respective clients; see Section \ref{sec: problem formulation} for the detailed FedMF model.

\vspace{-0.0cm}
\begin{enumerate}
	
	\item 
	
	We firstly propose a new FedMF algorithm, called FedMAvg,  that judiciously combines the alternating minimization method and the MA technique, where the former is a popular method for handling the MF model in \eqref{eqn: general prob} \cite{GLRM_2016}. 
	Specifically, in each communication round of FedMAvg, the clients 
	perform one round of local alternating minimization through multiple steps of local projected gradient descent (local PGD) with respect to $\Hb$ and $\Wb$, followed by averaging the locally learned model $\Wb$ at the server. Besides, partial client communication (PCC) is adopted to reduce the overall communication cost, where only a sampled subset of clients upload their models to the server in each round. We present a convergence analysis which explicitly characterizes how the number of local updates, PCC and non-i.i.d. data distribution can affect the algorithm convergence. The analysis suggests that a \emph{diminishing} number of local updates should be used for FedMAvg so that it can be less sensitive to non-i.i.d. data.
	Unlike the existing analyses \cite{CE_DDNN_2017,FedProx_2018,EDDL_DModelAvg_2018,Parallel_RSGD_2019,FedAvg_noniid_2019} which are for smooth unconstrained problems, our analysis overcomes the challenge due to the constraints and two-block variables in problem \eqref{eqn: general prob}.
	
	%
	
	
	\item Secondly, we propose a new GS based FedMF algorithm, called FedMGS, which improves upon the algorithms in \cite{SFMF_2019,DRMF_CF_2019} in terms of convergence speed and communication efficiency.
	In FedMGS, the clients are responsible for computing $\nabla_W\Phi$ and uploading it to the server which is in charge of updating $\Wb$.
	%
	We focus on MF problems with 
	the squared Frobenius norm cost, i.e., 
	$\Phi(\Xb, \Wb\Hb)=\frac{1}{N} \|\Xb - \Wb\Hb\|_{F}^2$.
	By exploring the fact that $\nabla_W\Phi$ has a linear separable structure, 
	one can allow both the clients and the server to perform multiple steps of PGD in each communication round, which thus can improve the convergence speed in a way similar to MA based methods. Analogously, in FedMGS, we allow only a sampled subset of clients to be active and communicate with the server, which reduces the communication cost.
	Convergence analysis reveals that FedMGS is inherently resilient to non-i.i.d. data distribution. 
	
\end{enumerate}

\vspace{-0.0cm}
To examine the performance of the proposed FedMF algorithms, we apply them to the \emph{federated data clustering} task and test them on both 
synthetic dataset and real datasets including the TDT2 document data \cite{LCCF_2011}, the TCGA  cancer gene  data \cite{TCGA_CGCD}, and the MNIST hand-writing digits data \cite{website_MNIST}. 
Extensive experiment results are presented, which not only provide useful insights on how various algorithm parameters and data distribution affect the algorithm performance, but also show the superiority of the proposed algorithms over the existing distributed clustering methods.

{\bf Synopsis:} Section \ref{sec: problem formulation}  presents the FedMF problem model and its application to data clustering.
Section \ref{sec: fedam} and Section \ref{sec: FedMGS} respectively present the proposed FedMAvg and FedMGS algorithms and their convergence analyses. Experiment results are presented in Section \ref{sec: simulation}, and lastly the conclusion is drawn in Section \ref{sec: conclusion}.

{\bf Notation:} 
$\mathbb{R}^{m \times n}$ denotes the set of $m$ by $n$ real-valued matrices. The $(i,j)$th entry of matrix $\Ab$ is denoted by 
$[\Ab]_{ij}$; 
superscript $\top$ stands for matrix transpose.
$\oneb$ denotes the all-one vector, $\zerob$ is the all-zero vector, and
$ \Ib_{m}$ is the $m$ by $m$ identity matrix, and
$\|\cdot\|_F$ is the matrix Frobenius norm.
$\langle \Ab,\Bb\rangle$ denotes the inner product between matrices $\Ab$ and $\Bb$.
$\lambda_{\max}(\Ab)$ stands for the maximum eigenvalue of $\Ab$. 

\section{Federated MF and Application to Clustering}\label{sec: problem formulation}
\subsection{FedMF Model}

By considering the FL setting, we assume that the data matrix can be partitioned as $\Xb=[\Xb_1,\Xb_2,\ldots,\Xb_P]$ and respectively owned by $P$ distributed clients. Specifically, each client $p$ owns non-overlapping data $\Xb_p\in \mathbb{R}^{M \times N_p}$, where $N_p$ is the number of samples of client $p$ and $\sum_{p = 1}^{P} N_p = N$. Besides, we assume that there is a server who coordinates the $P$ clients to accomplish the MF task with all the distributed data $\Xb_1,\Xb_2,\ldots,\Xb_P$ being considered. 
Note that, under the FL scenario, the number of clients $P$ could be large, the data size $N_p$, $p=1,\ldots,P$, could be unbalanced, and the data samples $\Xb_1,\Xb_2,\ldots,\Xb_P$ could be non-i.i.d. \cite{FL_Beyond_2015,FL_Ondevice_2016}.

Let $\Hb=[\Hb_1,\ldots,\Hb_P]$ be partitioned in the same fashion as $\Xb$, and let $\omega_p=N_p/N$, $p\in \Pc \triangleq \{ 1, \ldots, P\}$. Moreover, assume that $\Phi(\Xb,\Wb\Hb)=\sum_{p=1}^P \Phi_p(\Xb_p,\Wb\Hb_p)$, $R_H(\Hb)=\sum_{p=1}^P R_H(\Hb_p)$ and $\Hc=\Hc_1\times \Hc_2 \cdots  \times \Hc_P$ which are separable with respect to $\Hb_1,\ldots,\Hb_P$. Then, one can write the MF problem \eqref{eqn: general prob} as
{\begin{subequations}\label{eqn: distributed prob}
		\begin{align}
		\min_{\substack{\Wb,~ \Hb_p, \\
				p = 1, \ldots, P}} ~&F(\Wb, \Hb)\triangleq \sum_{p = 1}^{P}\omega_p F_p(\Wb, \Hb_p) \\
		{\rm s.t.}~&\Wb \in \Wc, \Hb_p \in \Hc_p, \forall p \in \Pc,
		\end{align}
\end{subequations}}

\noindent where 
{\begin{align}\label{eqn: obj of client p}
	\hspace{-0.3cm} F_p(\Wb, \Hb_p) = \frac{\Phi_p(\Xb_p,\Wb\Hb_p)}{\omega_p}  + \frac{R_H(\Hb_p)}{\omega_p}+R_W(\Wb)
	\end{align}}
\noindent is the local cost function of each client $p$. 
As seen, $\Wb$ is a shared variable whereas $\Hb_p$, $p \in \Pc$, are client local variables.

The FedMF algorithm should enable the server to coordinate the distributed clients to jointly solve the MF problem \eqref{eqn: distributed prob} without the need of the clients revealing their private raw data.
We should emphasize here that problem \eqref{eqn: distributed prob} is much more challenging to solve than the FL problems considered in the literature \cite{CE_DDNN_2017,FedProx_2018,EDDL_DModelAvg_2018, Parallel_RSGD_2019,FedAvg_noniid_2019} because problem \eqref{eqn: distributed prob} is non-convex and non-smooth (due to the constraints) and 
problem \eqref{eqn: distributed prob}  involves two blocks of variables $\Wb$ and $\Hb$. Thus, the existing FL algorithms that are designed for smooth problems with single-block variable cannot be applied to problem \eqref{eqn: distributed prob}. In addition, the GS based methods in  \cite{DRMF_CF_2019} and \cite{SFMF_2019} are not communication-efficient solutions to problem \eqref{eqn: distributed prob}, as mentioned in Section \ref{subsec: related work} and will be verified in Section \ref{sec: simulation}.

\subsection{Federated Clustering via FedMF}\label{sub: clustering}

As mentioned, the MF model \eqref{eqn: general prob} has many applications in signal processing and machine learning. In this paper, we are particularly interested in applying the MF model \eqref{eqn: general prob} for data clustering, in view of that clustering is one of the most fundamental data mining tasks.

Take clustering as the example.
When $\Phi(\Xb, \Wb\Hb)=\frac{1}{N} \|\Xb - \Wb\Hb\|_{F}^2$, $R_W(\Wb)= R_H(\Hb) =\zerob$ and $\Hc=\{\Hb~ |~ \oneb^\top\hb_j = 1, ~[\Hb]_{ij} \in \{0, 1\},~\forall i=1,\ldots,K, j=1,\ldots,N\}$,
the MF model \eqref{eqn: general prob} corresponds to the classical K-means formulation \cite{Pomplili_2014}. 
Specifically, in \eqref{eqn: general prob}, columns of $\Wb$ represent centroids of the $K$ clusters, while $\Hb \in \Hc$ is the cluster assignment matrix where
$[\Hb]_{ij}=1$ indicates that data sample $\xb_j$ is uniquely assigned to cluster $i$.
Thus, the K-means algorithm is equivalent to solving the above MF problem \eqref{eqn: general prob} via alternating minimization \cite{Kmeans_NMF_2015}.
Interestingly, recent studies have shown that structured MF models such as the orthogonal non-negative MF (NMF)  \cite{Pomplili_2014, Shuai_SNCP_2019} can outperform the K-means in many application scenarios.

However, there lacks algorithms that can perform clustering in FL network. In the literature, there are two main categories for distributed clustering. In the first category, the methods are simply parallel implementations of the centralized clustering algorithms, such as K-means \cite{PKmeans_KDD_2000,Kmeans||_2012,PKmeans_2014} and density based DBSCAN \cite{DBSCAN_1996}, but 
they usually assume a shared memory, which is opposite to the setting of FL.

Distributed clustering methods in the second category target at approximating the centralized clustering methods via constructing so-called coreset, which is a small-sized set of weighted samples whose cost approximates the cost of the original dataset. Thus, clustering over the coreset is approximately the same as clustering over the original dataset, which resolves the large-scale clustering issue. For example, in \cite{KMeans_DD_2016} distributed clients generate local coresets based on local data, and their union constitutes a global coreset, while in \cite{DKCoreset_2013}, a global coreset is directly constructed from locally clustering results. Impressively, these methods are communication efficient since the clients require to communicate with the server for one or two rounds only. Approximation ratios with respect to the referenced algorithms (such as K-means/K-median/K-centers) are also guaranteed \cite{DKCoreset_2013,KMeans_DD_2016,DKMeans_noise_2018, PA_DC_outlier_2018}. 
However, these coreset methods are in general no better than their referenced algorithms. 

In the next two sections, we present the proposed FedMF algorithms, and then examine their practical performance on the data clustering task in Section \ref{sec: simulation}.

\section{Federated MF by Model Averaging} \label{sec: fedam}

In this section, we develop a MA based FedMF algorithm, termed FedMAvg, and establish its theoretical property.

\vspace{-0.3cm}
\subsection{FedMAvg Algorithm}\label{sec: fedam development}

Straightforward application of the MA technique to the distributed MF problem \eqref{eqn: distributed prob} would lead to an iterative algorithm as follows. For round $s=1,2,\ldots$, each client $p$ obtains an approximate solution to the corresponding local subproblem of \eqref{eqn: distributed prob}, i.e.,
{
	\begin{subequations}\label{eqn: local_prob}
		\begin{align}
		(\Wb_p^s, \Hb_p^s )=  \arg\min_{\substack{\Wb, \Hb_p}} ~&F_p(\Wb, \Hb_p)\\
		{\rm s.t.}~&\Wb \in \Wc, \Hb_p \in \Hc_p.
		\end{align}
\end{subequations}}
Since $\Wb$ in \eqref{eqn: distributed prob} is the shared variable,  
the server collects and takes certain average of $\Wb_1^s,\ldots,\Wb_P^s$, denoted by ${\Wb}^s$, and broadcasts the average ${\Wb}^s$ to the clients for the next round of updates. 

There are many possible ways to handling problem \eqref{eqn: local_prob}.
One approach is simply employing one step of the alternating minimization; that is,  given ${ \Wb}^{s-1}$ in the previous round, each client $p$ performs
{
	\begin{subequations}\label{eqn: AM}
		\begin{align}
		\Hb_p^s&=\arg\min_{\substack{\Hb_p\in \Hc_p}} ~F_p({\Wb}^{s-1}, \Hb_p), \label{eqn: AM H}\\
		\Wb_p^s&=\arg\min_{\substack{\Wb \in \Wc}} ~F_p(\Wb, \Hb_p^s). \label{eqn: AM W}
		\end{align}
\end{subequations}} \vspace{-0.5cm}

In practice, it is sufficient to employ the simple PGD method for \eqref{eqn: AM H} and \eqref{eqn: AM W}, respectively. In particular, by following the same spirit as the multiple-step local SGD in {FedAvg} \cite{CE_DDNN_2017}, 
we propose to approximate \eqref{eqn: AM H} by $Q_1\geq 1$ consecutive steps of PGD with respect to $\Hb_p$, i.e., for $t=1, \ldots, Q_1$,
{\begin{align}
	\hspace{-0.1cm}	&\Hb_p^{s, t}\! =\! \Pc_{\Hc_p} \big\{\Hb_p^{s,t-1}\! -\! \frac{1}{c_p^{s}}{\nabla_{H_p}F_p(\Wb^{s-1}, \Hb_p^{s,t-1})}\big\}, \label{eqn: FedAM update of H1 0}
	\end{align}}where  $\Hb_p^{s, 0}=\Hb_p^{s-1}$; $c_p^t>0$ is the step size, and $\Pc_{\Hc}$ denotes the projection operation onto the sets $\Hc_p$. Denote $\Hb_p^{s}=\Hb_p^{s, Q_1}$ for all $p\in \Pc$.

Analogously, we approximate \eqref{eqn: AM W} by $Q_2\geq 1$ consecutive steps of GD (no projection) with respect to $\Wb$, i.e., for $t = Q_1 + 1 \ldots, Q,$
{\begin{align}
	\hspace{-0.1cm}&\Wb_p^{s, t} = \Wb_p^{s, t-1} - \frac{1}{d^s}{\nabla_{W}F_p(\Wb_p^{s, t-1}, \Hb_p^{s, Q_1})},\label{eqn: FedAM update of W2 0}
	\end{align}}where $Q=Q_1+Q_2$ and $d_s>0$ is a step size.
After a total number of $Q$ local model updates, each client $p$ sends its local model of $\Wb_p^s=\Wb_p^{s,Q}$ to the server. The server takes the weighted average and applies projection operation $\Pc_{\Wc}$ to it
{ \begin{align}
	\Wb^s = \Pc_{\Wc}\bigg(\sum_{p = 1}^{P}\omega_p\Wb_p^{s-1, Q}\bigg).
	\label{eqn: FedAM proj of W 0}
	\end{align}}\vspace{-0.3cm}

{\bf Diminishing $Q_2$:} As $\Wb$ is the shared variable, the local GD length $Q_2$ should not be too large since it may make the local variable deviate from the global one and slow down the algorithm convergence, especially in the presence of heterogeneous non-i.i.d. data. This insight was also found in the classical FedAvg algorithm \cite{CE_DDNN_2017,Parallel_RSGD_2019,FedAvg_noniid_2019}.
To overcome the issue, we propose to consider a diminishing $Q_2$; for example, we consider $Q_2^s=\lfloor \frac{\hat Q}{s}\rfloor+1$, where $\hat Q$ is a preset number. The intuition is that in the early iterations the clients should ``explore" more by performing more GD updates based on its local data, whereas when the algorithm is close to convergence, they should make small movements only to avoid model deviation. As will be demonstrated later by theoretical analysis and empirical experiments, the strategy of diminishing $Q_2$  can benefit the algorithm convergence significantly.

{\bf Partial client communication (PCC):}
After a total number of $Q^s=Q_1+Q_2^s$ local updates, each client $p$ sends $\Wb_p^{s,Q^s}$ to the server for model averaging.
Like FedAvg \cite{CE_DDNN_2017}, 
we let the server samples a small, fixed-size subset of clients (denoted by $\Ac^s$ with size $|\Ac^s| = m \ll P$) and ask them to upload their local models $\Wb_p^s$, $p\in \Ac^s$.
The server then simply takes the average of the uploaded messages by

\vspace{-0.3cm}
{\begin{align}
	\Wb^s = \Pc_{\Wc}\bigg(\frac{1}{m}\sum_{p \in \Ac^{s-1}}\Wb_p^{s-1, Q}\bigg),
	\label{eqn: FedAM proj of W PCC}
	\end{align}}\vspace{-0.3cm}

\noindent followed by projection onto $\Wc$.
Note that under PCC, the clients that are not selected are still active in updating their local variables by \eqref{eqn: FedAM update of H1 0}-\eqref{eqn: FedAM update of W2 0}, which is different from the PCP\cite{CE_DDNN_2017,FedAvg_noniid_2019} where non-selected clients are completely inactive. It will be shown that the PCC scheme actually can provide significant performance improvement, particularly in heterogeneous networks with non-i.i.d data.

The details of the proposed FedMAvg algorithm are summarized in Algorithm \ref{alg: model_avg}.

\begin{algorithm}[t!]
	\caption{Proposed FedMAvg algorithm} 
	\label{alg: model_avg}{
		\begin{algorithmic}\footnotesize 
			\STATE {\bfseries Input:} initial values of $\Wb_1^{0}=\cdots=\Wb_P^{0}$ at the server side,  initial values of $\{\Hb_p^{0}\}_{p=1}^P$ at the clients, 
			$\Ac^0 = \{1, \ldots, P\}$ and $\hat Q$.
			\FOR{round $s=1$ {\bfseries to} $S$}
			\STATE {\bfseries \underline{Server side:}} Compute
			{\small\begin{align} 
				\Wb^s = \Pc_{\Wc}\bigg(\frac{1}{m}\sum_{p \in \Ac^{s-1}}\Wb_p^{s-1}\bigg),
				\label{eqn: FedAM proj of W}
				\end{align}}
			\!\!\!\!\!\! and select a set of clients $\Ac^s$ (with size $|\Ac^s| = m$) by sampling with replacement according to probabilities $\{\omega_1, \ldots, \omega_P\}$, and broadcast 
			$\Wb^{s}$ to all clients.
			\STATE {\bfseries \underline{Client side:}}
			\FOR{client $p=1$ {\bfseries to} $P$ in parallel} 
			\STATE Set $\Hb_p^{s, 0} = \Hb_p^{s-1}$ and  $\Wb_p^{s, 0} = \Wb^s$. 
			\FOR{epoch $t = 1$ {\bfseries to} $Q_1$}
			\STATE 
			
			\vspace{-0.3cm}
			{\small\begin{align} 
				\hspace{-0.5cm}	\Hb_p^{s, t}\! &=\! \Pc_{\Hc_p} \big\{\Hb_p^{s,t-1}\! -\! \frac{\nabla_{H_p}F_p(\Wb_p^{s,t-1}, \Hb_p^{s,t-1})}{c_p^{s}}\big\}, \label{eqn: FedAM update of H1}\\
				\hspace{-0.5cm}	\Wb_p^{s, t} &= \Wb_p^{s, t-1}.\label{eqn: FedAM update of W1}
				\end{align}}
			\vspace{-0.4cm}
			
			\ENDFOR
			
			\FOR{epoch $t = Q_1 + 1$ {\bfseries to} $Q^s=Q_1+Q_2^s$}
			\STATE \vspace{-0.3cm}
			{\small\begin{align}
				\hspace{-0.1cm}&\Wb_p^{s, t} = \Wb_p^{s, t-1} - \frac{\nabla_{W}F_p(\Wb_p^{s, t-1}, \Hb_p^{s, t-1})}{ d^s},\label{eqn: FedAM update of W2}\\
				\hspace{-0.1cm}&\Hb_p^{s, t}=\Hb_p^{s, t-1}. \label{eqn: FedAM update of H2}
				\end{align}}
			\vspace{-0.4cm}
			\ENDFOR
			\STATE  Denote $\Wb_p^{s}=\Wb_p^{s, Q^s}$ and $\Hb_p^{s}=\Hb_p^{s, Q^s}$.
			\IF{client $p\in \Ac^s$}
			\STATE Upload $\Wb_p^{s}$ to the server.	
			\ENDIF
			\ENDFOR
			\ENDFOR
	\end{algorithmic}}\vspace{-1mm} 
\end{algorithm}

\vspace{-0.3cm} 
\begin{Rmk} \rm \label{rmk: model ave} 
	Rather than using the alternating minimization strategy \eqref{eqn: FedAM update of H1 0}-\eqref{eqn: FedAM update of W2 0},  one may instead approximate problem \eqref{eqn: local_prob}
	by applying $Q/2$ consecutive proximal alternating linearization minimization (PALM) steps \cite{PALM_2014} locally at each client $p$. That is, given ${\Wb}^{s,0}_p\triangleq {\Wb}^{s}$, and $\Hb^{s,0}_p\triangleq \Hb_p^{s-1,Q/2}$ in the round $s$, each client $p$ performs for $t=1,\ldots,Q/2$

	\vspace{-0.4cm}{ \begin{align}
		\Hb^{s,t}_p &=\Pc_{\Hc} \big\{\Hb_p^{s,t-1} - \frac{1}{c^{s}_p}\nabla_{H_p}F_p({\Wb}_p^{s,t-1}, \Hb_p^{s,t-1})\big\}, \label{eqn: PAML udpate of H oneshot} 
		\end{align}
		\begin{align}
		\Wb_p^{s,t} &=\Pc_\Wc \big\{{\Wb}_p^{s,t-1} - \frac{1}{d^{s}}\nabla_{W}F_{p}({\Wb}^{s,t-1}_p, \Hb_p^{s,t})\big\}. \label{eqn: PAML udpate of W oneshot}
		\end{align}}\vspace{-0.4cm}
	
	\noindent The above \eqref{eqn: PAML udpate of H oneshot}-\eqref{eqn: PAML udpate of W oneshot} are different from \eqref{eqn: FedAM update of H1 0}-\eqref{eqn: FedAM update of W2 0} in the order of updates.
	Intriguingly, our numerical experience suggests that \eqref{eqn: PAML udpate of H oneshot}-\eqref{eqn: PAML udpate of W oneshot}  may not be a good strategy. 
	To gain the insight, one can see that when $Q\to \infty$ the updates in \eqref{eqn: FedAM update of H1 0}-\eqref{eqn: FedAM update of W2 0} merely correspond to applying a single step of alternating minimization to the local problem \eqref{eqn: local_prob}, whereas applying $Q/2$ PALM steps \eqref{eqn: PAML udpate of H oneshot}-\eqref{eqn: PAML udpate of W oneshot} with $Q\to \infty$ would reach a stationary point of \eqref{eqn: local_prob} \cite{PALM_2014}.  
	Given solely locally observable data at the clients, the latter strategy in \eqref{eqn: PAML udpate of H oneshot}-\eqref{eqn: PAML udpate of W oneshot} would be too greedy and may not benefit the global algorithm convergence, especially in the presence of non-i.i.d. data.
\end{Rmk}
\subsection{Convergence Analysis of {FedMAvg}}

We first make some proper assumptions on problem \eqref{eqn: distributed prob}.  

\begin{assumption} \label{eqn: assumption1}
	All local cost functions $F_p$ are lower bounded, i.e., $F_p(\Wb,\Hb_p) \geq \underline{F} > {-\infty},~\forall~ \Wb\in \Wc, \Hb_p\in \Hc_p$, where the constraint sets $\Wc$ and $\Hc_p$, $p\in \Pc$, are compact and convex.
\end{assumption}

\begin{assumption}	\label{eqn: assumption2}
	$F_p$ are continuously differentiable in both $\Wb$ and $\Hb_p$.
	Moreover, $\nabla_{H_p} F_p(\Wb^{s}, \cdot)$ is Lipschitz continuous on $\Hc_p$ with constant $L_{H_p}^s$, and $\nabla_{W} F_p(\cdot, \Hb_p^{s,Q})$ is Lipschitz continuous on $\Wc$ with constant $L_{W_p}^s$. 
\end{assumption}

Note that by Assumption \ref{eqn: assumption2}, $\nabla_{W} F(\cdot, \Hb)$ is Lipschitz continuous with constant $L_{W}^s = \sqrt{\sum_{p=1}^{P} \omega_p (L_{W_p}^{s})^2}$. Since $\Wc$ and $\Hc_p$ are compact by Assumption \ref{eqn: assumption1}, there exist upper and lower bounds for $L_{W_p}^s$ and $L_{H_p}^s$, e.g., for all $p\in \Pc$,

\vspace{-0.5cm}
{
	\begin{align}\label{eqn: bound of Lip const}
	&\ol L_W \geq L_{W_p}^s \geq \underline{L}_{W} > 0,~\ol L_H \geq L_{H_p}^s \geq \underline{L}_{H} > 0.
	\end{align}
}
\vspace{-0.5cm}

In addition, under compact $\Wc$ and $\Hc_p$, we can have the following bounds
{\begin{align}
	&\|\nabla_{W} F_p(\Wb, \Hb_p) - \nabla_{W} F(\Wb, \Hb)\|_F^2\leq \zeta^2, \label{eqn: grad_var} \\
	&\|\nabla_{W} F(\Wb, \Hb)\|_F^2 \leq \phi^2, \label{eqn: grad_bound} 
	\end{align}}for all $\Wb \in \Wc$ and $\Hb\in \Hc$, where $\zeta$ and $\phi$ are some constants. Equation \eqref{eqn: grad_bound} means that the gradient of $F$ is bounded. Equation \eqref{eqn: grad_var} implies that the deviation between the local gradient $\nabla_WF_p$ and global gradient $\nabla_W F$ are also bounded. It is worth noting that the term in \eqref{eqn: grad_var} is usually used in the FL literature \cite{DPSGD_2017} to quantify the effect of non-i.i.d. data.
Different from \cite{DPSGD_2017} where \eqref{eqn: grad_var} and \eqref{eqn: grad_bound} are made as  assumptions, in our work we have \eqref{eqn: grad_var} and \eqref{eqn: grad_bound} to hold naturally since problem \eqref{eqn: distributed prob} is a constrained problem.

To build the convergence condition, we define the following sequence
{\begin{align}
	\wt\Wb^{s, t} = \Pc_{\Wc}\bigg(\frac{1}{m}\sum_{p \in \Ac^s}\Wb_p^{s, t}\bigg), ~\widetilde\Wb^{s, 0} = \Wb^s, \label{eqn: virtual_global}
	\end{align}}$t  = 1, \ldots, Q^s$, 
as the instantaneous weighted average of local models. Besides, 
we define the following terms as the optimality gap between a stationary solution of problem \eqref{eqn: distributed prob}

\vspace{-0.3cm}
{ \begin{align}
	&G_{H}(\wt \Wb^{s,t},\Hb^{s,t}) \triangleq \sum_{p =1 }^{P} \omega_p (c_p^s)^2\big\|\Hb_p^{s,t}- \Pc_{\Hc_p}\big(\Hb_p^{s,t} - \frac{1}{c_p^s}{\nabla_{H_p} F_p(\wt \Wb^{s,t}, \Hb_p^{s,t})}\big)\big\|_F^2,~ \forall t\in \Qc_1, \label{eqn: prox_H}\\
	&G_{W}(\wt \Wb^{s,t}, \Hb^{s, t}) \triangleq (d^s)^2\|\wt \Wb^{s, t}- \Pc_{\Wc}\big(\wt \Wb^{s, t} - \frac{1}{d^s}{\nabla_{W} F(\wt \Wb^{s,t}, \Hb^{s,t})}\big)\|_F^2, ~\forall t\in \Qc_2^s, \label{eqn: prox_W}
	\end{align}}
\vspace{-0.3cm}

\noindent where $\Qc_1 \triangleq \{1, \ldots, Q_1\}$ and $\Qc_2^s \triangleq \{Q_1 + 1, \ldots, Q^s\}$.
Note that if $G_{H}(\wt \Wb^{s,t},\Hb^{s,t})=G_{W}(\wt \Wb^{s,t}, \Hb^{s, t})=0$, then $(\wt \Wb^{s,t},\Hb^{s,t})$ is a stationary solution of problem \eqref{eqn: distributed prob}.
The main theoretical results for FedMAvg is given as follows.

\begin{Theorem} \label{thm: model_avg}
	Let $Q_2^s = \lfloor \frac{\hat{Q}}{s} \rfloor +1$ and $T = \sum_{s = 1}^{S} Q^s$ be the total number of gradient evaluations per client. Moreover, let $c_p^s = \frac{\gamma_1}{2}\ol L_H$, $d^s = \gamma_2L_{W}^s$,  where $\gamma_1 >1$ and  $\gamma_2 \geq Q_2^1\sqrt{2(7 + 4 \ol L_W^2/\underline{L}_W^2)}$, and let $\Ac^s$ (with $|\Ac^s|=m\leq P$) be obtained by  sampling with probability $\{\omega_1,\ldots,\omega_P\}$ with replacement.
	Then, under Assumptions \ref{eqn: assumption1} and \ref{eqn: assumption2}, the sequence $\{(\widetilde \Wb^{s, t}, \Hb^{s, t})\}$ of FedMAvg 
	satisfies 
	
	\vspace{-0.4cm}
	{\begin{align}
		&\frac{1}{T}\bigg[\sum_{s = 1}^{S}\sum_{t = 1}^{Q_1} \E[G_{H}(\wt \Wb^{s, t - 1}, \Hb_p^{s, t-1})] \notag \\
		&~~~~~~~~~~+\sum_{s = 1}^{S}\sum_{t = Q_1 + 1}^{Q^s}\E[G_{W}(\wt \Wb^{s, t - 1}, \Hb^{s, t-1})] \bigg]\notag 
		\\
		\leq& 
		\frac{D}{T}\bigg(F(\wt\Wb^{1, 0}, \Hb^{1, 0}) - \underline{F}\bigg)  + \bigg( \frac{8D\zeta^2}{m\gamma_2\underline{L}_W} +  \frac{96\zeta^2}{m} \bigg) \notag \\
		&+ \frac{2D(1+ 8/m)(\frac{11}{3}\zeta^2 + \phi^2)\sum_{s = 1}^{S}C_1^s}{T\gamma_2^3\underline{L}_W}  \notag \\
		& + \frac{(\frac{11}{3}\zeta^2 + \phi^2)\sum_{s = 1}^{S}C_2^s}{T\gamma_2^2} + \frac{3(\zeta^2 + \phi^2)\sum_{s = 1}^{S}C_1^s}{2T}, \label{thm: model_rate}
		\end{align}}\vspace{-0.2cm}
	
	\noindent where $D \triangleq \frac{\gamma_1^2 \ol L_{H}}{2(\gamma_1 - 1)} + \frac{6(\gamma_2^2 + 1)\ol L_W^2}{(\gamma_2 - 1)\underline{L}_W}$, and 
		\begin{subequations}
			\begin{align}
			C_1^s &\triangleq Q_2^s(Q_2^s - 1)(2Q_2^s -1), \label{eqn C1s}\\
			C_2^s &\triangleq 6(3Q_2^s(Q_2^s - 1)/2 + 4 + 32/m)C_1^s.\label{eqn C2s}
			\end{align}
	\end{subequations}
\end{Theorem}

{\bf Proof:} See Appendix \ref{proof of fedmavg}.

The bound in \eqref{thm: model_rate} shows that the local GD length $Q_1$, $Q_2^s$, non-i.i.d. data $\zeta$ and PCC $m$ all have strong impacts on the convergence of FedMAvg. One can notice that if constant $Q_2^s=Q_2=1$ is used, then $\sum_{s = 1}^{S}C_1^s=\sum_{s = 1}^{S}C_2^s=0$, which makes the last three terms in the right hand side (RHS) of \eqref{thm: model_rate} vanish to zero. 
On the other hand, if $T = SQ_1+\sum_{s = 1}^{S} Q_2^s$ is fixed, then increasing $Q_1$ or $Q_2^s$ can potentially reduce the required number of communication rounds $S$. Thus, there need proper choices of $Q_1$ or $Q_2^s$ for a good trade off between convergence performance and communication efficiency. 
One can see that if constant $Q_2^s=Q_2>1$ is used, then both $\sum_{s = 1}^{S}C_1^s$ and $\sum_{s = 1}^{S}C_2^s$ linearly increase with $S$ and are unbounded. On the contrary, if $Q_2^s=\lfloor \frac{\hat Q}{s}\rfloor+1$, then one can show ({see \cite[Section 3]{FedC}}) that both 
$
\sum_{s = 1}^{S} C_1^s = \mathcal{O}(\hat{Q}^3)<\infty, ~\sum_{s = 1}^{S} C_2^s = \mathcal{O}( \hat{Q}^5)<\infty,
$ and thereby the last three terms in the RHS of \eqref{thm: model_rate} can decrease sublinearly when $T$ is large.
Therefore, the diminishing $Q_2$ strategy can trade off between convergence performance and communication efficiency in a better way than constant $Q_2$.
%
%
%
%

It can also been seen from the 2nd term in the RHS of  \eqref{thm: model_rate} that PCC would slow down the algorithm convergence, especially under non-i.i.d. data. Interestingly, when $m$ increases, such negative effect can be reduced. In fact, when $|\Ac^s|=|m|=P$ (full participation), the term ${\footnotesize \big( \frac{8D\zeta^2}{m\gamma_2\underline{L}_W} +  \frac{96\zeta^2}{m} \big)}$ vanishes, and 
FedMAvg can converge to a stationary solution problem \eqref{eqn: distributed prob} in a sublinear rate. 

 \begin{Corollary} \label{corly: model_avg}
	Consider full participation in FedMAvg, i.e., $|\Ac^s| = P, ~\forall s$, and the use of weighted average like \eqref{eqn: FedAM proj of W 0}. Let $\wt \Wb^{s, t} = \Pc_{\Wc}\big(\sum_{p =  1}^{P} \omega_p\Wb_p^{s, t}\big)$ for all $s$ and $t$.
	Then, under the same setting as Theorem \ref{thm: model_avg}, the sequence $\{(\widetilde \Wb^{s, t}, \Hb^{s, t})\}$ of FedMAvg satisfies
	
	\vspace{-0.3cm}
	{\begin{align}
		&\frac{1}{T}\bigg[\sum_{s = 1}^{S}\sum_{t = 1}^{Q_1}G_{H}(\wt \Wb^{s, t-1}, \Hb^{s, t-1}) \notag \\
		&~~~~~~~~~~+\sum_{s = 1}^{S}\sum_{t = Q_1 + 1}^{Q^s}G_{W}(\wt \Wb^{s, t-1}, \Hb^{s, t-1})\bigg] \notag\\
		\leq &   \frac{D}{T}\bigg(F(\wt\Wb^{s,0}, \Hb^{s, 0}) - \underline{F} \bigg) \notag \\
		&+ \frac{1}{T} \bigg [  \frac{3}{2}({\zeta^2} + \phi^2)\sum_{s = 1}^{S}C_1^s +  \frac{(\frac{11\zeta^2}{3} + \phi^2)}{\gamma_2^2} \bigg( \frac{D \sum_{s = 1}^{S}C_1^s}{\gamma_2\underline{L}_W} + \sum_{s = 1}^{S}C_3^s\bigg) \bigg]. \label{corly: model_avg_rate}
		\end{align}}
	\vspace{-0.2cm}

	\noindent where {$C_3^s \triangleq 6(3Q_2^s(Q_2^s - 1)/2 + 2)C_1^s$}, {$D$ and $C_1^s$ are defined in Theorem 1.}
\end{Corollary}

{\bf Proof:} See \cite[Section 2]{FedC}.

\section{Federated MF by Gradient Sharing}\label{sec: FedMGS}
In this section, we focus on MF models with the squared Frobenius norm loss function $\Phi(\Xb, \Wb\Hb)=\frac{1}{N} \|\Xb - \Wb\Hb\|_{F}^2$.
By carefully exploiting the linear structure of $\nabla_W \Phi$, we 
present another FedMF algorithm, termed {FedMGS}, and its convergence analysis.

\vspace{-0.5cm}
\subsection{{FedMGS} Algorithm}
According to the GS principle  \cite{CE_DDNN_2017}, for problem \eqref{eqn: distributed prob} each client $p$ should compute the gradient $\nabla_W F_p$ and upload it to the server. After collecting $\nabla_W F_p$, $p\in \Pc$, the server performs one step of PGD with respect to $\Wb$, and broadcasts the new $\Wb$ to the clients. More specifically, given $\Wb^{s-1}$ at the clients, the client $p$ computes $\nabla_W F_p(\Wb^{s-1},\Hb_p^s)$ where

\vspace{-0.3cm} 
\begin{align}
\Hb_p^s&=\arg\min_{\substack{\Hb_p\in \Hc_p}} ~F_p({\Wb}^{s-1}, \Hb_p). \label{eqn: AM H2}
\end{align}
\vspace{-0.3cm}

\noindent The PGD performed by the server is

\vspace{-0.3cm}
{\begin{align}	
	\Wb^{s} &=\Pc_\Wc \bigg\{{\Wb}^{s-1} - \frac{1}{d^{s}}\sum_{p=1}^P\omega_p \nabla_{W}F_p({\Wb}^{s-1}, \Hb_p^{s})\bigg\}. \label{eqn: PAML udpate of W oneshot0}
	\end{align}}
\vspace{-0.3cm}

Analogous to \eqref{eqn: FedAM update of H1 0}, we can approximate \eqref{eqn: AM H2} by $Q_1\geq 1$ consecutive PGD steps with respect to $\Hb_p$; specifically, given  $\Hb_p^{s, 0}=\Hb_p^{s-1}$, each client $p$ performs for $t\in \Qc_1$, 

\vspace{-0.3cm}
{\begin{align}
	\hspace{-0.1cm}	&\Hb_p^{s, t}\! =\! \Pc_{\Hc_p} \big(\Hb_p^{s,t-1}\! -\! \frac{1}{c_p^{s}}{\nabla_{H_p}F_p(\Wb^{s-1}, \Hb_p^{s,t-1})}\big), \label{eqn: FedAM update of H1 2}
	\end{align}}
\vspace{-0.3cm}

\noindent and obtain $\Hb_p^{s}=\Hb_p^{s, Q_1}$. 
For example, the FedMF algorithm in  \cite{SFMF_2019,DRMF_CF_2019} adopts $Q_1=1$. However, as pointed out in \cite{CE_DDNN_2017}, the GS scheme will need a lot of communication rounds to produce a good model. 
To overcome this, we not only require $Q_1>1$, but also require the server to perform multiple steps of \eqref{eqn: PAML udpate of W oneshot0}, which however is not possible in general since the server cannot access the data and thus cannot obtain the new gradient $\nabla_W F(\cdot,\Hb^s)$. 

{\bf Linear gradient structure:} Intriguingly, for problem \eqref{eqn: distributed prob} with $\Phi(\Xb, \Wb\Hb)=\frac{1}{N} \|\Xb - \Wb\Hb\|_{F}^2$, we actually can allow the server to conduct multiple steps of PGD with respect to $\Wb$ in each communication round  if the linear gradient structure of $\nabla_W F$ is utilized.
Specifically, by \eqref{eqn: obj of client p}, we have

\vspace{-0.3cm}
{\begin{align}\label{eqn: gradient of F wrt W}
	&\nabla_{\Wb} F(\Wb, \Hb) = \sum_{p=1}^P \omega_p \nabla_W F_p(\Wb, \Hb_p)\notag \\
	&= 2\Wb \sum_{p = 1}^{P} \frac{\Hb_p\Hb_p^\top}{N}  - 2 \sum_{p=1}^{P}\frac{\Xb_p\Hb_p^\top}{N} + \nabla R_W(\Wb).
	\end{align}}
\vspace{-0.3cm}

\noindent Thus, it is sufficient for each client $p$ to send $\Hb_p^{s}(\Hb_p^{s})^\top$ and $\Xb_p(\Hb_p^{s})^\top$ to the server, who is then able to construct the gradient $\nabla_{W}F(\cdot,\Hb^{s})$
on its own by \eqref{eqn: gradient of F wrt W}. With this ability, the server can perform multiple PGD steps in each communication round. Particularly, given $\Wb^{s, 1}=\ldots=\Wb^{s, Q_1}=\Wb^{s-1}$, we let the server perform $Q_2 \geq 1$ consecutive steps of PGD, i.e., for $t\in \Qc_2=\{Q_1+1,\ldots,Q\}$

\vspace{-0.3cm}
{\begin{align}	
	\Wb^{s,t} &=\Pc_\Wc \big\{{\Wb}^{s,t-1} - \frac{1}{d^{s}}\nabla_{W}F({\Wb}^{s,t-1}, \Hb^{s, Q_1})\big\}. \label{eqn: PAML udpate of W oneshot2}
	\end{align}}\vspace{-0.3cm}

{\bf Partial client participation (PCP):}
We let the server select only a small subset of clients $\Ac^s$ (with size $m \ll P$) to participate in the FedMF task in each communication round. 
Different from the PCC in FedMAvg, the clients that are not selected in $\Ac^s$ are completely inactive in FedMGS (who neither perform local updates nor upload message to the server).

The FedMGS algorithm are summarized in Algorithm \ref{alg: gradient_share}.
\begin{algorithm}[H]
	\caption{Proposed {FedMGS} Algorithm} 
	\label{alg: gradient_share}
	\begin{algorithmic}\footnotesize 
		\STATE {\bfseries Input:} Initial values of $\Hb_1^{0,Q},\ldots,\Hb_P^{0,Q}$ at the clients,
		and initial value of $\Wb^{0,Q}$ 
		and
		
		\vspace{-0.3cm}
		{ \begin{align*}
			\Gb_1^0 = \sum\limits_{p = 1}^{P}\frac{2}{N}\Hb_p^{0, Q_1}(\Hb_p^{s, Q_1})^\top,~\Gb_2^0 = \sum\limits_{p = 1}^{P}\frac{2}{N}\Xb_p(\Hb_p^{0, Q_1})^\top,
			\end{align*}} 
		\vspace{-0.3cm}
		
		at the server.
		
		\FOR{round $s=1$ {\bfseries to} $S$}
		\STATE {\bfseries \underline{Server side:}} Select a 
		subset of clients $\Ac^s \subset \Pc$ (with size $|\Ac^s| = m$), and broadcast 
		$\Wb^{s}=\Wb^{s-1,Q}$ to the clients in $\Ac^s$.
		
		\STATE {\bfseries \underline{Client side:}}~   
		\FOR{client $p=1$ {\bfseries to} $P$ in parallel} 
		\IF{client $p\notin \Ac^s$}
		\STATE Set $\Hb_p^{s, t} = \Hb_p^{s-1, Q_1},~\forall t\in \Qc_1.$

		\ELSIF{client $p\in \Ac^s$}
		
		\STATE Set $\Hb_p^{s, 0} = \Hb_p^{s-1,Q}$.
		\FOR{epoch $t = 1$ {\bfseries to} $Q_1$}
		\STATE

		\vspace{-0.3cm}
		{\begin{align}
			\Hb_p^{s, t}\! =\! \Pc_{\Hc_p} \big\{\Hb_p^{s,t-1}\! -\! \frac{1}{c_p^{s}}{\nabla_{H_p}F_p(\Wb^{s}, \Hb_p^{s,t-1})}\big\}. \notag
			\end{align}}
		\vspace{-0.3cm}

		\ENDFOR
		\STATE Send the server 
		
		\vspace{-0.3cm}
		{\begin{align}\label{eqn: differential info 1}
			&\Ub_{p}^s=\Hb_p^{s, Q_1}(\Hb_p^{s, Q_1})^\top,~
			\Vb_{p}^s=\Xb_p(\Hb_p^{s, Q_1})^\top.
			\end{align}}
		\vspace{-0.3cm}
		

		\ENDIF

		\ENDFOR
		\STATE {\bfseries \underline{Server side:}}
		\STATE Set $\Wb^{s, t} = \Wb^{s},~ \forall t\in \Qc_1$, and compute 
		
		\vspace{-0.3cm}
		{\begin{align}
			&\Gb_1^s = \Gb_1^{s-1} + \frac{2}{N}\!\!\sum_{p \in \Ac^s}(\Ub_p^s -\Ub_p^{s-1}), \notag \\&\Gb_2^s = \Gb_2^{s-1}+\frac{2}{N}\!\!\sum_{p \in \Ac^s} (\Vb_p^s - \Vb_p^{s-1}).\notag 
			\end{align}}
		\vspace{-0.3cm}
		
		\FOR {epoch $t=Q_1 + 1$ {\bfseries to} $Q$}
		\STATE 
		
		\vspace{-0.3cm}
		{\begin{align}
			\label{eqn: fedCALM update of W1}
			&\!\!\!\!\!\!\Wb^{s, t} = \Pc_{\Wc}\{\Wb^{s, t-1} \!- \!\frac{1}{ d^s}{\nabla_{W}F(\Wb^{s, t-1}, \Hb^{s, Q_1})}\},
			\end{align}}		\vspace{-0.2cm}
		
		\noindent where 
		$\nabla_{W}F(\Wb^{s, t-1},\Hb^{s,Q_1}) = \Wb^{s, t-1}\Gb_1^s - \Gb_2^s +\nabla R_W(\Wb^{s,t-1})$.

		\ENDFOR
		\ENDFOR
	\end{algorithmic}
\end{algorithm}
\vspace{-0.0cm}

\vspace{-0.5cm}
\subsection{Convergence Analysis of FedMGS}

Here we establish the convergence conditions of FedMGS. For PCP, we assume that the server samples the clients with a uniform probability without replacement to obtain $\Ac^s$  in each communication round\footnote{Our analysis considers this uniform sampling without replacement, but can be easily extended to the case when the server samples clients using probability $\{\omega_1,\ldots,\omega_P\}$ with replacement.}.
%
Besides, we define the following virtual sequence assuming that all clients are active  in each round $s$, i.e., for all $p\in \Pc$ and $ t\in \Qc_1$,

\vspace{-0.3cm}
{\begin{align}\label{smeqn: virtual H}
	&\wt\Hb_p^{s, t} = \Pc_{\Hc_p} \big\{\wt\Hb_p^{s,t-1} - \frac{1}{c_p^{s}}{\nabla_{H_p}F_p(\Wb^{s,0}, \wt\Hb_p^{s,t-1})}\big\},  \notag \\
	&\wt\Hb_p^{s, 0} = \Hb_p^{s, 0}.
	\end{align} }
\vspace{-0.3cm}

The convergence result for FedMGS is stated below.

\begin{Theorem} \label{thm: gradient_share}
	Let $c_p^s = \frac{\gamma}{2}L_{H_p}^s$ and $d^s = \frac{\gamma}{2}L_{W}^s$, where  $\gamma > 1$, and that $\Ac^s$ (with $|\Ac^s|=m\leq P$) is obtained by uniform sampling without replacement. Then, under Assumptions \ref{eqn: assumption1} and \ref{eqn: assumption2}, 
	we have for FedMGS
	
	\vspace{-0.3cm}
	{\begin{align} 
		&\frac{1}{T}\bigg[\sum_{s = 1}^{S} \sum_{t = 1}^{Q_1}\E[G_{H}(\Wb^{s, t-1}, \wt \Hb^{s, t-1})] \notag 
		\\
		&~~~~~~~~~+\sum_{s = 1}^{S}\sum_{t = Q_1 + 1}^{Q} \E[G_{W}(\Wb^{s, t-1}, \Hb^{s, t-1})]\bigg] \notag \\
		\leq & \frac{1}{T}\bigg(\frac{P\gamma^2 \ol L_{H}}{2m(\gamma - 1)} + \frac{\gamma^2 \ol L_W}{2(\gamma - 1)}\bigg) \bigg(F(\Wb^{1, 0}, \Hb^{1, 0})- \underline{F}\bigg). \label{thm2: conv_rate} 	
		\end{align}}
	\vspace{-0.4cm}
\end{Theorem}

{\bf Proof:} {See \cite[Section 4]{FedC}}.

Since the RHS of \eqref{thm2: conv_rate}  is bounded and can decrease to zero as $T\to \infty$, 
both $\E[G_{H}(\Wb^{s, 0}, \wt \Hb^{s, 0}) ]= \E[G_{H}(\Wb^{s, 0}, \Hb^{s, 0})] \rightarrow 0$ and $\E[G_{W}(\Wb^{s, 0}, \Hb^{s, 0})] \rightarrow 0$ as $s \rightarrow \infty$;  this implies that FedMGS will converge to a stationary point of problem \eqref{eqn: distributed prob}. 
Interestingly, in contract to the FedMAvg which could suffer from the non-i.i.d. data, we can see that FedMGS is resilient to non-i.i.d. data. This point will be further examined via numerical experiments later. Besides, one can also see from \eqref{thm2: conv_rate} that  $Q_1>1$ and $Q_2>1$ can reduce the number of communication rounds if $T$ is fixed. 


\vspace{-0.2cm}
\begin{Rmk}\rm \label{remark 3}
	Although both FedMAvg and FedMGS are based on alternating minimization and gradient descent, they adopt very different strategies (MA and GS) for learning over the federated network. Theorem \ref{thm: model_avg} and Theorem \ref{thm: gradient_share} also suggest that the two algorithms have different convergence properties in the presence of non-i.i.d. data and  partial active clients. Moreover, the experiment results in Section \ref{sec: simulation} will show that FedMGS can exhibit favorable convergence behaviors than FedMAvg and better clustering performance. However, we should emphasize that FedMGS is restricted to problem \eqref{eqn: distributed prob} with the linear gradient structure in \eqref{eqn: gradient of F wrt W}. By contrast, FedMAvg can handle a broader range of MF problems of the form of \eqref{eqn: distributed prob} not limited to specific structured cost functions. 
	In the future, it will be interesting to apply FedMAvg to MF models with different cost functions such as the $\beta$-divergence \cite{NMF_beta_2011}.
	%
\end{Rmk}

\section{Experiment Results}
\label{sec: simulation}

In this section, we examine the convergence behavior and performance of the proposed algorithms by applying them to the data clustering problem described in Section \ref{sub: clustering}.
\vspace{-0.2cm}
\subsection{Experiment setup}
{\bf Model:} We consider the orthogonal NMF based clustering model in \cite[Eqn. (9)]{Shuai_SNCPJ_2019} which corresponds to 
problem \eqref{eqn: distributed prob}
with $R_W(\Wb)=0$,
%
%

\vspace{-0.4cm}
{\begin{align}
	&R_H(\Hb_p) \!=\! \frac{\rho}{2} \sum_{j=1}^{N_p}\bigg(\|\oneb^T\hb_{p,j}\|_2^2 - \|\hb_{p,j}\|_2^2\bigg)\! +\! \frac{\nu}{2} \|\Hb_p\|_F^2,  \label{sim: R(H)}\\
	& \Phi(\Xb, \Wb\Hb)=\frac{1}{N} \|\Xb - \Wb\Hb\|_{F}^2, \\
	&\Wc = \{\Wb \in \mathbb{R}^{M \times K} | \overline{W} \geq [\Wb]_{ij} \geq \underline{W}, \forall i,j \},  \notag \\
	&\Hc_p = \{\Hb_p \in \mathbb{R}^{K \times N_p}| [\Hb_p]_{ij} \geq 0, \forall i,j\}, \notag 
	\end{align}}
\vspace{-0.4cm}

\noindent where  
$\overline{W}$ (resp. $\underline{W}$) is set to the maximum (resp. minimum) value of $\Xb$, and $\rho, \nu > 0$ are two penalty parameters. 
If not mentioned specifically, we set $\rho=10^{-8} \times \frac{\|\Xb\|_F^2}{N}$ and $\nu = 10^{-10} \times \frac{\|\Xb\|_F^2}{N}$. Detailed explanation of $R_H(\Hb_p)$ and choice of parameters can be found in \cite{Shuai_SNCPJ_2019}. State-of-the-art distributed clustering methods will also be considered as benchmarks.

{\bf Datasets:} Four kinds of datasets are considered for evaluation, including synthetic data, the TDT2 data \cite{LCCF_2011}, the TCGA data \cite{TCGA_CGCD}, and the MNIST data \cite{website_MNIST}. 
Specifically, we follow the Gaussian linear model $\Xb = \Wb\Hb + \Eb$ in \cite{JNKM_2017} to generate a synthetic dataset with $M = 2000, N = 10000$ and $K=20$, where $\Eb \in \mathbb{R}^{M \times N}$ denotes the Gaussian noise and the signal to noise ratio (SNR) $=10\log_{10}(\|\Wb\Hb\|_F^{2}/\|\Eb\|_F^{2})$ dB is set to $-3$ dB. The TDT2 dataset is extracted from the TDT2 corpus which contains 9394 documents in the largest 30 categories, i.e. $K = 30, N = 9394, M = 5000$. The TCGA dataset is obtained from the Cancer Genome Atlas (TCGA) database which contains the gene expression data of 5314 cancer samples belonging to 20 cancer types, i.e. $K = 20, N = 5314, M = 5000$. Note that the features of the TCGA and TDT2 datasets are chosen as top-ranked ones by the Pearson’s Chi-Squared Test. Lastly, following \cite{FedAvg_noniid_2019}, we generate a MNIST dataset with $K = 10, N=10000, M = 784$. 

We distribute the samples of each dataset to $P=100$ clients in two ways: \textbf{Case 1}: we follow \cite{DKCoreset_2013} to obtain balanced and i.i.d. distributed data for the four datasets, respectively. 
\textbf{Case 2}: For the synthetic, TDT2 and TCGA datasets, we follow the similarity-based partition \cite{DKCoreset_2013} where the K-means algorithm is applied to the dataset to cluster it into $100$ clusters, and each of the cluster is assigned to one client. This leads to a highly unbalanced and non-i.i.d. dataset. For the MNIST dataset, we follow \cite{FedAvg_noniid_2019} to obtain a distributed data where each of the client
contains images of two digits only and the numbers of samples among clients are highly unbalanced. 

{\bf Parameter setting:}  
For FedMAvg, the step size $c_p^s$ and $d^s$ are set to $c_p^s=\frac{1}{2}\lambda_{\max}((\Wb_p^{s,0})^\top \Wb_p^{s,0})$, $d^s=5\lambda_{\max}(\Hb^{s, Q_1}(\Hb^{s, Q_1})^\top)$.
For FedMGS, it is set to $c_p^s=\frac{1}{2}\lambda_{\max}((\Wb^{s,0})^\top \Wb^{s,0})$ and $d^s=\frac{1}{2}\lambda_{\max}(\Hb^{s, Q_1}\\(\Hb^{s, Q_1})^\top)$.
The stopping condition for both algorithms is that the normalized change of the objective value
$\varepsilon = \frac{|F(\Wb^{s}, \Hb^{s, Q}) - F(\Wb^{s-1}, \Hb^{s-1, Q})|}{F(\Wb^{s-1}, \Hb^{s-1, Q})}$
is smaller than $10^{-8}$ or $500$ communication rounds are achieved.
All algorithms under test are initialized with 10 common, randomly generated initial points, and the averaged results are presented.

{\bf Communication cost:}
Only the uplink communication cost is considered since it is the primary bottleneck when $P$ is large.
We define the communication cost as the accumulated number of real values sent to the server.  For the $s$th round, the accumulated communication cost of FedMAvg is $s(mMK)$ while that of FedMGS is $s(mMK+mK^2)$.

Due to limited space, here we present results on the synthetic and the TCGA dataset only while relegating  the results on the TDT2 and MNIST datasets in \cite{FedC}. The simulation codes are available at \url{https://github.com/wshuai317/FedMF}.

\begin{figure} [t!]
	\centering
	\subfigure[\scriptsize syn, \textbf{Case 1}, $m=100$]{
		\includegraphics[width=6cm]{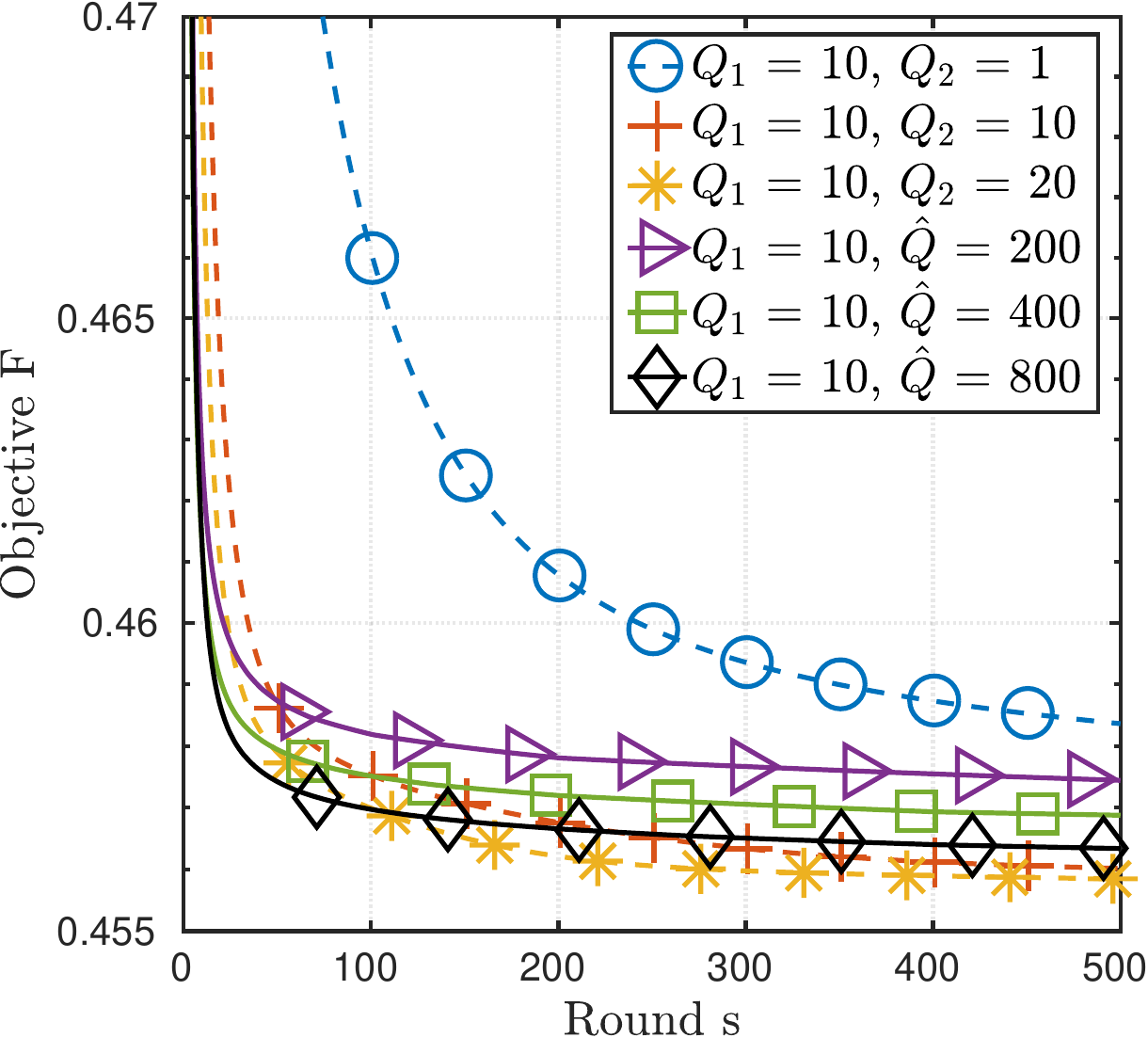}\label{fig:avg_con curves a}
	}\vspace{-0.1cm}	
	\subfigure[\scriptsize syn, \textbf{Case 2}, $m=100$]{
		\includegraphics[width=6cm]{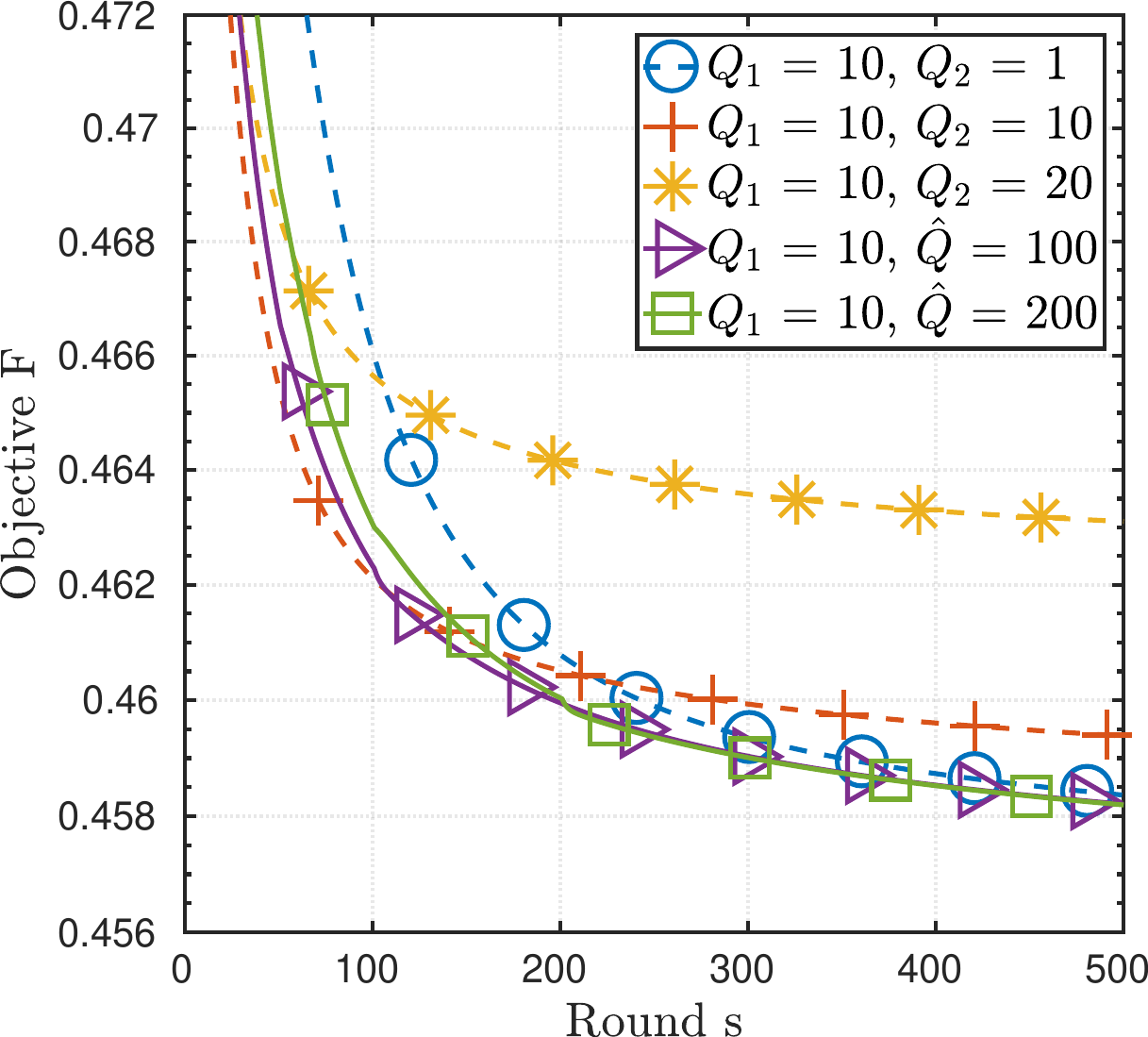} \label{fig:avg_con curves b}
	}\vspace{-0.1cm}
	\subfigure[\scriptsize  syn, \textbf{Case 2}, $m = 10$]{
		\includegraphics[width=6cm]{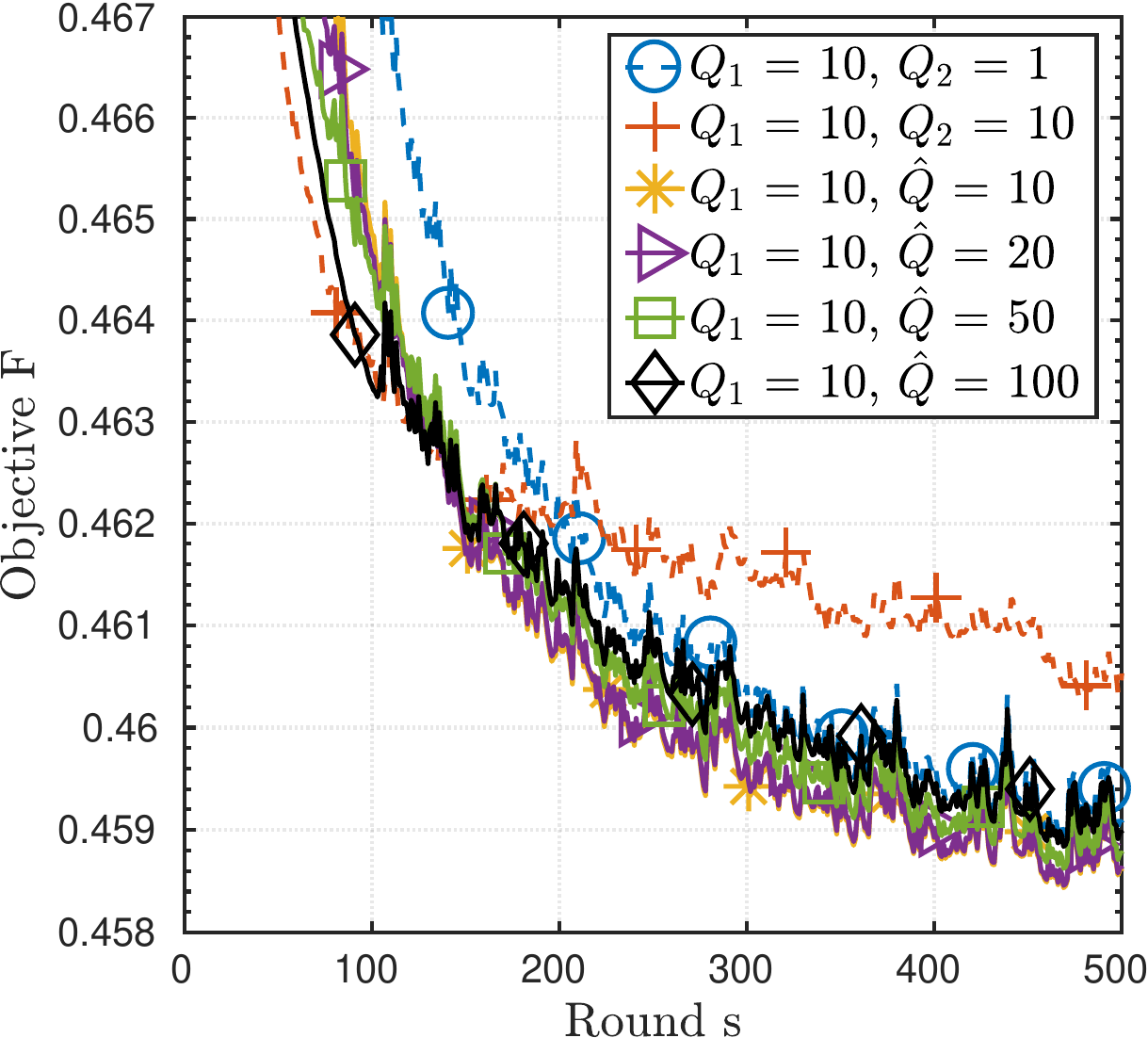} \label{fig:avg_con curves d}
	}\vspace{-0.1cm}
	\subfigure[\scriptsize syn, \textbf{Case 2}]{
		\includegraphics[width=6cm]{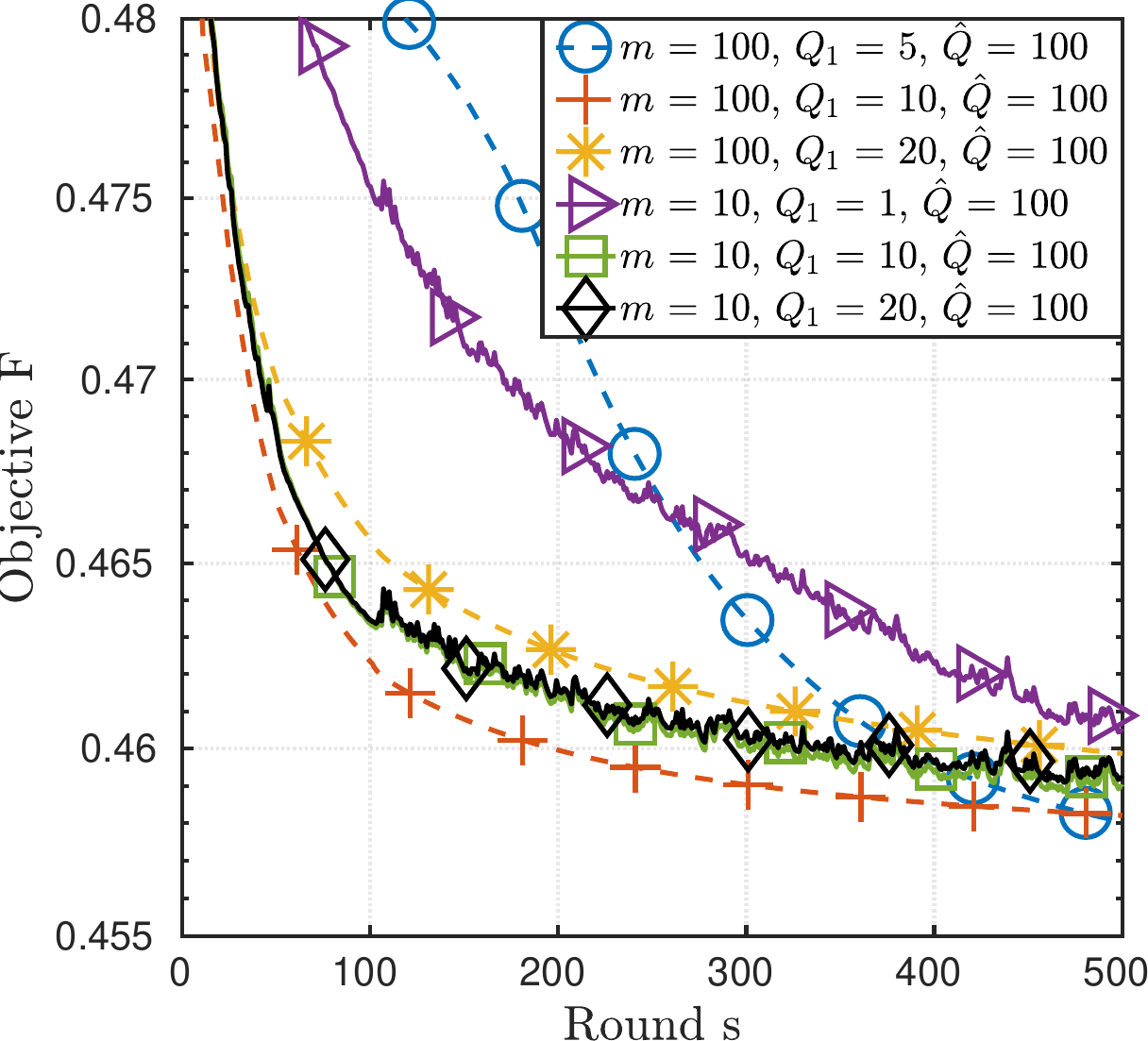} \label{fig:avg_con curves f}
	}\vspace{-0.1cm}
	\centering \caption{Convergence curve versus number of rounds of FedMAvg with different values of $Q_1$ and $\hat Q$. }\label{fig:FedAM_con curves}
	\vspace{-0.3cm}
\end{figure}

\begin{figure} [t!]
	\centering
	\subfigure[\scriptsize syn, \textbf{Case 2}, $m = 10$]{
		\includegraphics[width=6cm]{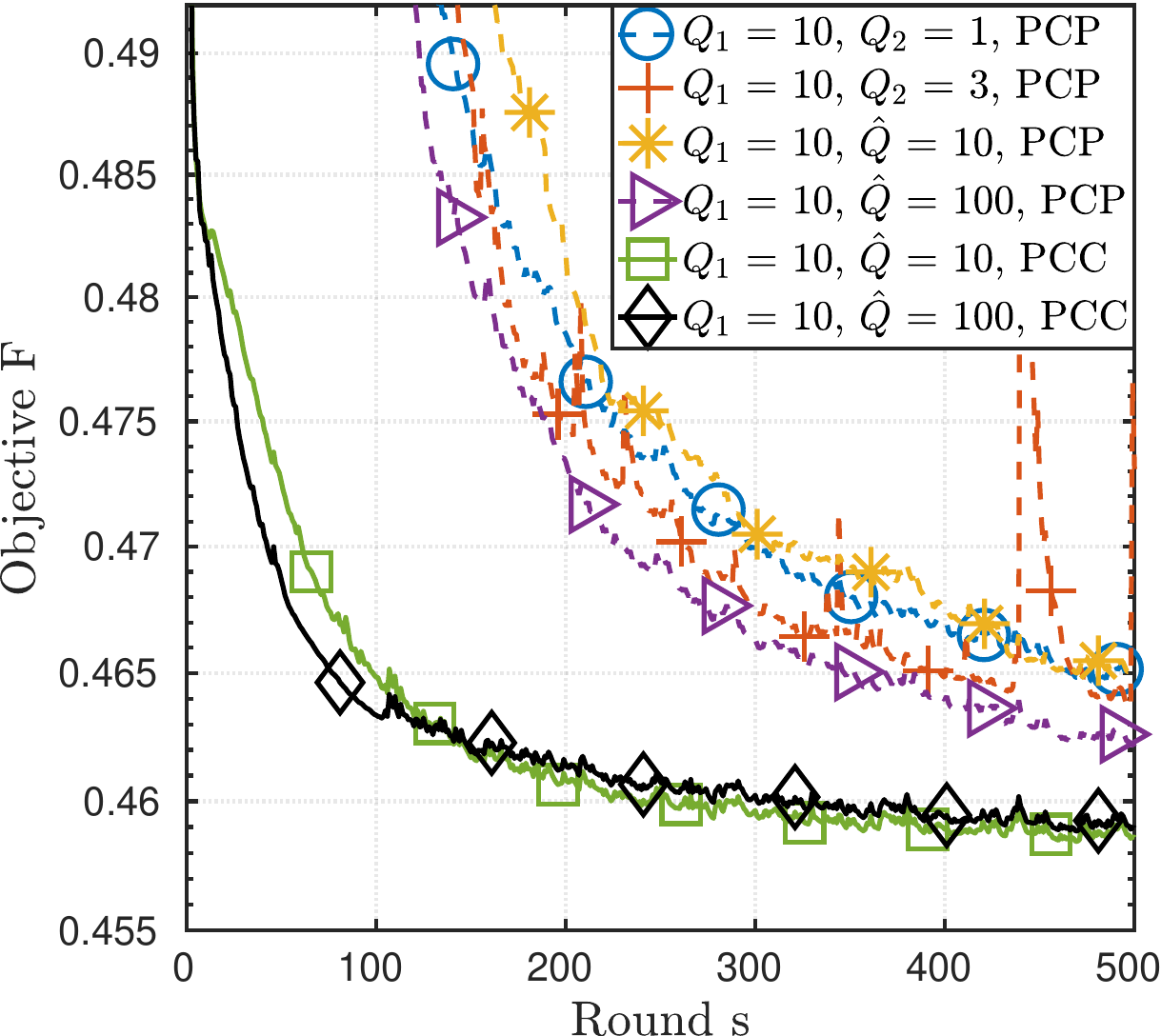} \label{fig:avg_con curves g}
	}\vspace{-0.1cm}
	\subfigure[\scriptsize syn, \textbf{Case 1}, $m = 100$]{
		\includegraphics[width=6cm]{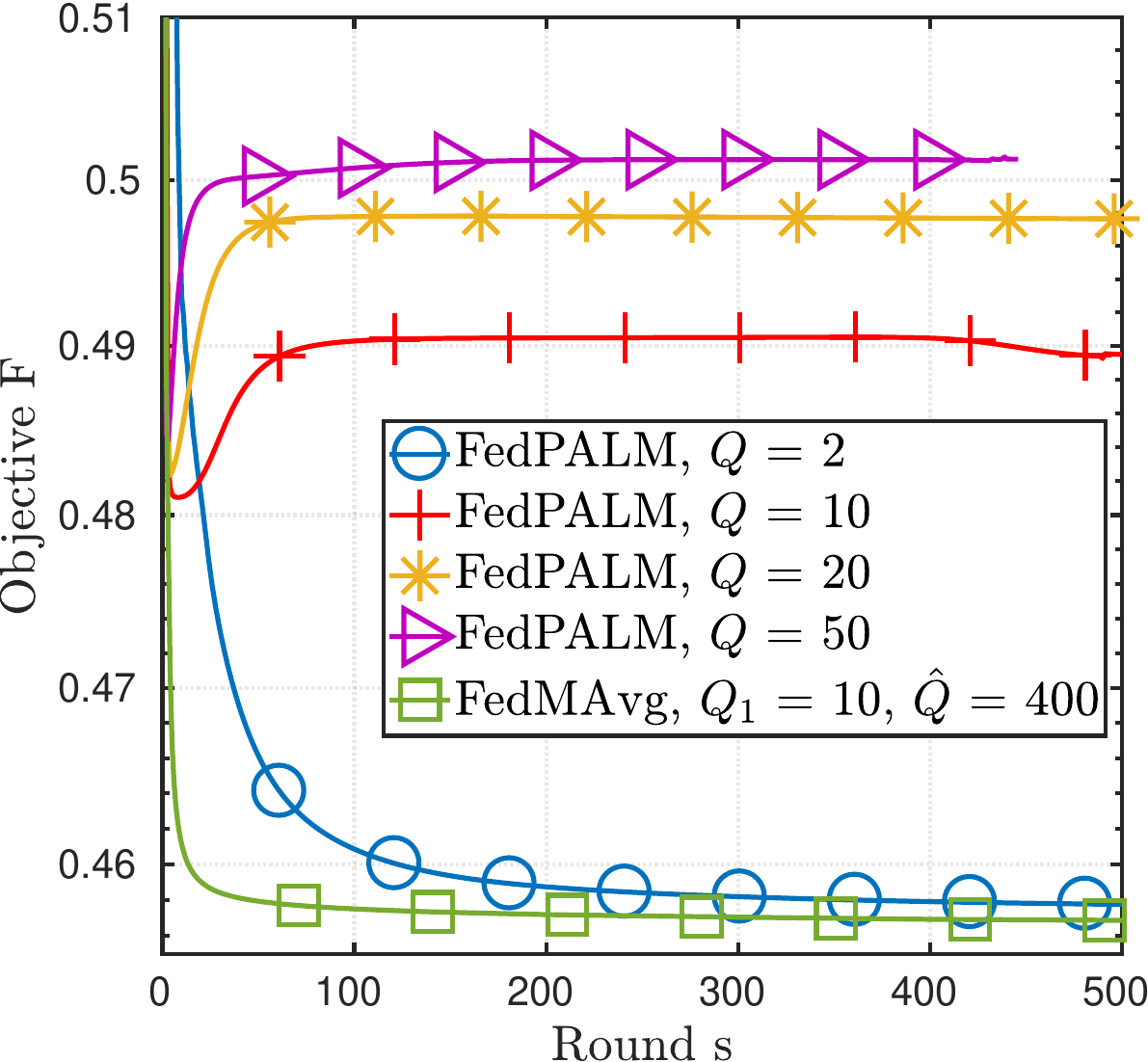} \label{fig:FedCPALM_con_curves_h}
	} \vspace{-0.1cm}
	\centering \caption{(a) Comparison between FedMAvg (PCP) and FedMAvg (PCC), and (b) comparison of FedMAvg and naive FedPALM mentioned in Remark \ref{rmk: model ave}. }\label{fig:FedAM_con curves}
	\vspace{-0.4cm}
\end{figure}  

\vspace{-0.3cm}
\subsection{Convergence of FedMAvg}
\label{subsec: conv of FedAM}
\vspace{-0.0cm}

In Fig. \ref{fig:avg_con curves a} and Fig. \ref{fig:avg_con curves b}, we present the convergence curves of FedMAvg for $Q_1=10$ and for constant $Q_2$ and diminishing $Q_2$, on \textbf{Case 1} and \textbf{Case 2} of synthetic data, respectively.
One can observe from Fig. \ref{fig:avg_con curves a} that for constant $Q_2$, increasing $Q_2$ can speed up the convergence under \textbf{Case 1} (i.i.d. data). On the contrary, one can see from Fig. \ref{fig:avg_con curves b} that under \textbf{Case 2} (non-i.i.d. data), increasing $Q_2$ can greatly cause  larger floors, whereas, with diminishing $Q_2$, a proper value of $\hat Q$ can not only speed up the convergence but also achieve a lower objective value. The choice of $\hat Q$ may depend on the dataset. As shown in  \cite[Fig. S1(a)]{FedC}, a small value of $\hat Q=5$ can achieve a good convergence behavior for the TCGA dataset.

Fig. \ref{fig:avg_con curves d} considers the PCC scheme with $m=10$. 
By comparing Fig. \ref{fig:avg_con curves d} with Fig. \ref{fig:avg_con curves b}, one can see that the degradation due to constant $Q_2$ becomes more evident when only partial clients communicate with the server, whereas the diminishing $Q_2$ scheme with a smaller value of $\hat{Q}$ can converge well.
Moreover, as displayed in Fig. \ref{fig:avg_con curves f}, increasing $Q_1$ property can also speed up the convergence for both full client  participation and PCC. While Fig. \ref{fig:avg_con curves f} is for \textbf{Case 2}, the same trend can be observed for \textbf{Case 1}; see \cite[Fig. S1(b)]{FedC}.
The above results well corroborate with Theorem \ref{thm: model_avg}. 

As discussed in Section \ref{sec: fedam development}, the adopted PCC scheme can yield better performance than the PCP scheme. To verify this, we present in Fig. \ref{fig:avg_con curves g} the comparison results of FedMAvg with PCC and PCP, respectively. One can observe that FedMAvg with PCC can significantly outperform that with PCP. 

%

Lastly, we verify the discussion in Remark \ref{rmk: model ave} that the proposed FedMAvg with sequential updating in \eqref{eqn: FedAM update of H1}-\eqref{eqn: FedAM update of H2} is better than the direct application of PALM \eqref{eqn: PAML udpate of H oneshot}-\eqref{eqn: PAML udpate of W oneshot}.  
In Fig. \ref{fig:FedCPALM_con_curves_h}, we denote the latter approach as ``FedPALM", and one can clearly see that FedPALM cannot perform well except for $Q=2$ (one step of updates of $\Wb_p$ and $\Hb_p$ per client).


\begin{figure} [t!]
	\centering
	\subfigure[\scriptsize syn, \textbf{Case 1}, $m = 100$]{
		\includegraphics[width=6cm]{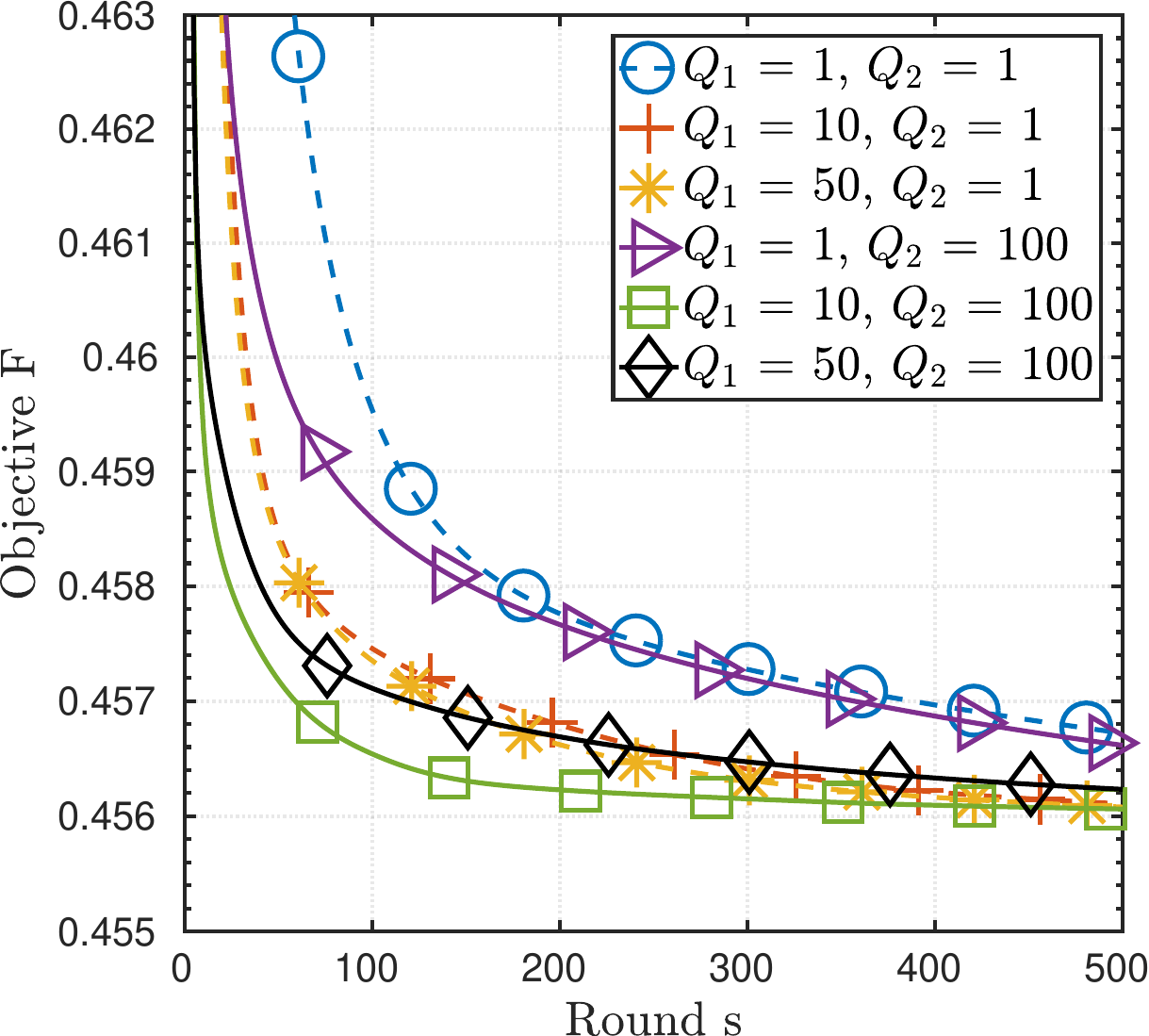} \label{fig: gs1_con_curves_a}
	} \vspace{-0.1cm}
	\subfigure[\scriptsize syn, \textbf{Case 1}, $m = 10$]{
		\includegraphics[width=6cm]{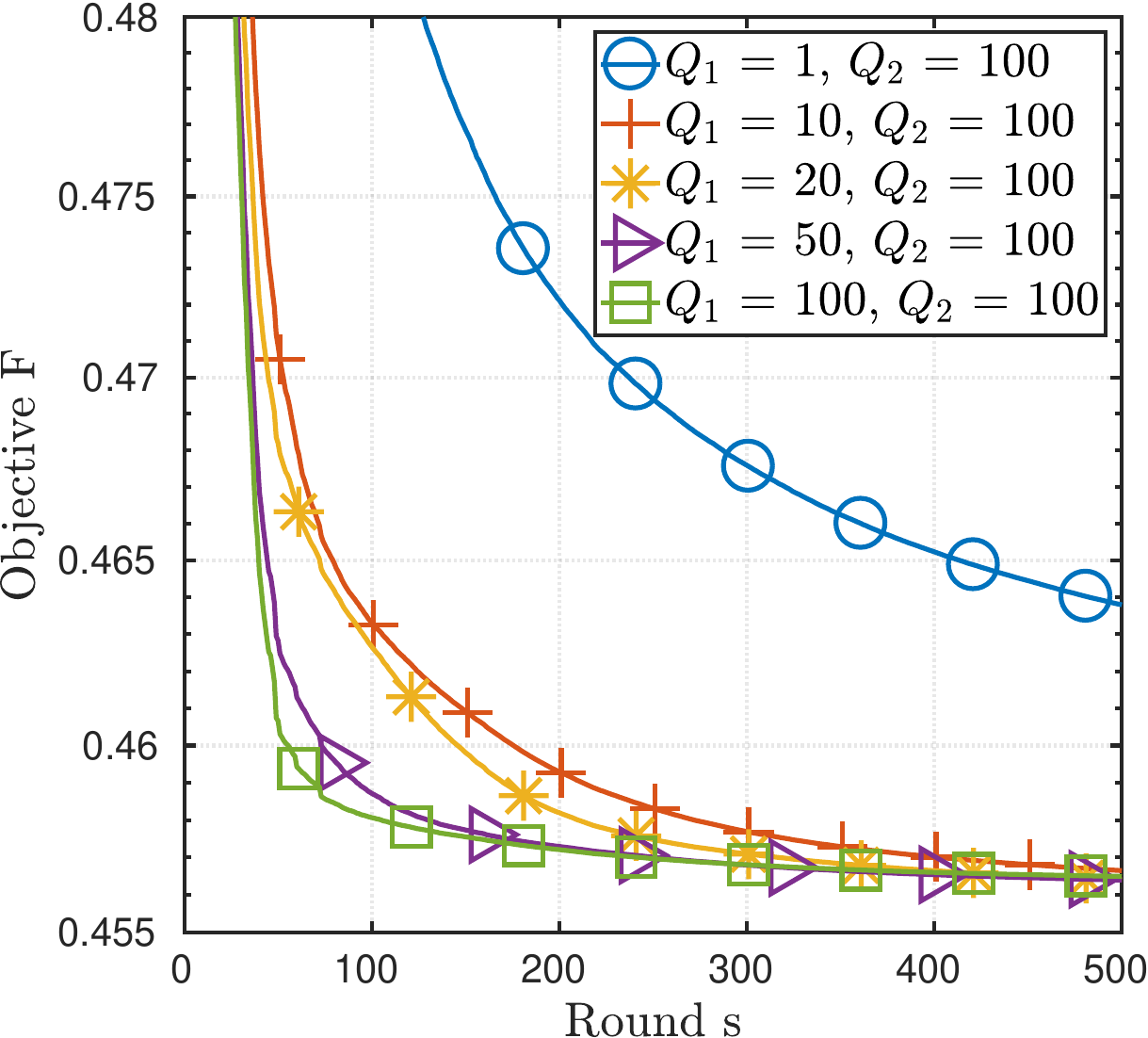}  \label{fig: gs1_con_curves_b}
	} \vspace{-0.1cm}
	\subfigure[\scriptsize  syn, \textbf{Case 2}, $m = 10$]{
		\includegraphics[width=6cm]{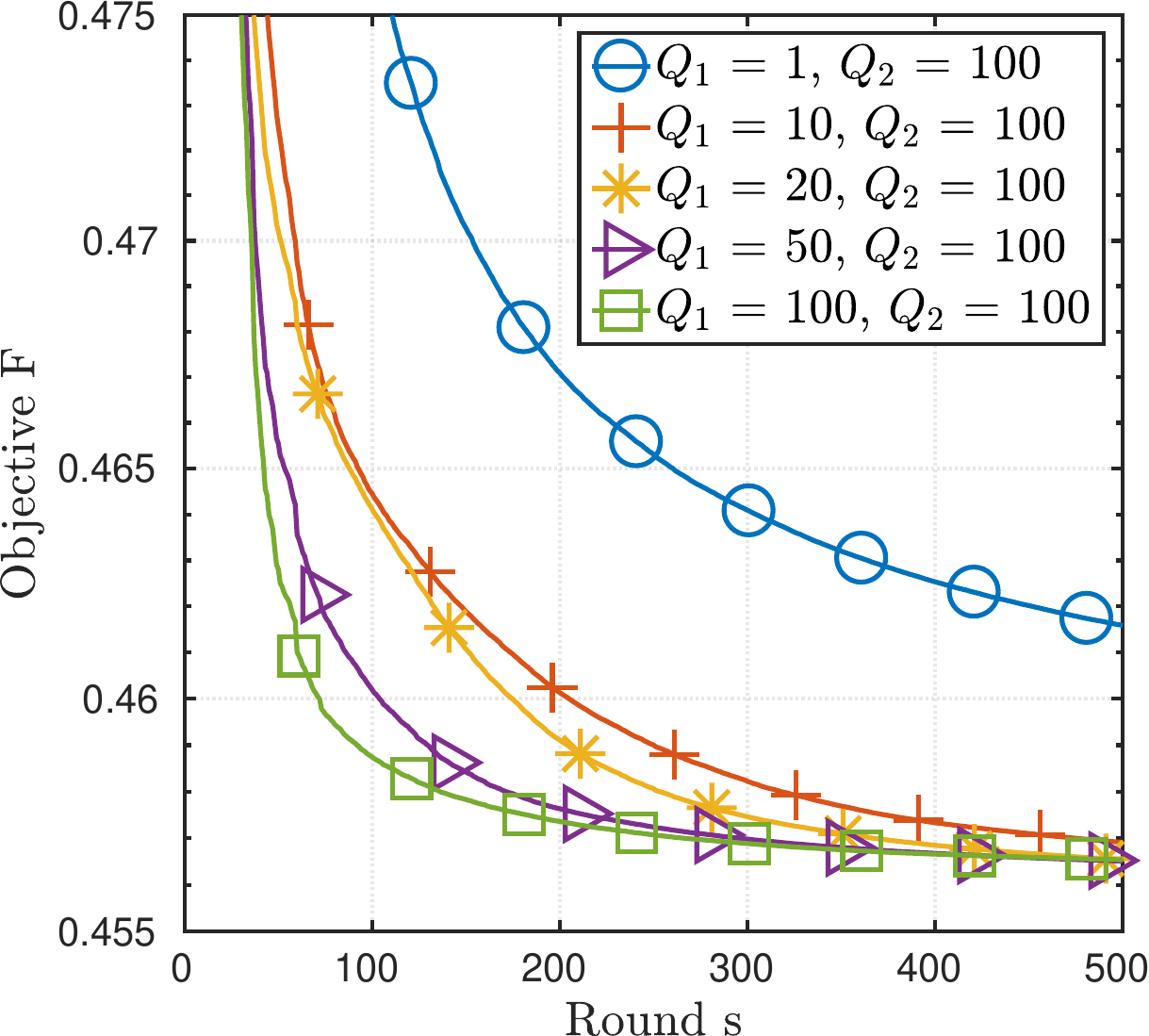} \label{fig: gs1_con_curves_c}
	} \vspace{-0.1cm}
	\subfigure[\scriptsize syn, \textbf{Case 1}, $m = 100$]{
		\includegraphics[width=6cm]{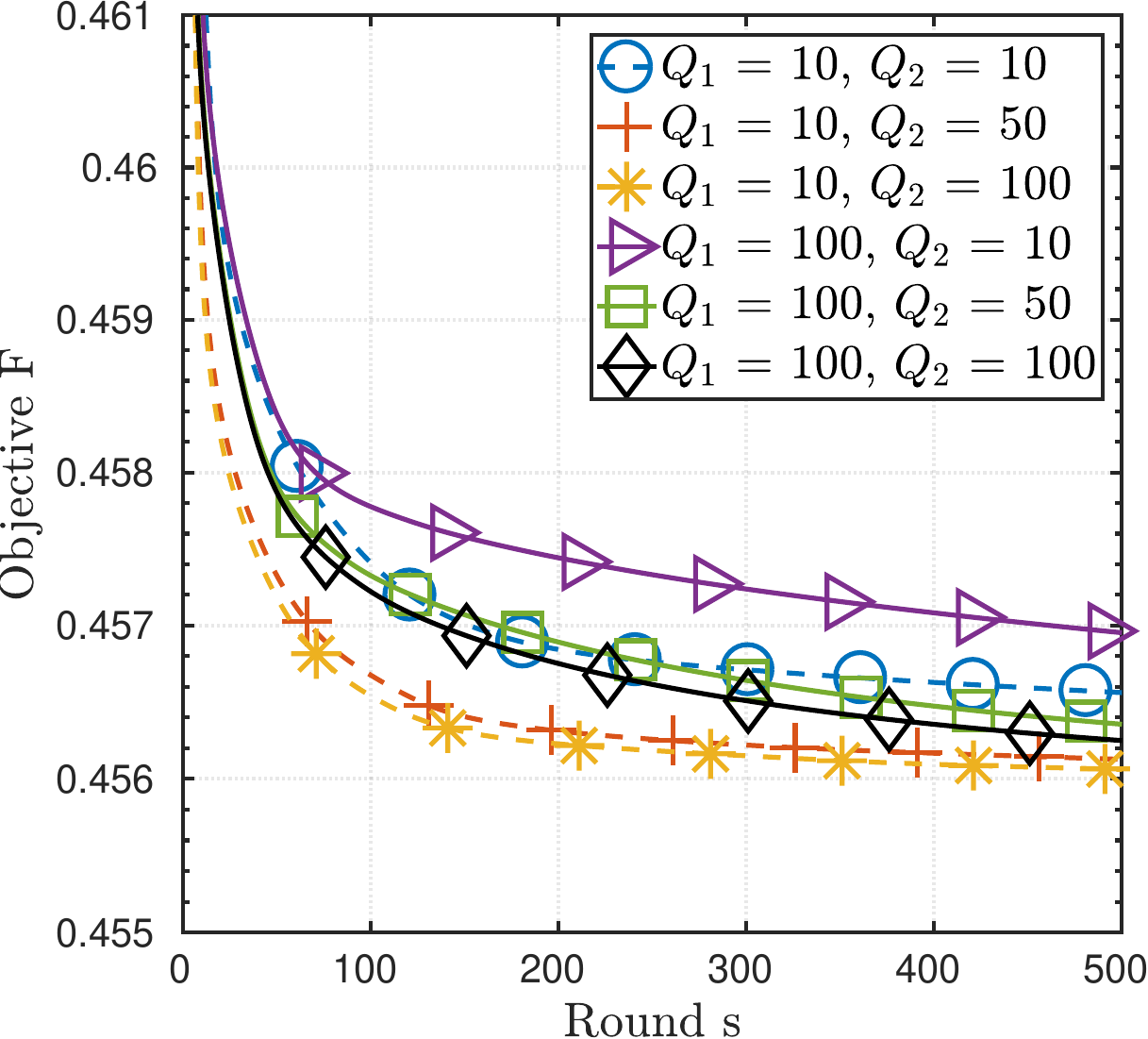}
		\label{fig: gs1_con_curves_d}
	} \vspace{-0.1cm}
	\subfigure[\scriptsize syn, \textbf{Case 1}, $m=10$]{
		\includegraphics[width=6cm]{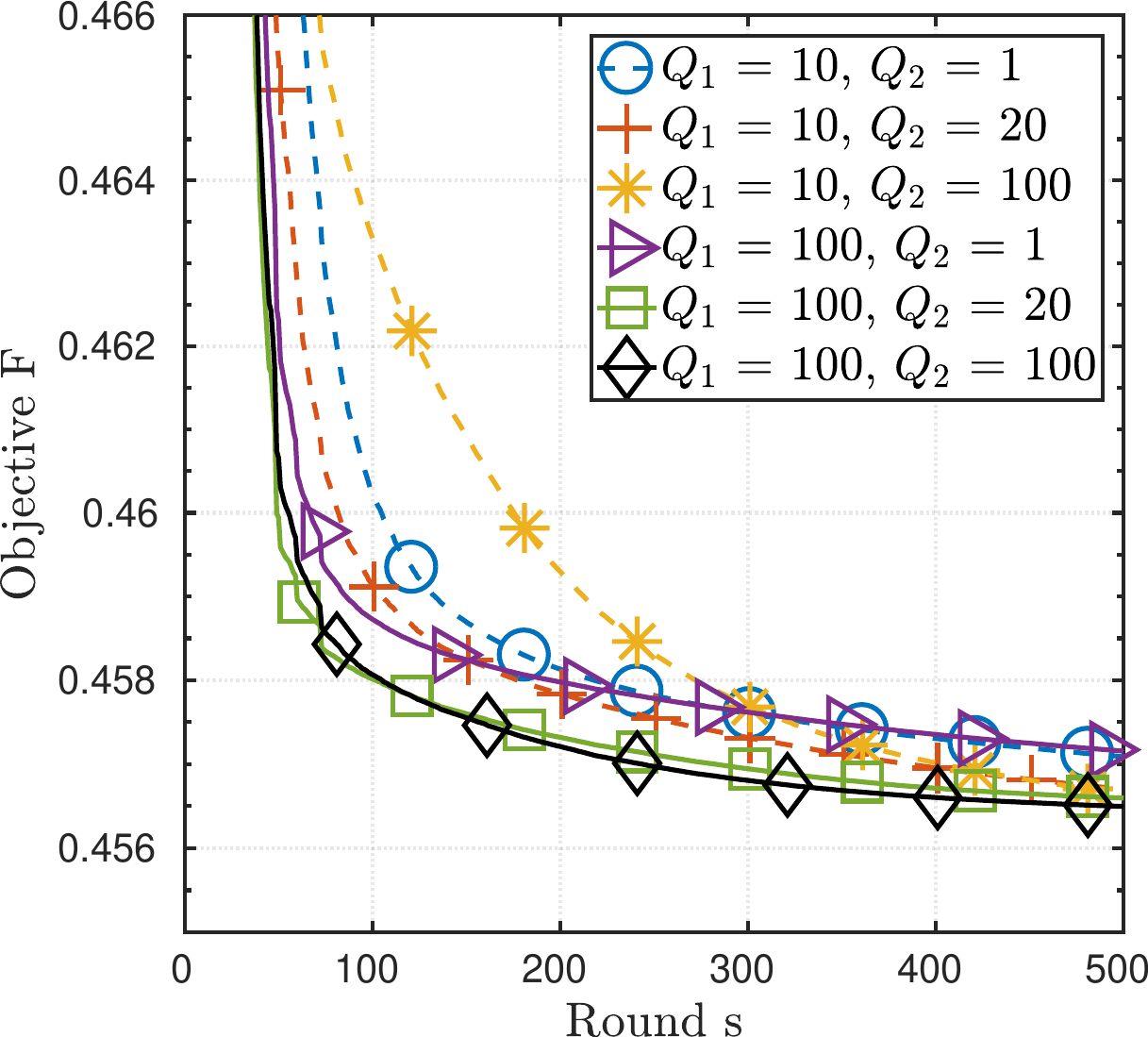} \label{fig: gs1_con_curves_e}
	} \vspace{-0.1cm}
	\subfigure[\scriptsize syn, \textbf{Case 2}, $m=10$]{
		\includegraphics[width=6cm]{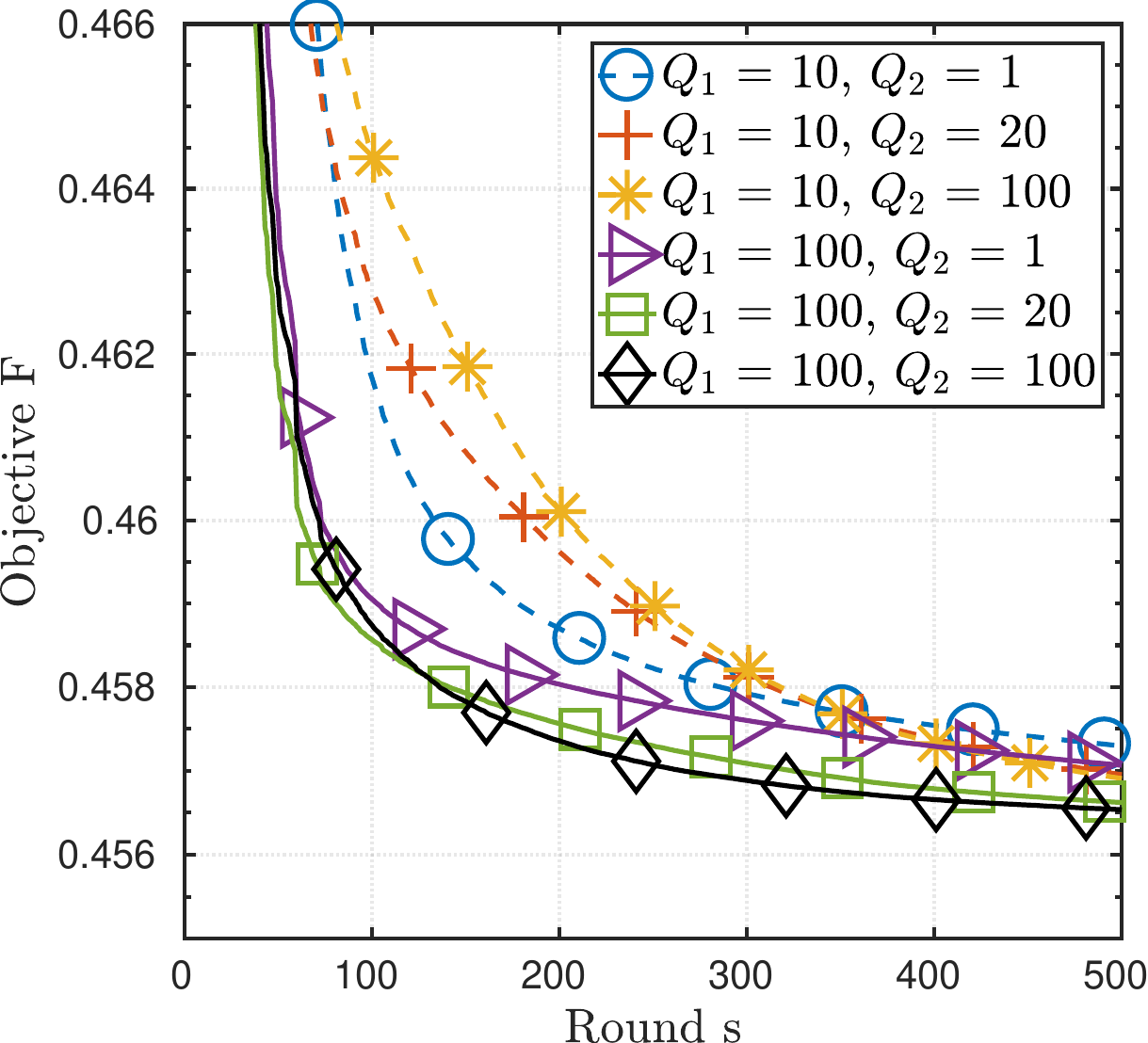} \label{fig: gs1_con_curves_f}
	} \vspace{-0.1cm}
	\caption{Convergence curve versus number of rounds of FedMGS for different values of $Q_1$ and $Q_2$.}\label{fig: gs1_con_curves}
	\vspace{-0.4cm}
\end{figure}

\vspace{-0.2cm}
\subsection{Convergence of FedMGS}
\label{subsec: conv of fedcgds}\vspace{-0.0cm}

In Fig. \ref{fig: gs1_con_curves}, the convergence curves of the FedMGS algorithm on the synthetic dataset are displayed. Fig. \ref{fig: gs1_con_curves_a} shows that increasing $Q_1$ can speed up the convergence, but the speedup with $Q_1>10$ is not as significant as that with $Q_1=10$. {One can also see that the method in \cite{DRMF_CF_2019} and \cite{SFMF_2019}, which corresponds to FedMGS with $Q_1 = Q_2 = 1$, converges much slower than FedMGS with $Q_1 > 1$ and $ Q_2 > 1$.}

As shown in Fig. \ref{fig: gs1_con_curves_b}, increasing $Q_1$ can monotonically improve the convergence rate when $m=10$; the same trend is also observed in Fig \ref{fig: gs1_con_curves_c} for Case 2 non-i.i.d. data. 
On the other hand, one can see from Fig. \ref{fig: gs1_con_curves_d} that FedMGS with $m=100$ (full client participation) and $Q_1=10$ or $100$ can have monotonically improved convergence speed when $Q_2$ increases. However, as shown in Fig. \ref{fig: gs1_con_curves_e}, when $m=10$, increasing $Q_2$ can improve the convergence only if $Q_1$ is also large ($Q_1 = 100$). We remark that similar insights apply to the Case 2 non-i.i.d data, which are shown in Fig \ref{fig: gs1_con_curves_f}. In summary, one can conclude that the algorithm convergence can benefit from a large $Q_2$ and small $Q_1$ when $m$ is large while from both large $Q_1$ and $Q_2$ when $m$ is small.

\vspace{-0.2cm}
\subsection{Effect of {i.i.d} and {non i.i.d} Data}
\label{subsec: effect of noniid}

\begin{figure}[t!]
	\centering
	\subfigure[\scriptsize FedMAvg, syn]{
		\includegraphics[width=6cm]{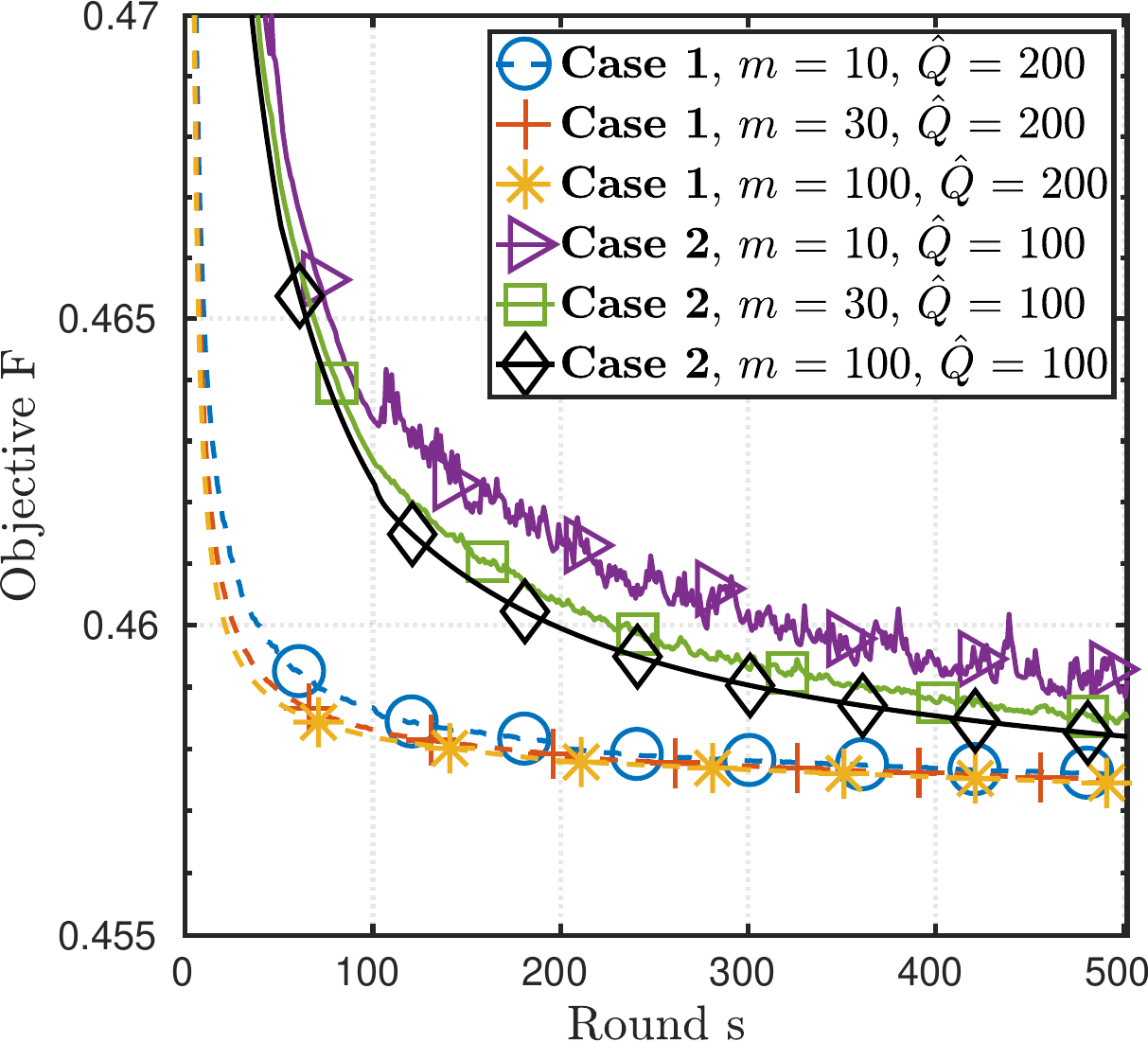}\label{fig: FedAM_snr3_iidvsnoniid}
	} \vspace{-0.2cm}
	\subfigure[\scriptsize FedMGS, syn]{
		\includegraphics[width=6cm]{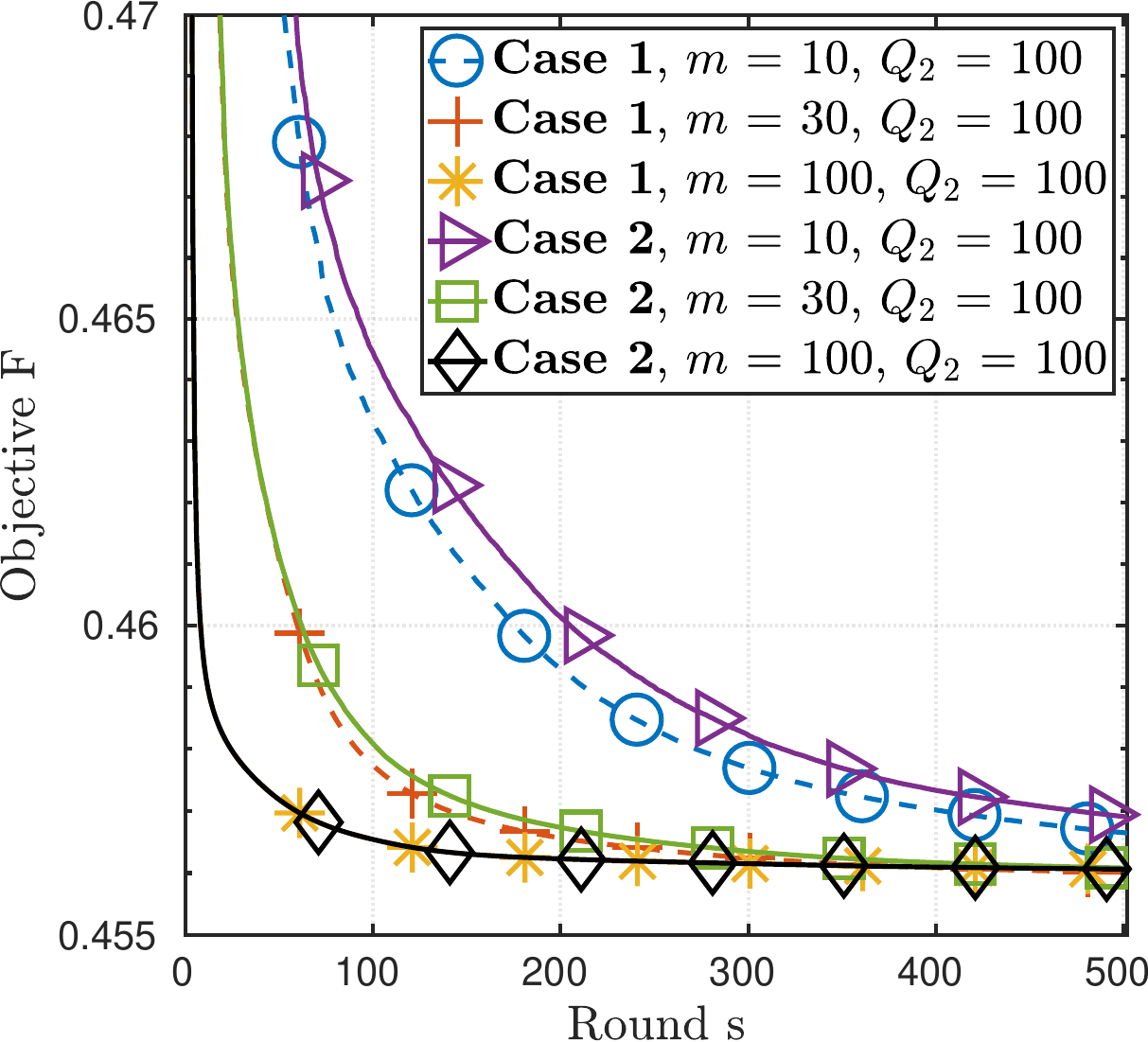}\label{fig: fedcgds_snr3_iidvsnoniid}
	} \vspace{-0.2cm}
	\subfigure[\scriptsize FedMAvg, TCGA]{
		\includegraphics[width=6cm]{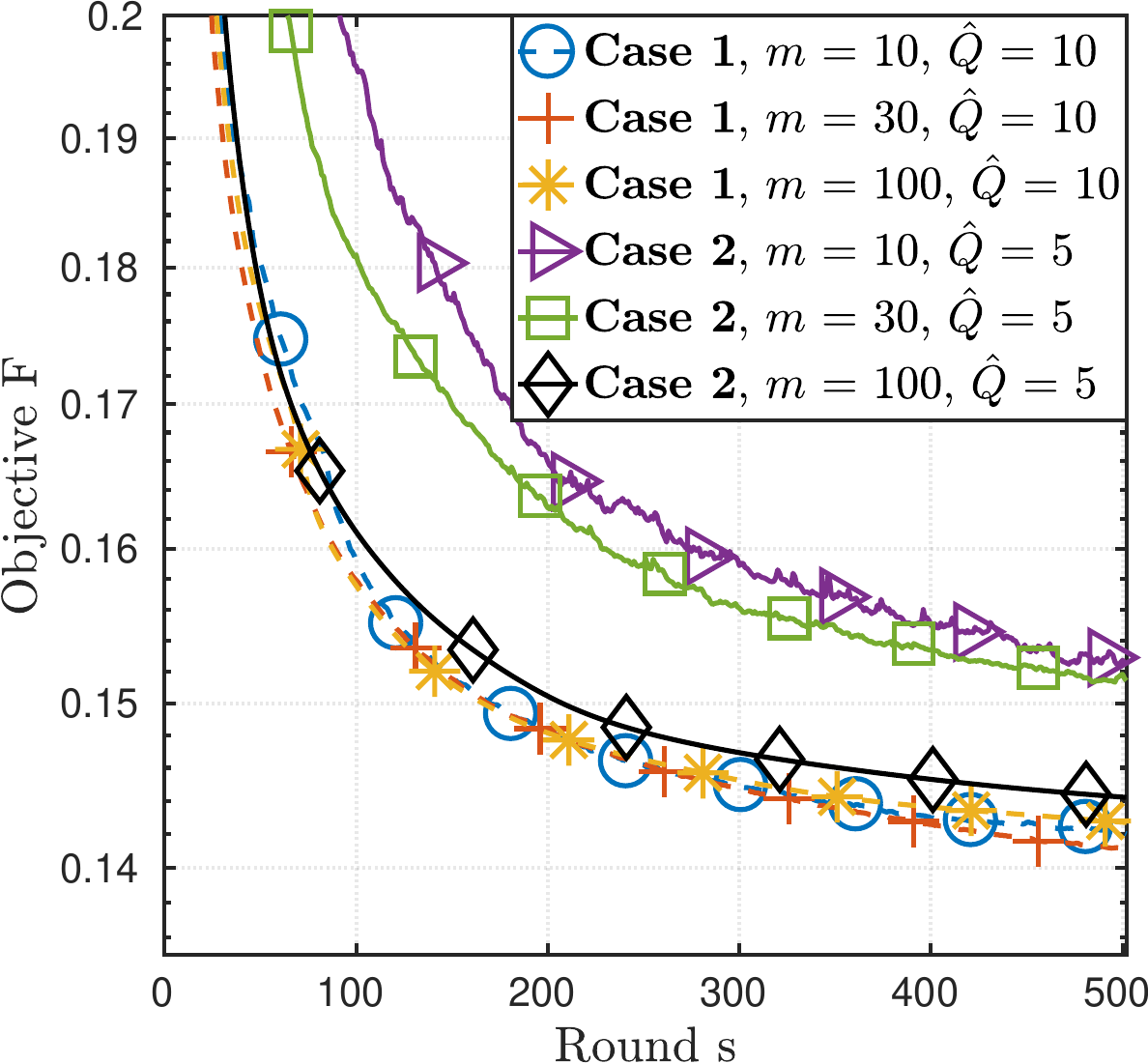} \label{fig: FedAM_TCGA_iidvsnoniid} 
	} \vspace{-0.0cm}
	\subfigure[\scriptsize FedMGS, TCGA]{
		\includegraphics[width=6cm]{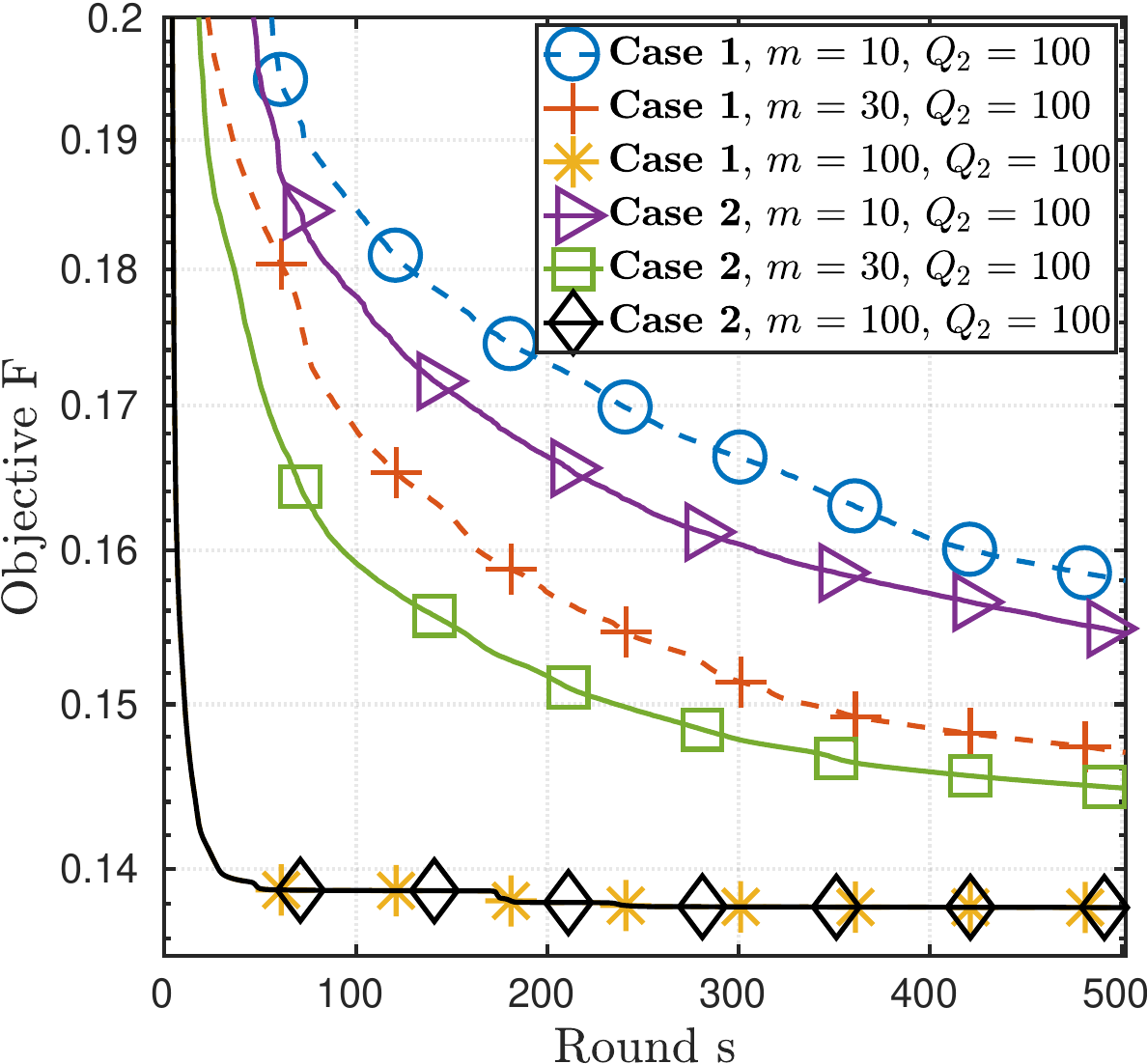} \label{fig: fedcgds_TCGA_iidvsnoniid} 
	} \vspace{-0.0cm}
	\caption{Convergence curve versus number of rounds FedMAvg and FedMGS on the four datasets. It is set that $Q_1 = 10$ for both FedMAvg and FedMGS for all datasets.
	}\label{fig:comparison_noniid}
	\vspace{-0.4cm}
\end{figure}

We further examine the performance of FedMAvg and FedMGS when faced with i.i.d (\textbf{Case 1}) and non-iid data (\textbf{Case 2}). 
One can see from Fig. \ref{fig: FedAM_snr3_iidvsnoniid} and Fig. \ref{fig: FedAM_TCGA_iidvsnoniid} that data distribution can have a considerable impact on the FedMAvg for both the synthetic and TCGA dataset (also see the results on TDT2 and MNIST datasets in \cite[Fig. S2]{FedC}) although full participation ($m=100$) may alleviate the degradation.

In Fig. \ref{fig: fedcgds_snr3_iidvsnoniid} and Fig. \ref{fig: fedcgds_TCGA_iidvsnoniid}, the results of FedMGS are shown, and it can be observed that FedMGS is more robust against the data distribution. In particular, when full participation ($m=100$), FedMGS can exhibit almost the same performance for both \textbf{Case 1} and \textbf{Case 2}. This is because when full participation, FedMGS is equivalent to applying the PALM to problem \eqref{eqn: distributed prob} over the network.

%

\vspace{-0.2cm}
\subsection{Comparison between FedMAvg and FedMGS}
\label{subsec: compariosn of two algs}

\begin{figure}[t!]
	\centering
	\subfigure[\scriptsize syn, \textbf{Case 2}, $F$ v.s. round $s$]{
		\includegraphics[width=6cm]{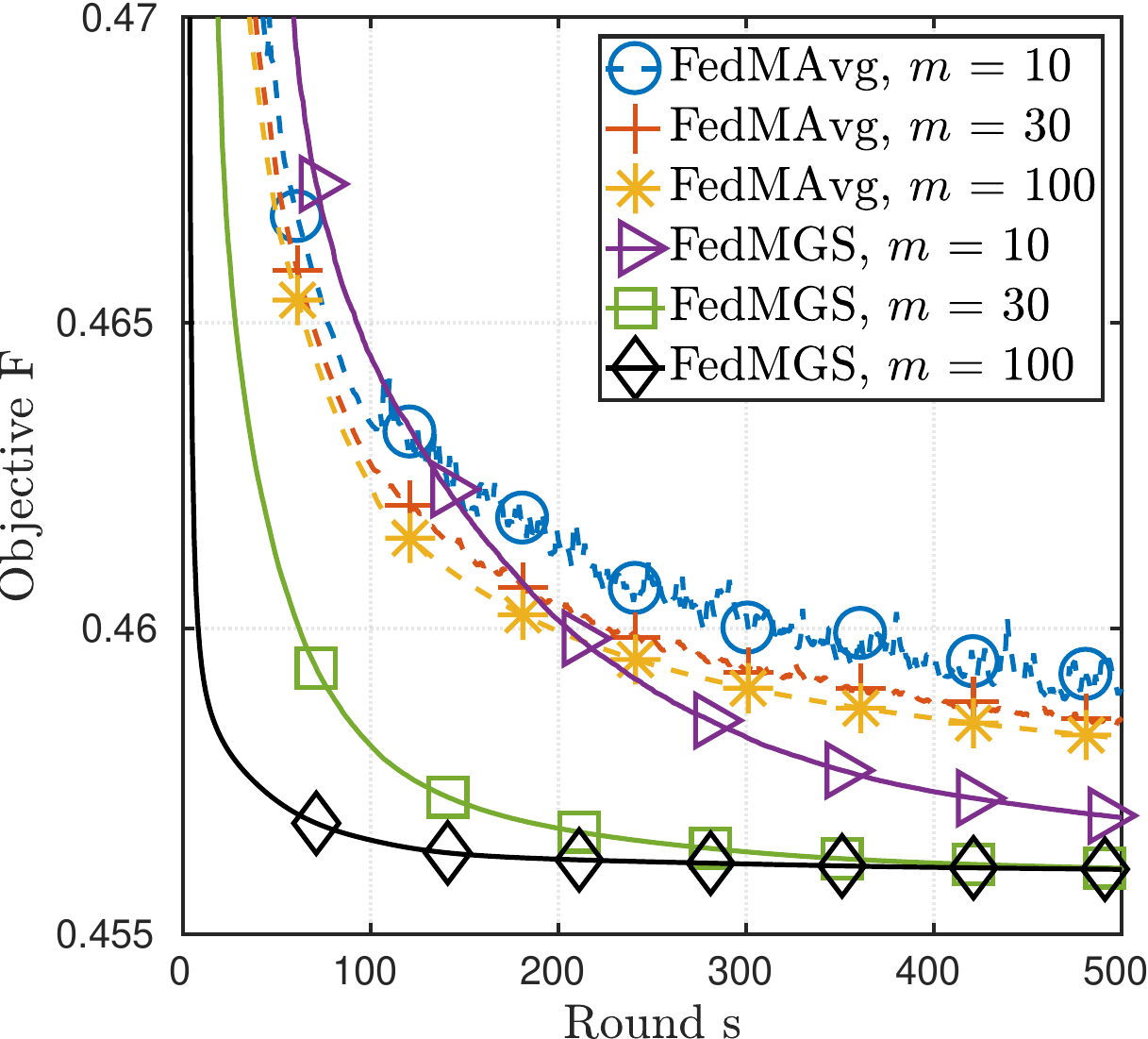}\label{fig:syn_comparison_a}
	} \vspace{-0.2cm}
	\subfigure[\scriptsize syn, \textbf{Case 2}, $F$ v.s. com. cost]{
		\includegraphics[width=6cm]{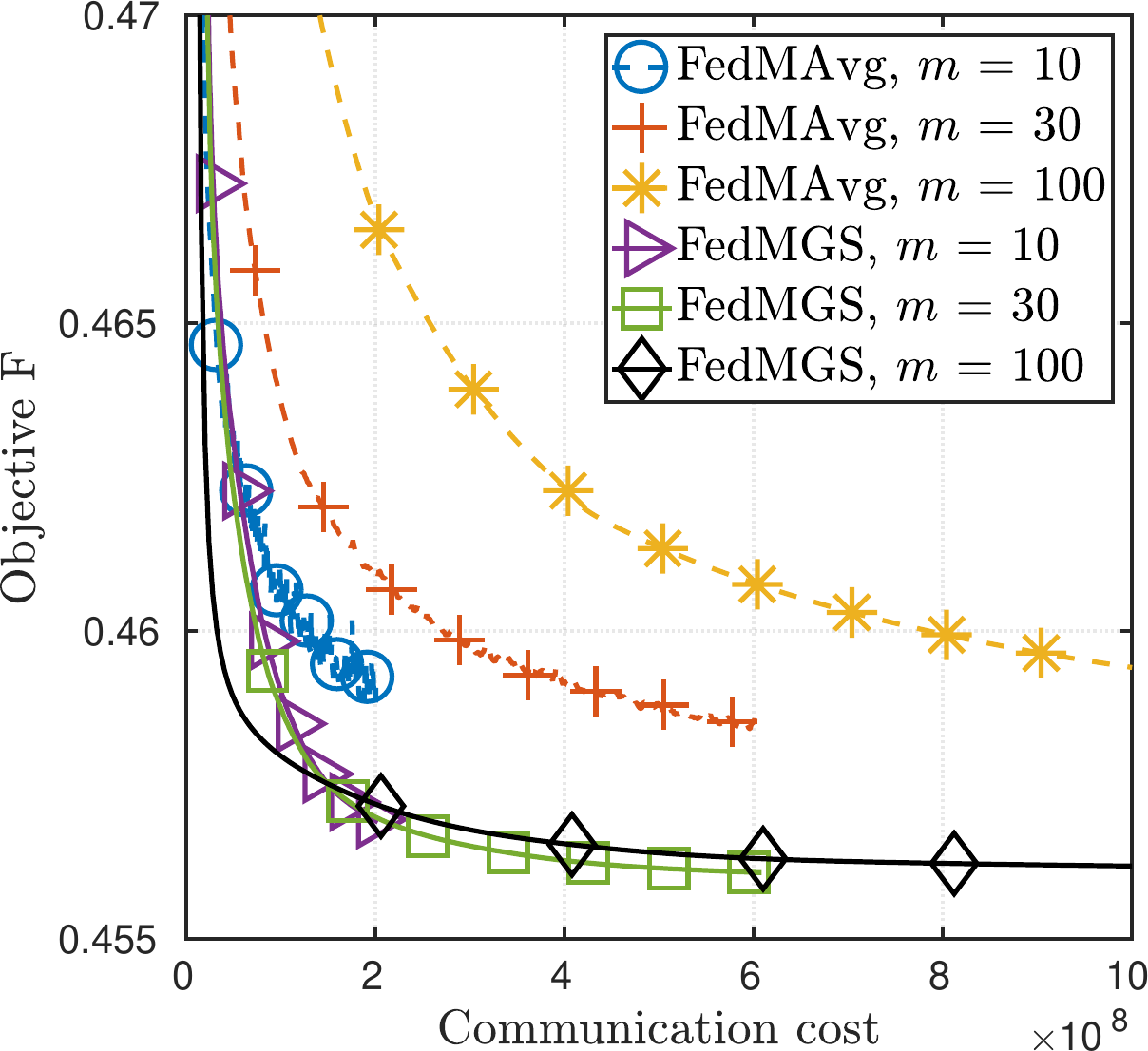}\label{fig:syn_comparison_b}
	} \vspace{-0.2cm}
	\subfigure[\scriptsize TCGA, \textbf{Case 2}, $F$ v.s. round $s$]{
		\includegraphics[width=6cm]{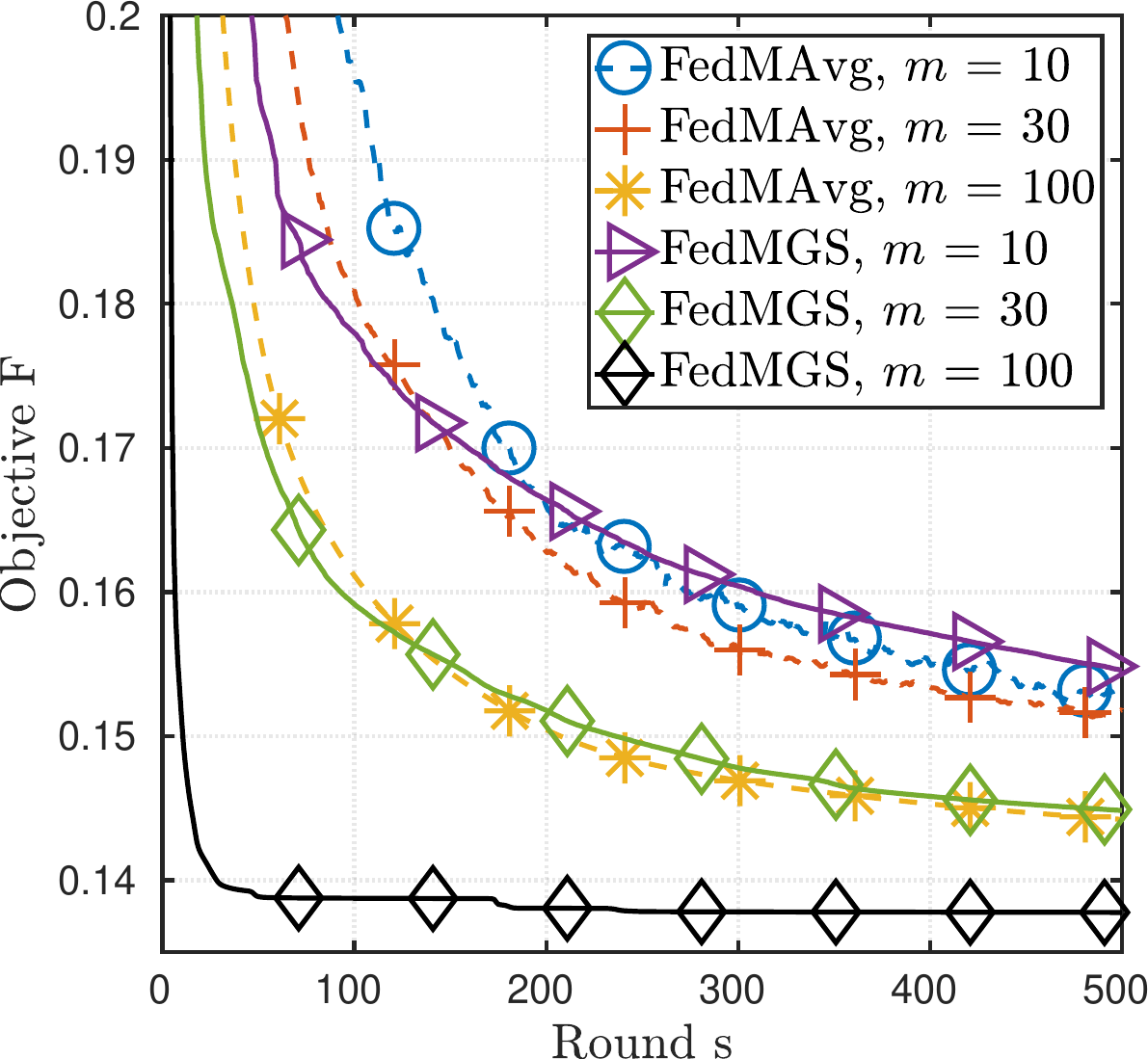} \label{fig:tcga_comparison_a} 
	} \vspace{-0.0cm}
	\subfigure[\scriptsize TCGA, \textbf{Case 2}, $F$ v.s. com. cost]{
		\includegraphics[width=6cm]{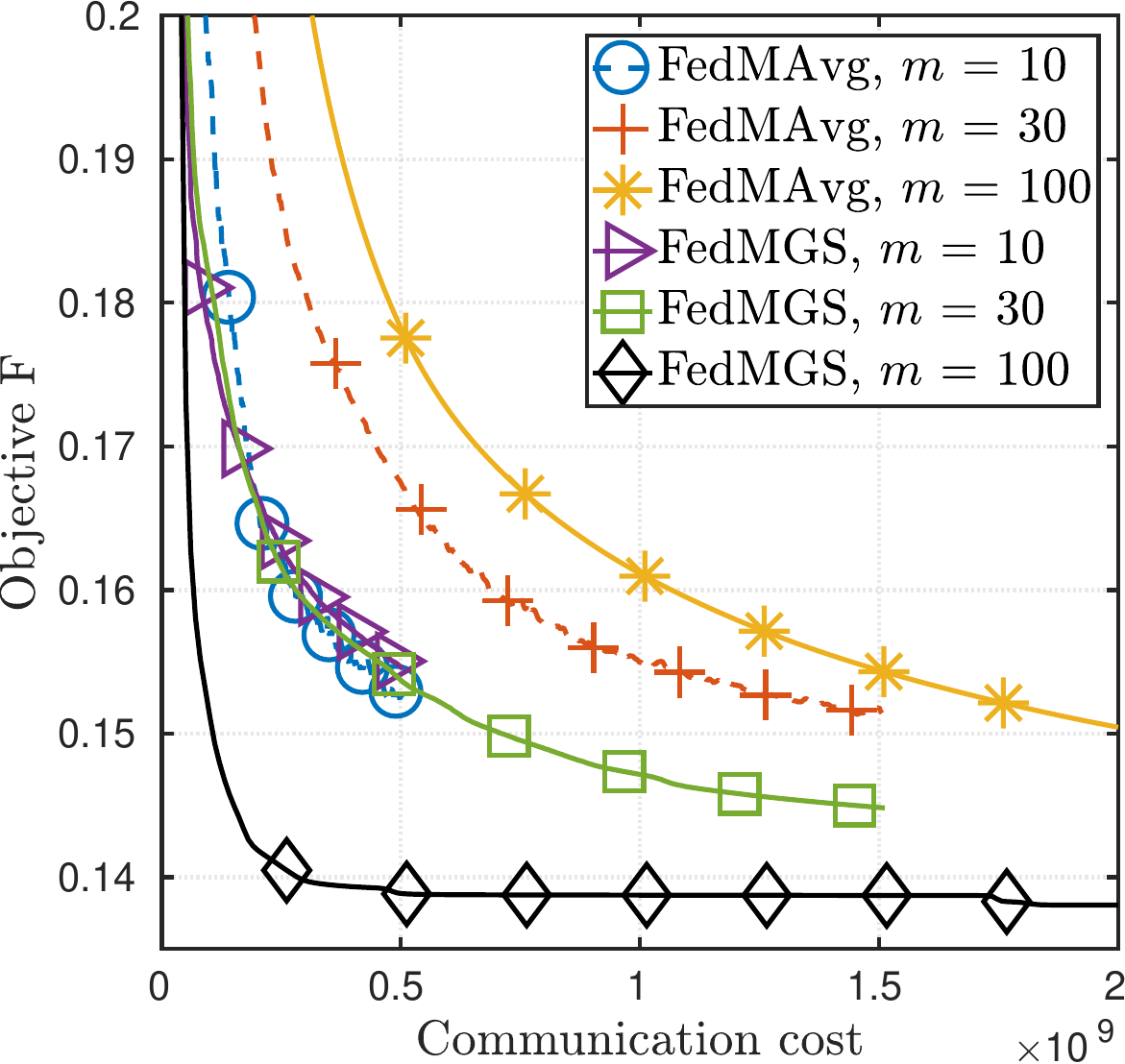} \label{fig:tcga_comparison_b} 
	} \vspace{-0.0cm}
	\caption{Convergence curve versus number of rounds and communication cost of FedMAvg and FedMGS under non-i.i.d data. }\label{fig:comparison0}
	\vspace{-0.3cm}
\end{figure}

In Fig. \ref{fig:comparison0}, we compare the FedMAvg and FedMGS on the non-i.i.d. synthetic and TCGA dataset (also see the results on TDT2 and MNIST datasets in \cite[Fig. S3]{FedC}). The setting of $Q_1$ and $Q_2$ $(\hat{Q})$ is the same as that in Section \ref{subsec: effect of noniid}. One can see from Fig. \ref{fig:syn_comparison_a} and Fig. \ref{fig:tcga_comparison_a} that on the synthetic data FedMGS performs significantly better than FedMAvg for almost all values of $m$ under test. One can also see that with increased $m$ both algorithms have improved convergence speed.

Fig. \ref{fig:syn_comparison_b} and Fig. \ref{fig:tcga_comparison_b} re-plot the same results but with respect to the communication 
cost. One can see that PCC are quite effective in reducing the communication cost for FedMAvg since a smaller value of $m$ can yield comparable performance as full participation ($m=100$) at the expense of significantly fewer communication costs.
The effect of PCP for FedMGS is not that significant because the convergence speed of FedMGS with $m=100$ is far faster than that with $m<100$ on these two dataset. However, one can see from \cite[Fig. S3]{FedC} that on the MNIST dataset, the FedMGS with $m=10$ can save the communication cost than that with $m=100$.

\begin{figure}[t!]
	\centering
	\subfigure[\scriptsize syn, \textbf{Case 2}]{
		\includegraphics[width=6cm]{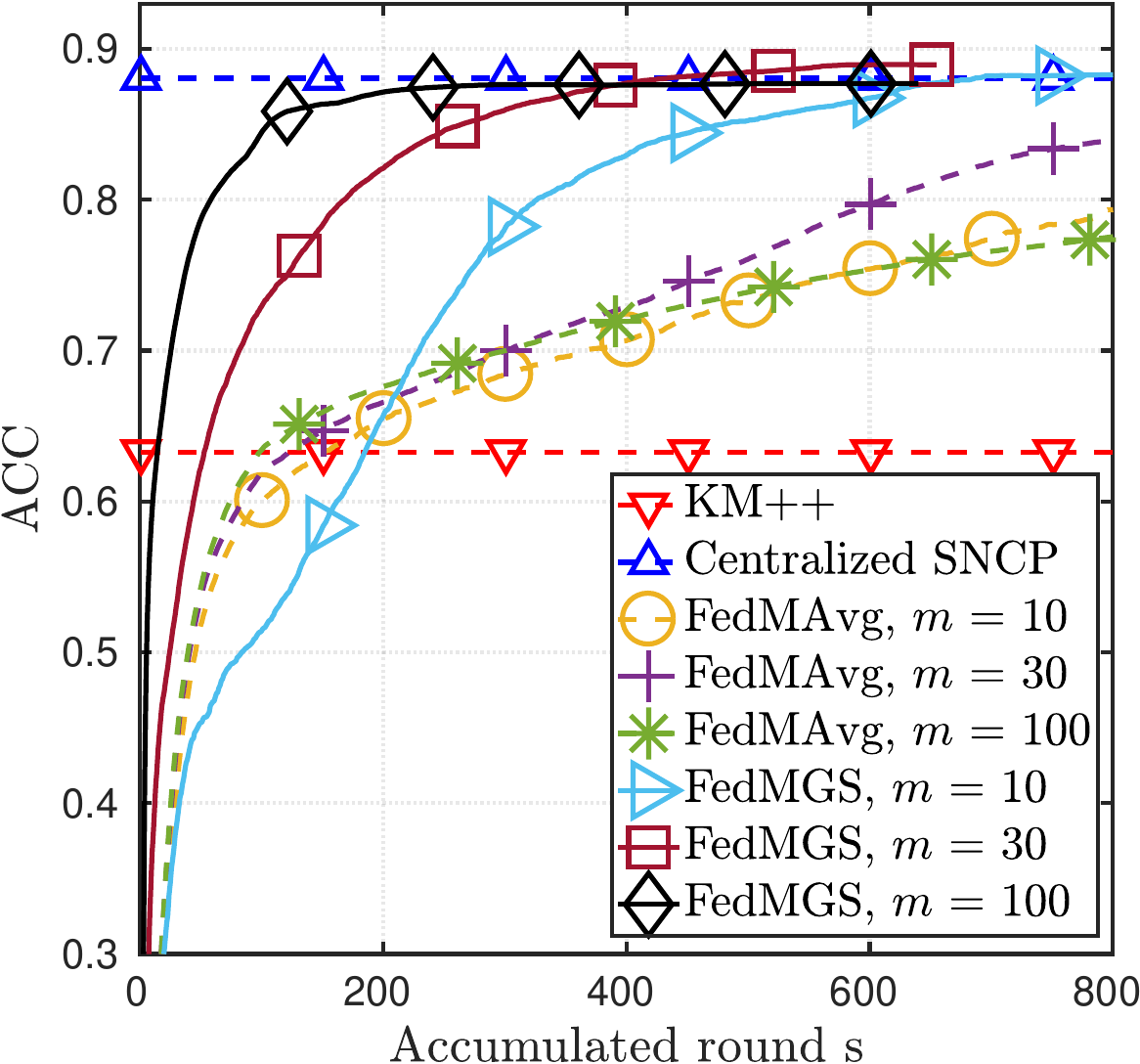}
	}\vspace{-0.1cm}
	\subfigure[\scriptsize syn, \textbf{Case 2}]{
		\includegraphics[width=6cm]{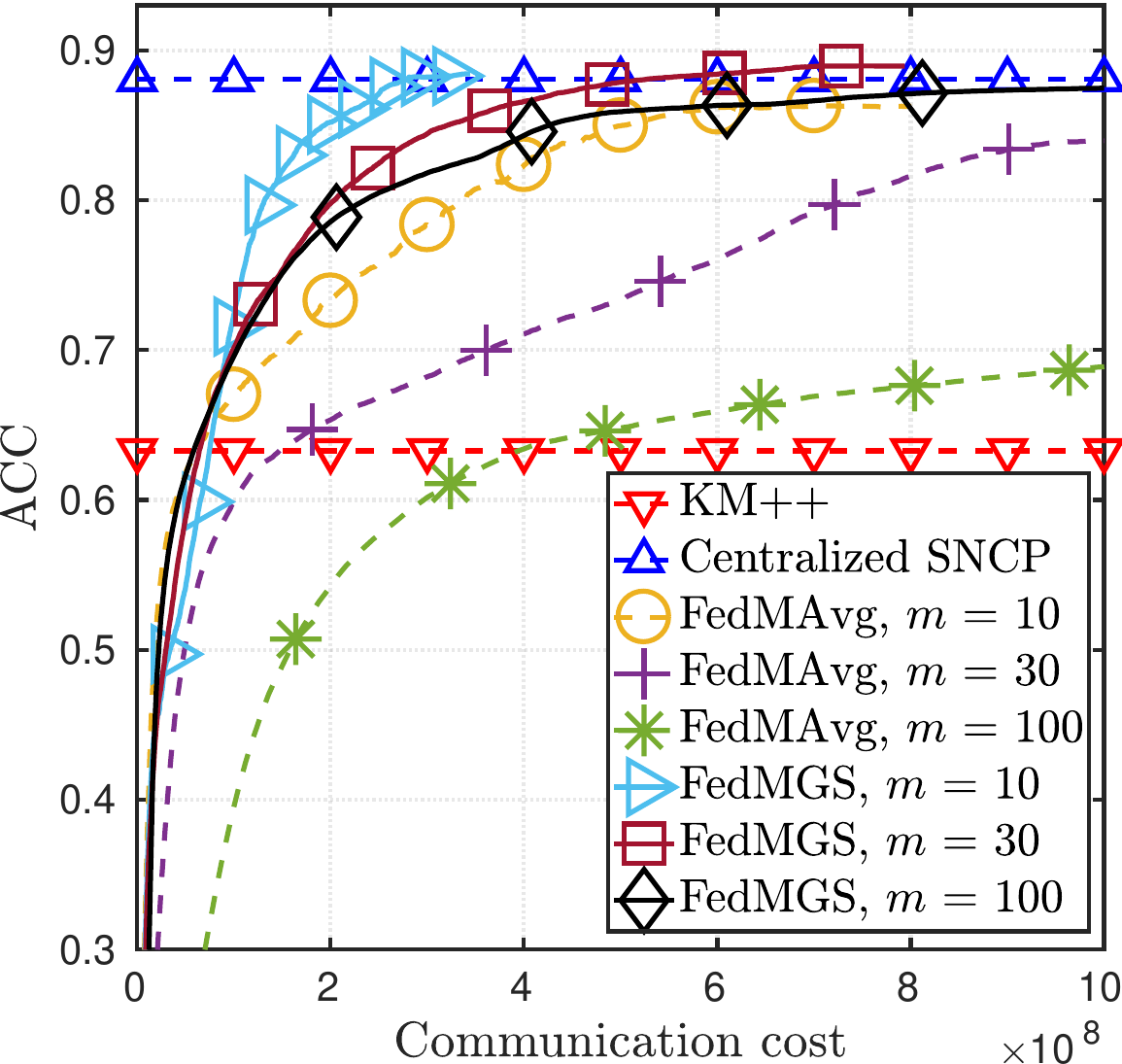}
	}\vspace{-0.1cm}
	\subfigure[\scriptsize TCGA, \textbf{Case 2}]{
		\includegraphics[width=6cm]{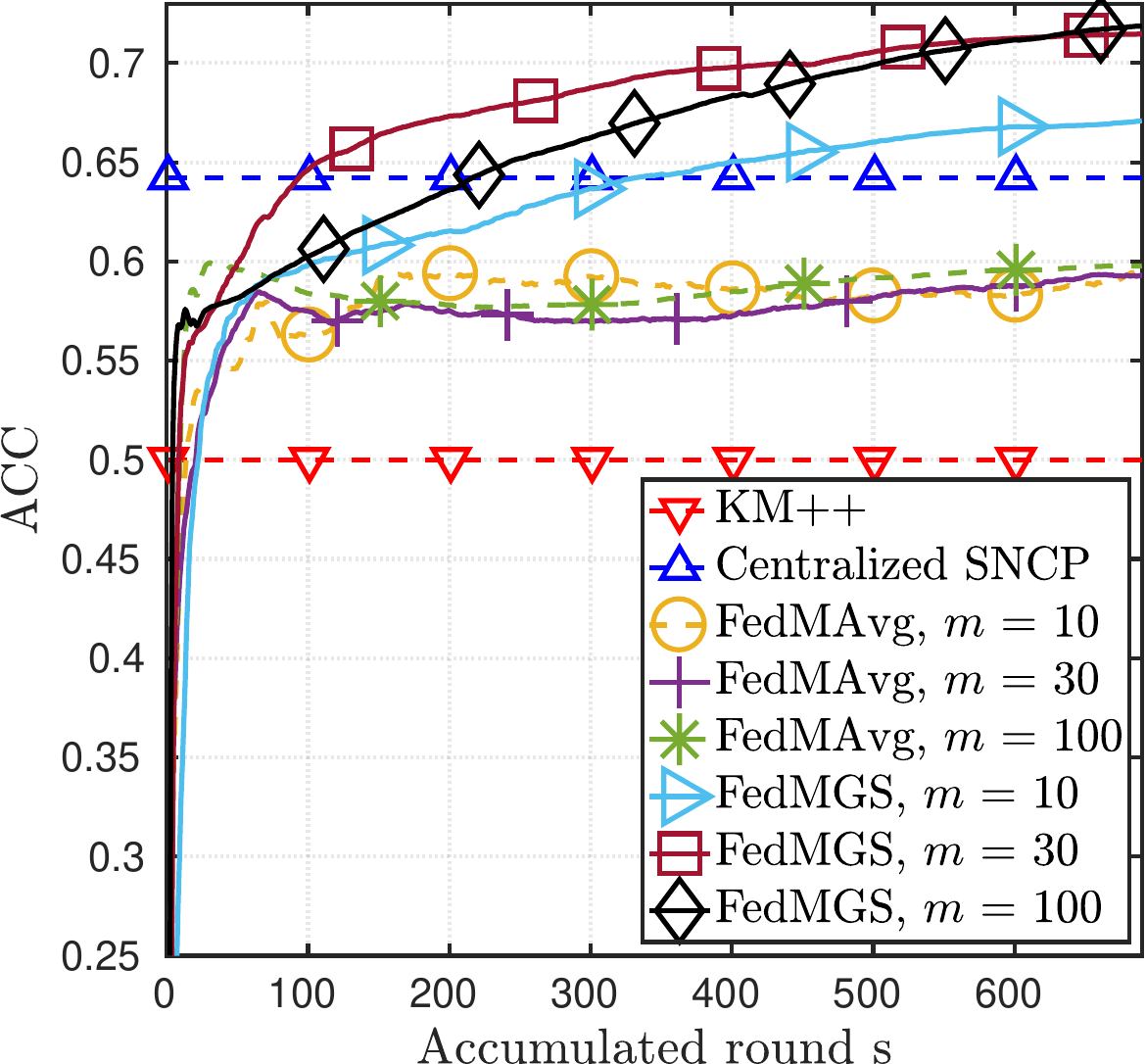}
	}\vspace{-0.1cm}
	\subfigure[\scriptsize TCGA, \textbf{Case 2}]{
		\includegraphics[width=6cm]{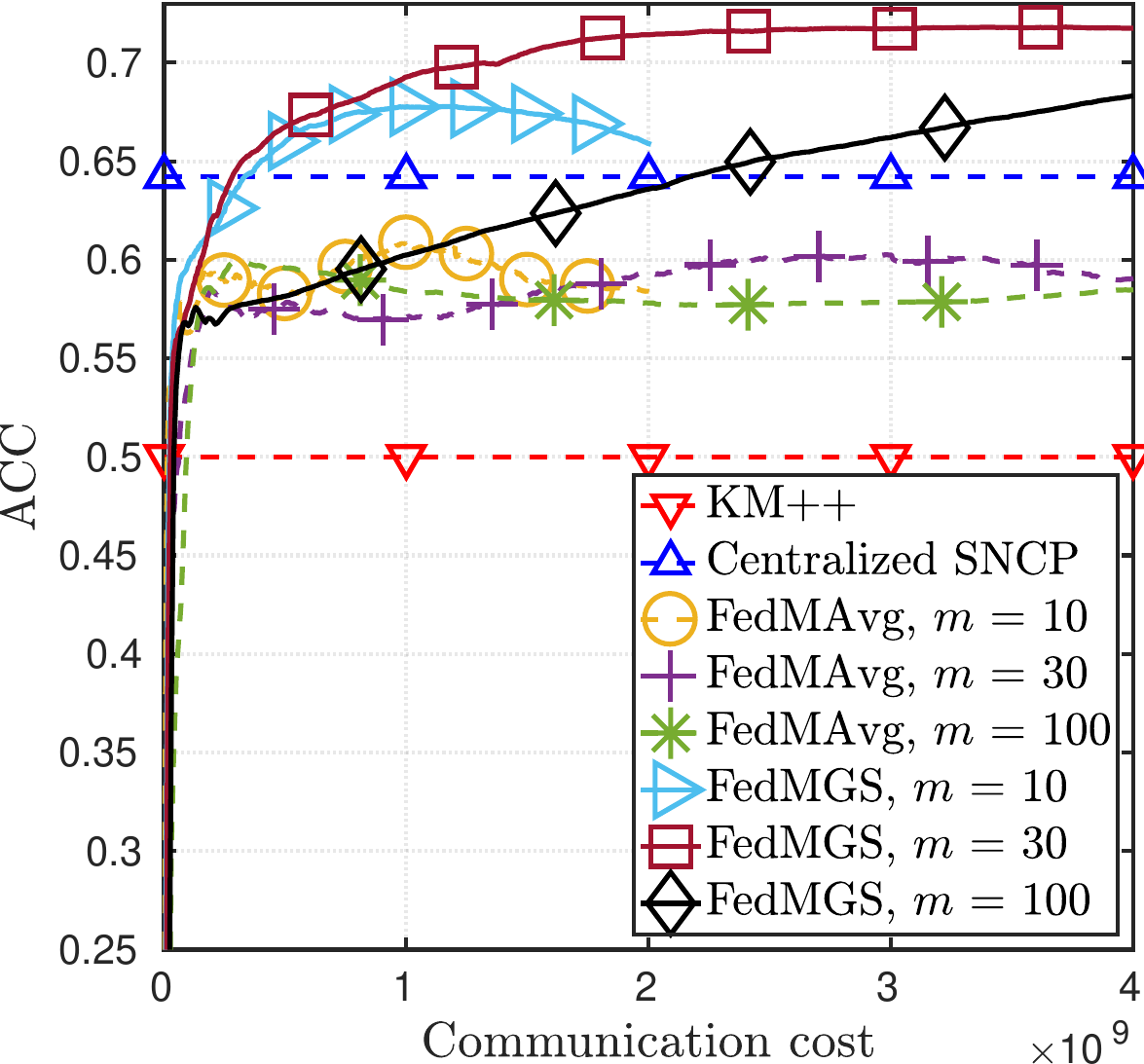}
	}\vspace{-0.1cm}
	\caption{Clustering accuracy versus number of accumulated rounds or communication cost of FedMAvg and FedMGS for different datasets.}\label{fig:accuracy}
	\vspace{-0.6cm}
\end{figure}

\vspace{-0.2cm}
\subsection{Clustering Performance}
\label{subsec: cluster performance}
\vspace{-0.0cm}
To evaluating the clustering performance of the proposed algorithms, we follow the successive non-convex penalty (SNCP) approach in \cite{Shuai_SNCPJ_2019} to gradually increase the penalty parameter $\rho$ in \eqref{sim: R(H)} whenever problem \eqref{eqn: distributed prob} is solved with sufficiently small $\varepsilon$. Specifically, the initial $\rho$ is set to $10^{-8}$ and is updated by $\rho = 1.5 \times \rho$ whenever $\varepsilon < 1 \times 10^{-5}$ (resp. $\varepsilon < 5 \times 10^{-5}$ ) for FedMAvg (resp. FedMGS). The stopping condition is set to $\varepsilon < 1 \times 10^{-8}$. The setting of $Q_1$ and $Q_2$ $(\hat{Q})$ is the same as that in Section \ref{subsec: effect of noniid}. In addition, the centralized SNCP method \cite[Algorithm \ref{alg: model_avg} \& 2]{Shuai_SNCPJ_2019} and the popular K-means++ \cite{K++_2007} are also implemented as two benchmarks.

Fig. \ref{fig:accuracy} presents the clustering accuracy (ACC) \cite{LCCF_2011} versus accumulated round number and communication cost on the synthetic and TCGA datasets (also see the results on TDT2 and MNIST datasets in \cite[Fig. S4]{FedC}). One can observe that FedMGS outperforms FedMAvg and achieves much higher clustering accuracy than K-means++. From Fig. \ref{fig:accuracy}(a)-(b), one can see that FedMGS yields comparable clustering accuracy as the centralized SNCP. Surprisingly, one can see from Fig. \ref{fig:accuracy}(c)-(d) that FedMGS can even perform better than the centralized SNCP on the TCGA data. More importantly, with $m = 10$ or $30$, both of the FedMGS and FedMAvg can quickly reach a higher clustering accuracy with lower communication cost than their counterparts with $m = 100$.

\vspace{-0.2cm}
\subsection{Comparison with Existing Distributed Clustering Methods}

\begin{table}[t!] 
	\centering 
	\caption{Clustering accuracy (\%) of the considered five methods on the four datasets (\textbf{Case 2}). 
	}\vspace{-0.2cm}
	\setlength{\tabcolsep}{3.0mm}
	\label{table: clustering quality}
	\begin{tabular}{|c|c|c|c|c|}
		\hline    \rowcolor{gray!50}       
		Dataset      &      syn      &      TDT2         &       TCGA        &  MNIST\\  \hline\hline 
		KM$||$     &     69.4  &      32.6       &     50.1  &   46.8  \\ \cline{1-5}
		BEL         &    38.7     &   33.3          &        42.0   &  46.1  \\ \cline{1-5}
		CAL        &     63.6      &       32.5     &         48.2   &      46.6  \\ \cline{1-5}
		\tabincell{c}{FedMAvg ($m = 10$)}       &     86.2          &      47.2     &     58.4  &  48.8   \\ \cline{1-5}
		\tabincell{c}{FedMAvg ($m = 100$)}       &     85.5            &    52.1     &   62.9    &    40.0   \\ \cline{1-5}
		\tabincell{c}{FedMGS ($m = 10$)}      & \textbf{88.4} &51.5  & 67.7 &\textbf{50.0}\\ \cline{1-5}
		\tabincell{c}{FedMGS ($m = 100$)}      & 87.7  & \textbf{53.5} & \textbf{72.2}  &49.1  \\ \cline{1-5}			     
		\hline
	\end{tabular}
	\vspace{-0.35cm}
\end{table}

We further examine the clustering performance of the proposed FedMAvg and FedMGS methods against three state-of-the-art distributed clustering methods, including KM$||$ \cite{Kmeans||_2012}, BEL \cite{DKCoreset_2013}, and CAL \cite{PA_DC_outlier_2018}. These methods require only few rounds of communications between the clients and the server. For example, both of BEL and CAL need merely one round of communication to obtain the clustering results. The distributed version of KM$||$ is similar to the parallel counterpart of the K-means++ which requires few communication rounds as well.

\begin{figure}[t!]
	\centering
	\subfigure[\scriptsize Ground truth]{
		\includegraphics[width=5cm]{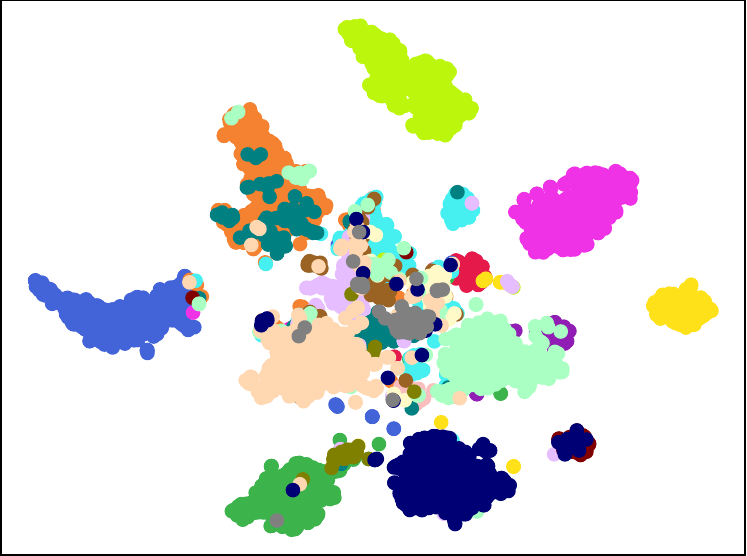}
	}\vspace{-0.05cm}
	\subfigure[\scriptsize KM$||$]{
		\includegraphics[width=5cm]{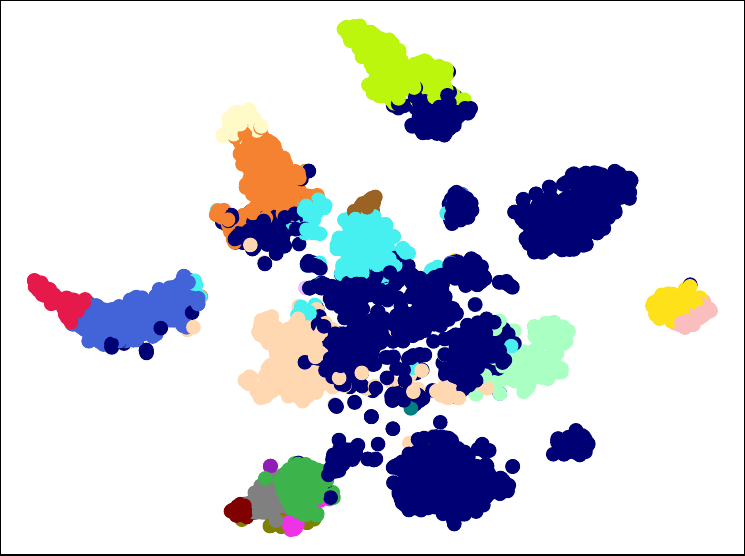}
	}\vspace{-0.05cm}
	\subfigure[\scriptsize BEL]{
		\includegraphics[width=5cm]{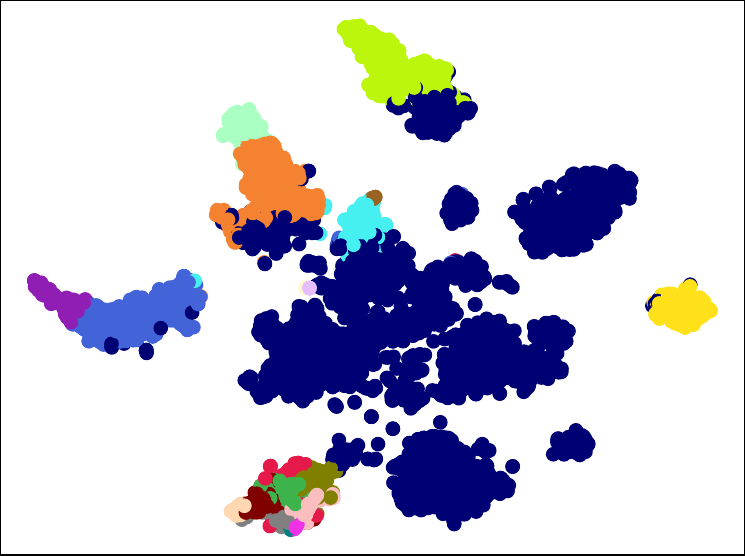}
	}\vspace{-0.05cm}
	\subfigure[\scriptsize CAL]{
		\includegraphics[width=5cm]{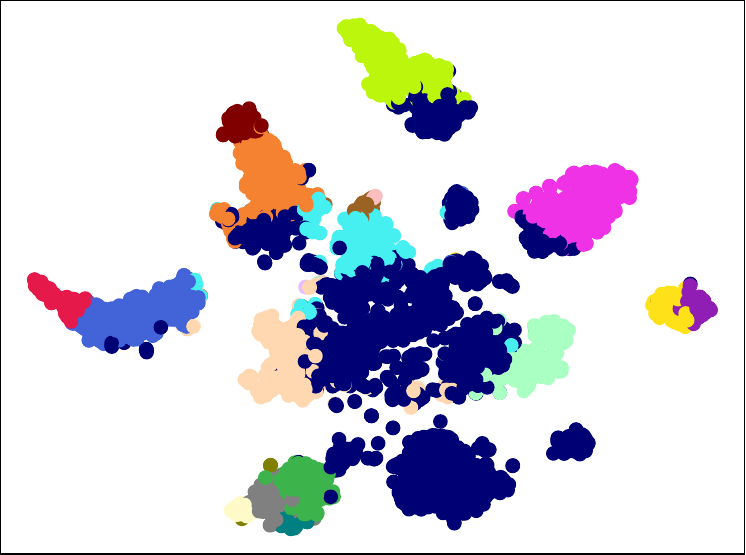}
	}\vspace{-0.05cm}
	\subfigure[\scriptsize FedMAvg ($m = 100$)]{
		\includegraphics[width=5cm]{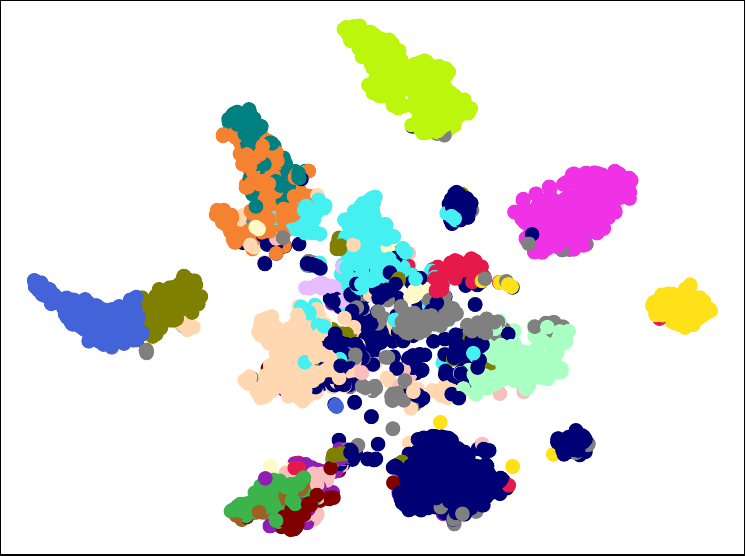}
	}\vspace{-0.05cm}
	\subfigure[\scriptsize FedMGS ($m = 100$)]{
		\includegraphics[width=5cm]{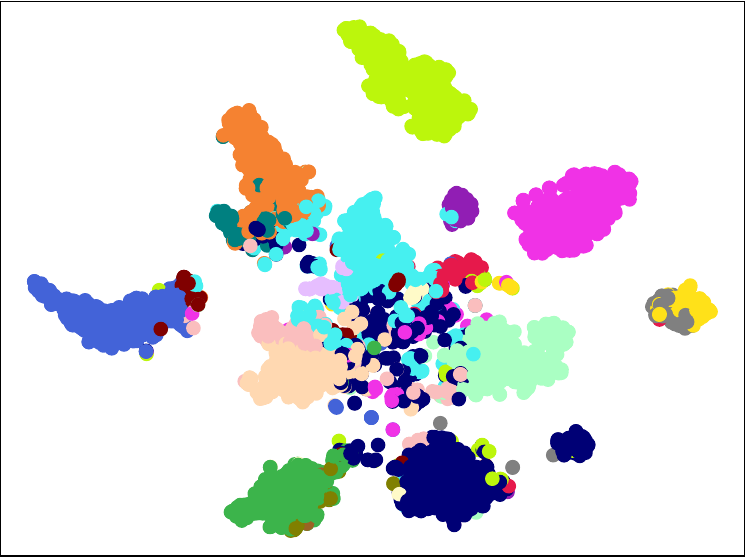}
	}\vspace{-0.05cm}
	\caption{Visualized clustering results (by t-SNE  \cite{TSNE_2008}) of the considered five methods on the TCGA dataset (Case 2).}\label{fig: cluster_tcga}
	\vspace{-0.5cm}
\end{figure}

Table \ref{table: clustering quality} shows the detailed clustering accuracy of all methods on the four datasets under \textbf{Case 2} over 100 clients. Note that the parameter settings of FedMAvg and FedMGS are the same as those in Section \ref{subsec: cluster performance}.
As seen in Table \ref{table: clustering quality}, the proposed FedMGS achieves the highest clustering accuracy. 
Besides, although FedMAvg provides worse performance than FedMGS, it still outperforms the other three distributed clustering methods considerably.   Fig. \ref{fig: cluster_tcga} presents the visualized clustering results (by t-SNE  \cite{TSNE_2008}) of all methods on the TCGA dataset under \textbf{Case 2}. One can see that both of the FedMAvg and FedMGS greatly outperform the other three distributed counterparts.

\vspace{-0.3cm}
\section{Conclusion}\label{sec: conclusion}

In this paper, we have presented FedMAvg and FedMGS algorithms for FedMF problems and considered their application to the fundamental data clustering task. 
We have also  provided theoretical convergence analyses for the two algorithms. The analysis results have explicitly characterized the impacts of non-i.i.d. data, partial client communication as well as local GD number on the convergence of of the two algorithms. They have also suggested a diminishing $Q_2$ strategy for FedMAvg to deal with the non-i.i.d. data, and also imply that FedMGS is less sensitive to data distribution.
Extensive experiment results have demonstrated consistent convergence behaviors of the proposed algorithms on both synthetic and real datasets, showing insights on the values of $Q_1$ and $Q_2$ $(\hat Q)$ that can improve the convergence speed of both algorithms. It has also been shown that PCC/PCP can significantly reduce the communication cost of both algorithms. 

As the future works, it is worthwhile to devise FedMF algorithms for general MF models that can handle outlier and noisy data \cite{DMC_RF_2015} as well as considering other MF applications, such as item recommendation and biological data analysis.  Enhancing privacy and security \cite{SFMF_2019} of FedMF algorithms is also of great importance.

\appendices
\section{Proof of Theorem \ref{thm: model_avg}}\label{proof of fedmavg}

According to \eqref{eqn: virtual_global}, we define

\vspace{-0.2cm}
{\begin{align} \label{smeqn1}
	\wt\Wb^{s, t} = \Pc_{\Wc}(\ol\Wb^{s, t}),~\ol\Wb^{s, t}  = \frac{1}{m}\sum\limits_{p \in \Ac^s}  \Wb_p^{s, t},
	\end{align}} 
for all $s$ and $t = \{0\}\cup \Qc_1 \cup \Qc_2^s$. 
Then, by \eqref{eqn: FedAM update of W1}, we have
{\begin{align}\label{smeqn2_2}
	\wt\Wb^{s, t} =\ol\Wb^{s, 0} &= \Wb^s = \Wb_p^{s,0},~\forall t \in \{0\} \cup \Qc_1, p \in \Pc. 
	\end{align}}
We also define 
{\begin{align}
	\Ec^{s-1} = & \{\Ac^1, \ldots, \Ac^{s - 1}, \{\Hb^{1,t}\}_{t=0}^{Q^{1}},\ldots,\{\Hb^{s-1,t}\}_{t=0}^{Q^{s-1}}, \notag \\
	&~\{\Wb^{1,t}\}_{t=0}^{Q^1},\ldots,\{\Wb^{s-1,t}\}_{t=0}^{Q^{s-1}}, \notag \\ &~\{c_p^1\}_{p=1}^P,\ldots,\{c_p^{s-1}\}_{p=1}^P,d^1,\ldots,d^{s-1}\} \label{eqn: def_ec}
	\end{align}}
\vspace{-0.3cm}

\noindent as the collection of historical events up to round $(s-1)$.
\vspace{0.1cm}

\noindent \underline{\bf Objective Descent w.r.t. $\Hb$:}
According to \cite[Lemma 3.2]{PALM_2014}, \eqref{eqn: FedAM update of H1}, \eqref{eqn: FedAM update of W1} and \eqref{smeqn2_2} imply
{\begin{align} 
	&F_p(\wt\Wb^{s,t}, \Hb_p^{s, t}) - F_p(\wt\Wb^{s,t-1}, \Hb_p^{s, t-1}) \notag \\ 
	\leq& -\frac{\gamma_1 - 1}{2}\ol L_H \|\Hb_p^{s, t-1} - \Hb_p^{s, t}\|_F^2, \forall t \in \Qc_1, \label{thm1: descent_H}
	\end{align}}since $c_p^s= \frac{\gamma_1 \ol L_H}{2}$.
Summing up \eqref{thm1: descent_H} from $t = 1$ to $Q_1$ and taking expectation  of it conditional on $\Ec^{s-1}$ yields

\vspace{-0.3cm}
{\begin{align}
	&\E[F_p(\wt\Wb^{s,Q_1}, \Hb_p^{s, Q_1})|\Ec^{s-1}] - \E[F_p(\wt\Wb^{s,0}, \Hb_p^{s, 0})|\Ec^{s-1}] \notag \\ 
	\leq& -\frac{\gamma_1 - 1}{2} \sum_{t = 1}^{Q_1} \ol L_H\E[\|\Hb_p^{s, t-1} - \Hb_p^{s, t}\|_F^2|\Ec^{s-1}].
	\label{thm1: obj_descent_Hp}
	\end{align}}
\vspace{-0.3cm}

As a result, the objective function $F$ descends with local updates of $\Hb$ as follows

\vspace{-0.3cm}
{\begin{align}
	&\E[F(\wt\Wb^{s,Q_1}, \Hb^{s, Q_1})|\Ec^{s-1}] - \E[F(\wt\Wb^{s,0}, \Hb^{s, 0})|\Ec^{s-1}] \notag  \\
	\leq &-\frac{\gamma_1 - 1}{2} \sum_{t = 1}^{Q_1}\sum_{p = 1}^{P} \omega_p \ol L_H \E[\|\Hb_p^{s, t - 1} - \Hb_p^{s, t}\|_F^2|\Ec^{s-1}]. \label{thm1: descent_H2}
	\end{align}}
\vspace{-0.3cm}

\noindent \underline{\bf Objective Descent w.r.t. $\Wb$:}
Note by (13) that $\Hb_p^{s, t} = \Hb_p^{s, t-1}, \forall t \in \Qc_2^s$.
Since $\nabla_{W} F(\cdot, \Hb^{s,Q})$ is Lipschitz continuous under Assumption \ref{eqn: assumption2},
by the descent lemma \cite[Lemma 3.1]{PALM_2014}, we have

\vspace{-0.4cm}
{
	\begin{align}
	&F(\wt\Wb^{s, t},\Hb^{s, t}) \leq F(\wt\Wb^{s, t-1}, \Hb^{s, t-1}) + \frac{L_{W}^s}{2} \|\wt\Wb^{s, t} - \wt\Wb^{s, t-1}\|_F^2 \notag \\
	&~~~~~~~~~~~~~+ \underbrace{\langle \nabla_{W} F(\wt\Wb^{s, t-1}, \Hb^{s, t-1}), \wt\Wb^{s, t} - \wt\Wb^{s, t-1} \rangle}_{\triangleq \rm (a)} .\label{thm1: lip_W} 
	\end{align}}
\vspace{-0.3cm}

The term ${\rm (a)}$ can be bounded by the following lemma which is proved in \cite[Section 1.1]{FedC}.
\begin{Lemma}
	For any $s$ and $t \in \Qc_2^s$, the following inequality holds.
	
	\vspace{-0.4cm}
	{ 
		\begin{align}
		&\langle \nabla_{W} F(\wt\Wb^{s, t-1}, \Hb^{s, t-1}), \wt\Wb^{s, t} - \wt\Wb^{s, t-1}\rangle \notag \\
		\leq& - d^s \|\wt\Wb^{s, t} - \wt\Wb^{s, t-1}\|_F^2 \notag \\
		&+ \bigg \langle \nabla_{W} F(\wt\Wb^{s, t-1}, \Hb^{s, t-1}) -\frac{1}{m}\sum_{p \in \Ac^s}\nabla_{W}F_p(\Wb_p^{s, t-1}, \Hb_p^{s, t-1}), \wt\Wb^{s, t} - \wt\Wb^{s, t-1} \bigg\rangle. \label{thm1: descent_immediate}
		\end{align}}
	\vspace{-0.3cm}
	
\end{Lemma} 

Thus, substituting \eqref{thm1: descent_immediate} into \eqref{thm1: lip_W} gives rise to

\vspace{-0.3cm}
{
	\begin{align}
	&~~~~F(\wt\Wb^{s, t}, \Hb^{s, t}) \notag \\
	& \leq F(\wt\Wb^{s, t-1}, \Hb^{s, t-1}) -\bigg (d^s - \frac{L_{W}^s}{2}\bigg)\|\wt\Wb^{s, t} - \wt\Wb^{s, t-1}\|_F^2 \notag 
	\\
	&~~+\bigg \langle \nabla_{W} F(\wt\Wb^{s, t-1}, \Hb^{s, t-1}) - \frac{1}{m}\sum_{p \in \Ac^s} \nabla_{W}F_p(\Wb_p^{s, t-1}, \Hb_p^{s, t-1}), \wt\Wb^{s, t} - \wt\Wb^{s, t-1} \bigg\rangle \notag \\
	&\leq F(\wt\Wb^{s, t-1}, \Hb^{s, t-1}) - \frac{d^s - L_{W}^s}{2}\|\wt\Wb^{s, t} - \wt\Wb^{s, t-1}\|_F^2 \notag \\ 
	&~~+\frac{1}{2d^s}\bigg\| \nabla_{W} F(\wt\Wb^{s, t-1}, \Hb^{s, t-1})\! -\! \frac{1}{m}\sum_{p \in \Ac^s}\!\!\nabla_{W}\!F_p(\Wb_p^{s, t-1}, \Hb_p^{s, t-1})\bigg\|_F^2,
	\label{thm1: descent_Ws1}
	\end{align}}
\vspace{-0.3cm}

\noindent where \eqref{thm1: descent_Ws1} holds since $\langle \xb, \yb \rangle \leq \frac{1}{2c}\|\xb\|_2^2 + \frac{c}{2}\|\yb\|_2^2, \forall c > 0$.
Then, taking the expectation over the two sides of \eqref{thm1: descent_Ws1} conditioned on $\Ec^{s-1}$ yields

\vspace{-0.3cm}
{
	\begin{align}
	&\E[F(\wt\Wb^{s, t}, \Hb^{s, t})|\Ec^{s-1}] - \E[F(\wt\Wb^{s, t-1}, \Hb^{s, t-1})|\Ec^{s-1}]\notag \\
	\leq &  - \E\bigg[\frac{d^s - L_{W}^s}{2}\|\wt\Wb^{s, t} - \wt\Wb^{s, t-1}\|_F^2\bigg|\Ec^{s-1}\bigg] \notag \\ 
	&~~+   \frac{1}{2d^s}\E\bigg[\bigg\| \nabla_{W} F(\wt\Wb^{s, t-1}, \Hb^{s, t-1}) - \frac{1}{m}\sum_{p \in \Ac^s} \nabla_{W}F_p(\Wb_p^{s, t-1}, \Hb_p^{s, t-1})\bigg\|_F^2\bigg|\Ec^{s-1}\bigg]. \label{thm1: descent_W}
	\end{align}}
\vspace{-0.3cm}

We then proceed with the following lemma which is proved in \cite[Section 1.2]{FedC}.
\vspace{-0.3cm}
\begin{Lemma} \label{lemm1}
	For any $s$ and $t \in \Qc_2^s$, 
	
	\vspace{-0.3cm}
	{\begin{align}
		&\E\bigg[\bigg\| \nabla_{W} F(\wt\Wb^{s, t-1}, \Hb^{s, t-1}) \notag \\
		&~~~~- \frac{1}{m}\sum_{p \in \Ac^s} \nabla_{W}F_p(\Wb_p^{s, t-1}, \Hb_p^{s, t-1})\bigg\|_F^2\bigg|\Ec^{s-1}\bigg] \notag \\
		\leq & \bigg(2 + \frac{16}{m}\bigg) \sum_{p = 1}^{P}\omega_p(L_{W_p}^s)^2\E[\| \wt\Wb^{s, t-1} -  \Wb_p^{s, t-1}\|_F^2|\Ec^{s-1}] +  \frac{16}{m}\zeta^2. \notag 
		\end{align}}
	\vspace{-0.5cm}
	
\end{Lemma}
By applying Lemma \ref{lemm1}, we have 

\vspace{-0.3cm}
{\begin{align}
	&\E[F(\wt\Wb^{s, t}, \Hb^{s, t})|\Ec^{s-1}] - \E[F(\wt\Wb^{s, t-1}, \Hb^{s, t-1})|\Ec^{s-1}] \notag \\
	\leq &  - \E\bigg[\frac{d^s - L_{W}^s}{2}\|\wt\Wb^{s, t} - \wt\Wb^{s, t-1}\|_F^2\bigg|\Ec^{s-1}\bigg] \notag \\ 
	&~~+   \bigg(\frac{1 + 8/m}{d^s}\bigg) \sum_{p = 1}^{P}\omega_p(L_{W_p}^s)^2\E[\| \wt\Wb^{s, t-1} -  \Wb_p^{s, t-1}\|_F^2|\Ec^{s-1}] + \frac{8\zeta^2}{md^s}, ~\forall t \in \Qc_2^s. \label{thm1: descent_W1}
	\end{align}}
\vspace{-0.3cm}

\noindent Summing \eqref{thm1: descent_W1} up from $t = Q_1 +1$ to $Q^s$ yields

\vspace{-0.3cm}
{\small 
	\begin{align}
	&~~\E[F(\wt\Wb^{s, Q^s}, \Hb^{s, Q^s})|\Ec^{s-1}]-\E[F(\wt\Wb^{s, Q_1}, \Hb^{s, Q_1})|\Ec^{s-1}] \notag \\
	&\leq   - \bigg(\frac{d^s - L_{W}^s}{2}\bigg) \E\bigg[\sum_{t = Q_1 + 1}^{Q^s}\|\wt\Wb^{s, t} - \wt\Wb^{s, t-1}\|_F^2\bigg|\Ec^{s-1}\bigg]  + \frac{8Q_2^s\zeta^2}{md^s} \notag \\ 
	&~~ +  \frac{1 + 8/m}{d^s} \underbrace{ \sum_{t = Q_1 +1}^{Q^s}\sum_{p = 1}^{P}\omega_p(L_{W_p}^s)^2\E[\| \wt\Wb^{s, t-1} -  \Wb_p^{s, t-1}\|_F^2|\Ec^{s-1}]}_{\rm \triangleq (b)}.  \label{thm1: descent_W2}
	\end{align}}
\vspace{-0.3cm}

\noindent The term $\rm (b)$ can be bounded with the following lemma which is proved in \cite[Section 1.3]{FedC}.

\vspace{-0.3cm}
\begin{Lemma}
	Let $\gamma_2 \geq Q_2^s\sqrt{2(7 + 4 \ol L_W^2/\underline{L}_W^2)}$. It holds that
	
	\vspace{-0.3cm}
	{\begin{align}
		&\sum_{t = Q_1 +1}^{Q^s}\sum_{p = 1}^{P}\omega_p(L_{W_p}^s)^2\E[\| \wt\Wb^{s, t-1} -  \Wb_p^{s, t-1}\|_F^2|\Ec^{s-1}] \notag \\
		\leq & \frac{2C_1^s(\frac{11\zeta^2}{3} + \phi^2)}{\gamma_2^2},
		\end{align}} 
	\vspace{-0.3cm}	
	
	\noindent where $C_1^s \triangleq Q_2^s(Q_2^s - 1)(2Q_2^s -1)$.
\end{Lemma}	
Since $\gamma_2  = \max\{Q_2^1\sqrt{2(7 + 4 \ol L_W^2/\underline{L}_W^2)}, \sqrt{T}\}$ and $Q_2^s = \lfloor \frac{\hat{Q}}{s} \rfloor +1$, we have  $\gamma_2 \geq Q_2^s\sqrt{2(7 + 4 \ol L_W^2/\underline{L}_W^2)}$. Thus, \eqref{thm1: descent_W2} becomes

\vspace{-0.3cm}
{\begin{align}
	&\E[F(\wt\Wb^{s, Q^s}, \Hb^{s, Q^s})|\Ec^{s-1}] - \E[F(\wt\Wb^{s, Q_1}, \Hb^{s, Q_1})|\Ec^{s-1}]\notag \\
	\leq &  - \bigg(\frac{d^s - L_W^s}{2}\bigg)\sum_{t = Q_1 + 1}^{Q^s}\E[\|\wt\Wb^{s, t} - \wt\Wb^{s, t-1}\|_F^2|\Ec^{s-1}] + \frac{ 2(1 + 8/m)(\frac{11\zeta^2}{3} + \phi^2)C_1^s}{d^s\gamma_2^2 } + \frac{8Q_2^s\zeta^2}{md^s} \\
	\leq &  - \frac{\gamma_2 - 1}{2}\sum_{t = Q_1 + 1}^{Q^s}L_{W}^s\E[\|\wt\Wb^{s, t} - \wt\Wb^{s, t-1}\|_F^2|\Ec^{s-1}] + \frac{ 2(1 + 8/m)(\frac{11\zeta^2}{3} + \phi^2)C_1^s}{\gamma_2^3 L_W^s } + \frac{8Q_2^s\zeta^2}{m\gamma_2 L_W^s}, \label{thm1: descent_W4}
	\end{align}}\vspace{-0.3cm}

\noindent where \eqref{thm1: descent_W4} follows since $d^s = \gamma_2 L_W^s$.
Then combing \eqref{thm1: descent_H2} and \eqref{thm1: descent_W4} and taking expectation over two sides yields

\vspace{-0.4cm}
{\begin{align}
	&\frac{\gamma_1 - 1}{2} \sum_{t = 1}^{Q_1} \sum_{p = 1}^{P}\omega_p\ol L_H\E[\|\Hb_p^{s, t-1} - \Hb_p^{s, t}\|_F^2] \notag \\
	&~~~~~~+ \frac{\gamma_2 - 1}{2}\sum_{t = Q_1 + 1}^{Q^s}L_{W}^s\E[\|\wt\Wb^{s, t} - \wt\Wb^{s, t-1}\|_F^2] \notag \\
	\leq & \E[F(\wt\Wb^{s, 0}, \Hb^{s, 0})] -\E[F(\wt\Wb^{s, Q^s}, \Hb^{s, Q^s})] \notag \\
	&~~+ \frac{ 2(1 + 8/m)(\frac{11\zeta^2}{3} + \phi^2)C_1^s}{\gamma_2^3 L_W^s }  + \frac{8Q_2^s\zeta^2}{m\gamma_2 L_W^s}.\label{thm1: descent_WH}
	\end{align}}\vspace{-0.3cm}

\noindent\underline{\bf Derivation of the Main Result:} 
We next derive the convergence in terms of the optimal gap functions in \eqref{eqn: prox_H} and \eqref{eqn: prox_W}.
Since $c_p^s = \frac{\gamma_1 \ol L_H}{2}$, we have from \eqref{thm1: descent_WH} that

\vspace{-0.4cm}
{
	\begin{align}
	&\sum_{t = 1}^{Q_1}\E[G_{H}(\wt \Wb^{s,t-1}, \Hb^{s,t-1})] \notag \\
	=& \sum_{t = 1}^{Q_1}\sum_{p=1}^{P}\omega_p (c_p^s)^2 \E[\|\Hb_p^{s, t-1} - \Hb_p^{s, t}\|_F^2] \notag %
	\\
	\leq & \frac{\gamma_1^2 \ol L_H}{2(\gamma_1 - 1)}\bigg(\E[F(\wt\Wb^{s, 0}, \Hb^{s, 0})] -\E[F(\wt\Wb^{s, Q^s}, \Hb^{s, Q^s})]\bigg) \notag \\
	&+ \frac{(1 + 8/m)(\frac{11\zeta^2}{3} + \phi^2)C_1^s\gamma_1^2\ol L_H}{\gamma_2^3(\gamma_1 - 1) L_{W}^s}  + \frac{4Q_2^s\zeta^2 \gamma_1^2\ol L_H}{m\gamma_2(\gamma_1 - 1)L_{W}^s},\label{thm1: descent_Hs}
	\end{align}}
\vspace{-0.3cm}

\noindent Then, summing   \eqref{thm1: descent_Hs}  up from $s = 1$ to $S$ yields

\vspace{-0.3cm}
{ 
	\begin{align}
	& \sum_{s = 1}^{S}\sum_{t = 1}^{Q_1} \E[G_{H}(\wt \Wb^{s,t-1}, \Hb^{s,t-1})] \notag \\
	\leq & \frac{\gamma_1^2 \ol L_{H}}{2(\gamma_1 - 1)}\bigg(F(\wt\Wb^{1, 0}, \Hb^{1, 0}) - \underline{F}\bigg)  + \frac{4\zeta^2\gamma_1^2 \ol L_{H}\sum_{s = 1}^{S}Q_2^s}{m\gamma_2(\gamma_1 - 1)\underline{L}_{W}} \notag \\
	&+ \frac{(1+ 8/m)(\frac{11}{3}\zeta^2 + \phi^2)\gamma_1^2\ol L_{H}\sum_{s = 1}^{S}C_1^s}{\gamma_2^3(\gamma_1 - 1) \underline{L}_{W}} . \label{thm1: descent_Hs_prox}
	\end{align}}
\vspace{-0.3cm}

\noindent 
where \eqref{thm1: descent_Hs_prox} follows from \eqref{eqn: bound of Lip const}.

Similarly, we can also have from \eqref{thm1: descent_WH} that
%
%
%

\vspace{-0.4cm}
{ \begin{align}
	&\sum_{s = 1}^{S}\sum_{t = Q_1 + 1}^{Q^s}\E[\|\wt\Wb^{s, t} - \wt\Wb^{s, t-1}\|_F^2] \notag \\
	\leq & \frac{2}{(\gamma_2 - 1)\underline{L}_W}\bigg(F(\wt\Wb^{1, 0}, \Hb^{1, 0}) -\E[F(\wt\Wb^{S +1, 0}, \Hb^{S + 1, 0})] \bigg)\notag %
	\\
	&~+ \sum_{s = 1}^{S} \E\bigg[\frac{4(1 + 8/m)(\frac{11\zeta^2}{3} + \phi^2)C_1^s}{\gamma_2^3(\gamma_2 - 1) (L_{W}^s)^2}  + \frac{16Q_2^s\zeta^2}{m\gamma_2(\gamma_2 - 1)(L_{W}^s)^2}\bigg] \notag \\
	\leq & \frac{2}{(\gamma_2 - 1)\underline{L}_W}\bigg(F(\wt\Wb^{1, 0}, \Hb^{1, 0}) - \underline{F}\bigg)  + \frac{16\zeta^2\sum_{s = 1}^{S}Q_2^s}{m\gamma_2(\gamma_2 - 1)\underline{L}_{W}^2}\notag \\
	&~ + \frac{4(1 + 8/m)(\frac{11\zeta^2}{3} + \phi^2)\sum_{s = 1}^{S}C_1^s}{\gamma_2^3(\gamma_2 - 1) \underline{L}_{W}^2}. \label{thm1: descent_Ws2}
	\end{align}}\vspace{-0.3cm}

\noindent Then, we need the following lemma to bound the optimality gap $G_{W}(\wt \Wb^{s,t}, \Hb^{s, t})$, which is proved  in \cite[Section 1.5]{FedC}.

\vspace{-0.3cm}
\begin{Lemma} \label{lem: prox_W_trans}
	For $|\Ac^s| < P$, we have
	
	\vspace{-0.3cm}
	{
		\begin{align}
		&\sum_{s = 1}^{S}\sum_{t = Q_1+1}^{Q^s}\E[G_{W}(\wt \Wb^{s-1, t}, \Hb^{s-1, t})] \notag \\
		&\leq    3(\gamma_2^2 +2)\ol L_W^2\sum_{s = 1}^{S}\sum_{t = Q_1 + 1}^{Q^s}\E[ \|\wt \Wb^{s,t} -\wt \Wb^{s,t-1}\|_F^2] \notag \\
		&~~ + \frac{(\frac{11}{3}\zeta^2 + \phi^2)\sum_{s = 1}^{S}C_2^s}{\gamma_2^2} + \frac{3(\zeta^2 + \phi^2)\sum_{s = 1}^{S}C_1^s}{2} + \frac{96\zeta^2}{m}\sum_{s = 1}^{S}Q_2^s,\notag 
		\end{align}}\vspace{-0.4cm}
	
	\noindent where $C_2^s \triangleq 6(3Q_2^s(Q_2^s - 1)/2 + 4 + 32/m)C_1^s$.
\end{Lemma}

By applying Lemma \ref{lem: prox_W_trans}, we obtain

\vspace{-0.3cm}
{
	\begin{align}
	&\sum_{s = 1}^{S}\sum_{t = Q_1 + 1}^{Q^s}\E[G_{W}(\wt \Wb^{s, t - 1}, \Hb^{s, t-1})] \notag \\
	\leq & 3(\gamma_2^2 +2)\ol L_W^2\bigg[\frac{2}{(\gamma_2 - 1)\underline{L}_W}\bigg(F(\wt\Wb^{1, 0}, \Hb^{1, 0}) - \underline{F}\bigg) \notag \\
	&+ \frac{4(1 + 8/m)(\frac{11}{3}\zeta^2 + \phi^2)\sum_{s = 1}^{S}C_1^s}{\gamma_2^3(\gamma_2 - 1) \underline{L}_{W}^2}  + \frac{16\zeta^2\sum_{s = 1}^{S}Q_2^s}{m\gamma_2(\gamma_2 - 1)\underline{L}_{W}^2}\bigg] \notag\\
	& + \frac{(\frac{11}{3}\zeta^2 + \phi^2)\sum_{s = 1}^{S}C_2^s}{\gamma_2^2} + \frac{3(\zeta^2 + \phi^2)\sum_{s = 1}^{S}C_1^s}{2} +  \frac{96\zeta^2}{m} \sum_{s = 1}^{S}Q_2^s \notag 
		\end{align}
	\begin{align}
	\leq & \frac{6(\gamma_2^2 +2)\ol L_W^2}{(\gamma_2 - 1)\underline{L}_W}\bigg(F(\wt\Wb^{1, 0}, \Hb^{1, 0}) - \underline{F}\bigg) \notag \\
	&+ \frac{12(\gamma_2^2 +2)(1 + 8 / m)(\frac{11}{3}\zeta^2 + \phi^2)\ol L_W^2\sum_{s = 1}^{S}C_1^s}{\gamma_2^3(\gamma_2 - 1)\underline{L}_W^2} \notag \\
	&+ \frac{48(\gamma_2^2 +2)\zeta^2\ol L_W^2\sum_{s = 1}^{S}Q_2^s}{m \gamma_2(\gamma_2 - 1)\underline{L}_W^2}\notag \\
	& + \frac{(\frac{11}{3}\zeta^2 + \phi^2)\sum_{s = 1}^{S}C_2^s}{\gamma_2^2} + \frac{3(\zeta^2 + \phi^2)\sum_{s = 1}^{S}C_1^s}{2}  +  \frac{96\zeta^2}{m}\sum_{s = 1}^{S}Q_2^s.   \label{thm1: descent_Ws_prox}
	\end{align}}\vspace{-0.1cm}

Combing \eqref{thm1: descent_Hs_prox} and \eqref{thm1: descent_Ws_prox} and dividing two sides of it by $T = \sum_{s = 1}^{S}Q^s =SQ_1+  \sum_{s = 1}^{S}Q_2^s$ yields

\vspace{-0.3cm}
{
	\begin{align}
	& \frac{1}{T}\bigg[\sum_{s = 1}^{S}\sum_{t = 1}^{Q_1} \E[G_{H}(\wt \Wb^{s,t-1}, \Hb^{s, t-1})] +\sum_{s = 1}^{S}\sum_{t = Q_1 + 1}^{Q^s}\E[G_{W}(\wt \Wb^{s,t - 1}, \Hb^{s, t - 1})] \bigg]\notag \\
	\leq & \bigg(\frac{\gamma_1^2 \ol L_{H}}{2(\gamma_1 - 1)} + \frac{6(\gamma_2^2 + 1)\ol L_W^2}{(\gamma_2 - 1)\underline{L}_W}\bigg) \bigg[\frac{1}{T}\bigg(F(\wt\Wb^{1, 0}, \Hb^{1, 0}) - \underline{F}\bigg)\notag \\
	&+ \frac{2(1+ 8/m)(\frac{11}{3}\zeta^2 + \phi^2)\sum_{s = 1}^{S}C_1^s}{T\gamma_2^3\underline{L}_W} + \frac{8\zeta^2\sum_{s = 1}^{S}Q_2^s}{Tm\gamma_2\underline{L}_W} \bigg] \notag \\
	& + \frac{(\frac{11}{3}\zeta^2 + \phi^2)\sum_{s = 1}^{S}C_2^s}{T\gamma_2^2} + \frac{3(\zeta^2 + \phi^2)\sum_{s = 1}^{S}C_1^s}{2T}  +  \frac{96\zeta^2}{m}\frac{\sum_{s = 1}^{S}Q_2^s}{T}
	\notag \\
	\leq & \frac{D}{T}\bigg(F(\wt\Wb^{1, 0}, \Hb^{1, 0}) - \underline{F}\bigg)  + \frac{8D\zeta^2}{m\gamma_2\underline{L}_W} +  \frac{96\zeta^2}{m} \notag \\
	&+ \frac{2D(1+ 8/m)(\frac{11}{3}\zeta^2 + \phi^2)\sum_{s = 1}^{S}C_1^s}{T\gamma_2^3\underline{L}_W}  \notag 
	\\
	& + \frac{(\frac{11}{3}\zeta^2 + \phi^2)\sum_{s = 1}^{S}C_2^s}{T\gamma_2^2} + \frac{3(\zeta^2 + \phi^2)\sum_{s = 1}^{S}C_1^s}{2T}, \label{thm1: rate2}
	\end{align}}
\vspace{-0.3cm}

\noindent where  $D \triangleq \frac{\gamma_1^2 \ol L_{H}}{2(\gamma_1 - 1)} + \frac{6(\gamma_2^2 + 1)\ol L_W^2}{(\gamma_2 - 1)\underline{L}_W}$, and the 2nd and 3rd terms in the right hand side of \eqref{thm1: rate2} follow because $\sum_{s = 1}^{S} Q_2^s/T \leq 1$. 
%
This completes the poof.
\hfill $\blacksquare$

\vspace{-0.1cm}

\bibliography{refs20,refs10}

\end{document}


\bibliographystyle{IEEEtran}
%

\begin{center}
\LARGE
\bf Supplementary Materials: Proofs

\end{center}
%
%

{\setcounter{section}{0}}

{\setcounter{equation}{0}
	\renewcommand{\theequation}{S.\arabic{equation}}

\section{Proofs of Lemmas for Theorem 1}

\subsection{Proof of Lemma 1}

Firstly, by (13) and (37), we have
	\begin{align}
	0 = \frac{1}{m}\sum_{p \in \Ac^s} \nabla_{W}F_p(\Wb_p^{s, t-1}, \Hb_p^{s, t-1}) + d^s (\ol\Wb^{s, t} - \ol\Wb^{s, t-1}).  \label{thm1: local optimality}
	\end{align}
Secondly, consider the following term
	\begin{align}
	&\langle \nabla_{W} F(\wt\Wb^{s, t-1}, \Hb^{s, t-1}) + d^s (\wt\Wb^{s, t} - \wt\Wb^{s, t-1}), \wt\Wb^{s, t-1} - \wt\Wb^{s, t} \rangle \notag \\
	= & \bigg\langle \nabla_{W} F(\wt\Wb^{s, t-1}, \Hb^{s, t-1}) + d^s (\wt\Wb^{s, t} - \wt\Wb^{s, t-1})  \notag \\
	&- \frac{1}{m}\sum_{p \in \Ac^s} \nabla_{W}F_p(\Wb_p^{s, t-1}, \Hb_p^{s, t-1})-d^s (\ol\Wb^{s, t} - \ol\Wb^{s, t-1}), \wt\Wb^{s, t-1} - \wt\Wb^{s, t} \bigg\rangle \label{thm1: inner_prod1} \\
	= &  \bigg\langle \nabla_{W} F(\wt\Wb^{s, t-1}, \Hb^{s, t-1}) - \frac{1}{m}\sum_{p \in \Ac^s} \nabla_{W}F_p(\wt\Wb_p^{s, t-1}, \Hb_p^{s, t-1}), \wt\Wb^{s, t-1} - \wt\Wb^{s, t} \bigg\rangle \notag \\
	& +d^s \langle \wt\Wb^{s, t} - \ol\Wb^{s, t}, \wt\Wb^{s, t-1} - \wt\Wb^{s, t} \rangle + d^s \langle \wt\Wb^{s, t-1} - \ol\Wb^{s, t-1}, \wt\Wb^{s, t} - \wt\Wb^{s, t-1} \rangle \notag \\
	\geq &  \bigg\langle \nabla_{W} F(\wt\Wb^{s, t-1}, \Hb^{s, t-1}) - \frac{1}{m}\sum_{p \in \Ac^s} \omega_p\nabla_{W}F_p(\wt\Wb_p^{s, t-1}, \Hb_p^{s, t-1}), \wt\Wb^{s, t-1} - \wt\Wb^{s, t} \bigg\rangle, \label{thm1: inner_prod2}
	\end{align}
where \eqref{thm1: inner_prod1} holds due to \eqref{thm1: local optimality}, and \eqref{thm1: inner_prod2} follows because 
\begin{align}\label{thm1: opt proj}
&\langle \wt\Wb^{s, t} - \ol\Wb^{s, t}, \wt\Wb^{s, t-1} - \wt\Wb^{s, t} \rangle \geq 0, \notag \\
&\langle \wt\Wb^{s, t-1} - \ol\Wb^{s, t-1}, \wt\Wb^{s, t} - \wt\Wb^{s, t-1} \rangle \geq 0.
\end{align}
Inequalities in \eqref{thm1: opt proj} are obtained by the fact that 
$\wt\Wb^{s, t} =\Pc_{\Wc}\{\ol\Wb^{s, t}\}$ and $\wt\Wb^{s, t-1} = \Pc_{\Wc}\{\ol\Wb^{s, t-1}\}$, and the application of the optimality condition 
$\langle \xb^\star - \zb, \xb - \xb^{\star} \rangle \geq 0, \forall x \in \Xc $
of the projection problem $\xb^{\star} = \arg\min\limits_{\xb \in \Xc} ~\frac{1}{2} \|\xb - \zb \|_2^2$,
where $\Xc$ is a closed convex set \cite[Proposition 3.1.1]{BK:NLPBertsekasV2}. 
Rearranging the terms in \eqref{thm1: inner_prod2} yields
\begin{align}
&\langle \nabla_{W} F(\wt\Wb^{s, t-1}, \Hb^{s, t-1}), \wt\Wb^{s, t} - \wt\Wb^{s, t-1}\rangle \notag \\
\leq& - d^s \|\wt\Wb^{s, t} - \wt\Wb^{s, t-1}\|_F^2 \notag \\
&+ \bigg\langle \nabla_{W} F(\wt\Wb^{s, t-1}, \Hb^{s, t-1}) -\frac{1}{m}\sum_{p \in \Ac^s}\nabla_{W}F_p(\Wb_p^{s, t-1}, \Hb_p^{s, t-1}), \wt\Wb^{s, t} - \wt\Wb^{s, t-1}\bigg \rangle. \label{thm1: descent_immediate}
\end{align}
\hfill $\blacksquare$

\subsection{Proof of Lemma 2}

Firstly, we have $\forall t \in \Qc_2^s$,
\begin{align}
& \E\bigg[\bigg\| \nabla_{W} F(\wt\Wb^{s, t-1}, \Hb^{s, t-1}) - \frac{1}{m}\sum_{p \in \Ac^s}\nabla_{W}F_p(\Wb_p^{s, t-1}, \Hb_p^{s, t-1})\bigg\|_F^2\bigg|\Ec^{s-1}\bigg] \notag \\
=& \E\bigg[\bigg\| \nabla_{W} F(\wt\Wb^{s, t-1}, \Hb^{s, t-1}) - \sum_{p = 1}^{P}\omega_p \nabla_{W}F_p(\Wb_p^{s, t-1}, \Hb_p^{s, t-1}) \notag \\
&~~~~~+\sum_{p = 1}^{P}\omega_p \nabla_{W}F_p(\Wb_p^{s, t-1}, \Hb_p^{s, t-1})- \frac{1}{m}\sum_{p \in \Ac^s} \nabla_{W}F_p(\Wb_p^{s, t-1}, \Hb_p^{s, t-1})\bigg\|_F^2\bigg|\Ec^{s-1}\bigg] \notag \\
\leq & 2 \underbrace{\E\bigg[\bigg\| \nabla_{W} F(\wt\Wb^{s, t-1}, \Hb^{s, t-1}) - \sum_{p = 1}^{P}\omega_p \nabla_{W}F_p(\Wb_p^{s, t-1}, \Hb_p^{s, t-1})\bigg\|_F^2\bigg|\Ec^{s-1}\bigg]}_{\rm \triangleq (S.a)} \notag \\
& +2\underbrace{\E\bigg[\bigg\|\sum_{p = 1}^{P}\omega_p \nabla_{W}F_p(\Wb_p^{s, t-1}, \Hb_p^{s, t-1})- \frac{1}{m}\sum_{p \in \Ac^s}\nabla_{W}F_p(\Wb_p^{s, t-1}, \Hb_p^{s, t-1})\bigg\|_F^2\bigg|\Ec^{s-1}\bigg]}_{\rm\triangleq (S.b)}, \label{thm3: g_bound}
\end{align}
where the inequality \eqref{thm3: g_bound} follows because of the basic inequality $\|\sum_{i = 1}^{n}\ab_i\|_2^2 \leq n\sum_{i = 1}^{n}\|\ab_i\|_2^2$.
The term ${\rm (S.a)}$ can be bounded by
\begin{align}
&\E\bigg[\bigg\| \nabla_{W} F(\wt\Wb^{s, t-1}, \Hb^{s, t-1}) - \sum_{p = 1}^{P}\omega_p \nabla_{W}F_p(\Wb_p^{s, t-1}, \Hb_p^{s, t-1})\bigg\|_F^2\bigg|\Ec^{s-1}\bigg] \notag \\
=&\E\bigg[\bigg\| \sum_{p = 1}^{P}\omega_p \bigg(\nabla_{W} F_p(\wt\Wb^{s, t-1}, \Hb_p^{s, t-1}) -  \nabla_{W}F_p(\Wb_p^{s, t-1}, \Hb_p^{s, t-1})\bigg)\bigg\|_F^2\bigg|\Ec^{s-1}\bigg] \notag \\
\leq&\sum_{p = 1}^{P}\omega_p\E[\| \nabla_{W} F_p(\wt\Wb^{s, t-1}, \Hb_p^{s, t-1}) -  \nabla_{W}F_p(\Wb_p^{s, t-1}, \Hb_p^{s, t-1})\|_F^2|\Ec^{s-1}] \notag \\
\leq&\sum_{p = 1}^{P}\omega_p(L_{W_p}^s)^2\E[\| \wt\Wb^{s, t-1} -  \Wb_p^{s, t-1}\|_F^2|\Ec^{s-1}]. \label{lem2: Sa_bound1}
\end{align}
We can also bound the term  ${\rm (S.b)}$ by
\begin{align}
&\E\bigg[\bigg\|\sum_{p = 1}^{P} \omega_p \nabla_{W}F_p(\Wb_p^{s, t-1}, \Hb_p^{s, t-1}) - \frac{1}{m}\sum_{p \in \Ac^s} \nabla_{W}F_p(\Wb_p^{s, t-1}, \Hb_p^{s,t-1})\bigg\|_F^2\bigg|\Ec^{s-1}\bigg] \notag \\
= & \frac{1}{m^2}\E\bigg[\bigg\|\sum_{p^\prime \in \Ac^s}\bigg(\sum_{p = 1}^{P} \omega_p \nabla_{W}F_p(\Wb_p^{s, t-1}, \Hb_p^{s, t-1}) -   \nabla_{W}F_{p^\prime}(\Wb_{p^\prime}^{s, t-1}, \Hb_{p^\prime}^{s, t-1})\bigg)\bigg\|_F^2\bigg|\Ec^{s-1}\bigg] \notag \\
= & \frac{1}{m^2}\E\bigg[\sum_{p^\prime \in \Ac^s}\bigg\|\sum_{p = 1}^{P} \omega_p \nabla_{W}F_p(\Wb_p^{s, t-1}, \Hb_p^{s, t-1}) -   \nabla_{W}F_{p^\prime}(\Wb_{p^\prime}^{s, t-1}, \Hb_{p^\prime}^{s, t-1})\bigg\|_F^2\bigg|\Ec^{s-1}\bigg] \notag 
\end{align}
\begin{align}
&+ \E\bigg[\sum_{p_1 \ne p_2 \in \Ac^s}\bigg\langle\sum_{p = 1}^{P} \omega_p \nabla_{W}F_p(\Wb_p^{s, t-1}, \Hb_p^{s, t-1}) -   \nabla_{W}F_{p_1}(\Wb_{p_1}^{s, t-1}, \Hb_{p_1}^{s, t-1}), \notag \\
&~~~~~~~~~~~~~~~~\sum_{p = 1}^{P} \omega_p \nabla_{W}F_p(\Wb_p^{s, t-1}, \Hb_p^{s, t-1}) -   \nabla_{W}F_{p_2}(\Wb_{p_2}^{s, t-1}, \Hb_{p_2}^{s, t-1}) \bigg \rangle\bigg|\Ec^{s-1}\bigg]\notag \\
= & \frac{1}{m^2}\E\bigg[\sum_{p^\prime \in \Ac^s}\bigg\|\sum_{p = 1}^{P} \omega_p \nabla_{W}F_p(\Wb_p^{s, t-1}, \Hb_p^{s, t-1}) -   \nabla_{W}F_{p^\prime}(\Wb_{p^\prime}^{s, t-1}, \Hb_{p^\prime}^{s, t-1})\bigg\|_F^2|\Ec^{s-1}\bigg] \label{lem2: Sb_bound1} \\
= & \frac{1}{m^2} m \sum_{i = 1}^{P}\omega_i \bigg\|\sum_{p = 1}^{P} \omega_p \nabla_{W}F_p(\Wb_p^{s, t-1}, \Hb_p^{s, t-1}) -   \nabla_{W}F_{i}(\Wb_{i}^{s, t-1}, \Hb_{i}^{s, t-1})\bigg\|_F^2  \label{lem2: Sb_bound2}\\
\leq & \frac{1}{m}  \sum_{i = 1}^{P}\omega_i \sum_{p = 1}^{P} \omega_p\underbrace{\| \nabla_{W}F_p(\Wb_p^{s, t-1}, \Hb_p^{s, t-1}) -   \nabla_{W}F_{i}(\Wb_{i}^{s, t-1}, \Hb_{i}^{s, t-1})\|_F^2 }_{\triangleq \rm (S.c)} \label{lem2: Sb_bound3},
\end{align}
where \eqref{lem2: Sb_bound1} and \eqref{lem2: Sb_bound2} follow due to independent sampling with replacement and unbiasedness, i.e., $\E_{p^\prime}[ \nabla_{W}F_{p^\prime}(\Wb_{p^\prime}^{s, t-1}, \Hb_{p^\prime}^{s, t-1})|\Ec^{s -1}]=\sum_{p =1}^{P}\omega_p \nabla_{W}F_p(\Wb_p^{s, t-1}, \Hb_p^{s, t-1})$;  \eqref{lem2: Sb_bound3} follows because of convexity of $\|\cdot\|_2^2$. The term $\rm (S.c)$ can be bounded as follows.
\begin{align}
{\rm (S.c)} =& \|\nabla_{W}F_i(\Wb_i^{s, t-1}, \Hb_i^{s, t-1}) - \nabla_{W} F_p(\Wb_p^{s, t-1}, \Hb_p^{s, t-1})\|_F^2 \notag \\
\leq & \bigg\|\nabla_{W}F_i(\Wb_i^{s, t-1}, \Hb_i^{s, t-1}) - \nabla_{W}F_i(\wt\Wb^{s, t-1}, \Hb_i^{s, t-1}) \notag \\ 
&~~+ \nabla_{W}F_i(\wt\Wb^{s, t-1}, \Hb_i^{s, t-1}) - \nabla_{W}F(\wt\Wb^{s, t-1}, \Hb^{s, t-1}) +  \nabla_{W}F(\wt\Wb^{s, j}, \Hb^{s, t-1})\notag \\
&~~- \bigg(\nabla_{W}F_p(\Wb_p^{s,t-1}, \Hb_p^{s, t-1}) - \nabla_{W}F_p(\wt\Wb^{s,t-1}, \Hb_p^{s, t-1}) \notag \\
&~~+ \nabla_{W}F_p(\wt\Wb^{s, t-1}, \Hb_p^{s, t-1}) - \nabla_{W}F(\wt\Wb^{s,t-1}, \Hb^{s, t-1}) + \nabla_{W}F(\wt\Wb^{s, t-1}, \Hb^{s, t-1})\bigg)\bigg\|_F^2 \notag \\
\leq& 4\|\nabla_{W}F_i(\Wb_i^{s, t-1}, \Hb_i^{s, t-1}) - \nabla_{W}F_i(\wt\Wb^{s, t-1}, \Hb_i^{s, t-1})\|_F^2 \notag \\
&~~+ 4\|\nabla_{W}F_i(\wt\Wb^{s, t-1}, \Hb_i^{s, t-1}) - \nabla_{W}F(\wt\Wb^{s, t-1}, \Hb^{s, t-1})\|_F^2 \notag \\
&~~+ 4\|\nabla_{W}F_p(\Wb_p^{s, t-1}, \Hb_p^{s, t-1}) - \nabla_{W}F_p(\wt\Wb^{s, t-1}, \Hb_p^{s, t-1})\|_F^2 \notag \\
&~~+ 4\|\nabla_{W}F_p(\wt\Wb^{s, t-1}, \Hb_p^{s, t-1}) - \nabla_{W}F(\wt\Wb^{s, t-1}, \Hb^{s, t-1})\|_F^2 \notag \\
\leq&4(L_{W_i}^s)^2\|\wt\Wb^{s, t-1} - \Wb_i^{s, t-1}\|_F^2 + 8\zeta^2 + 4(L_{W_p}^s)^2\|\wt\Wb^{s, t-1} - \Wb_p^{s, t-1}\|_F^2, \label{lemS1: bound_c}
\end{align}
where the first term and the third term in the right hand side (RHS) of \eqref{lemS1: bound_c} come from the Lipschitz continuity of $\nabla_{W} F_p(\cdot, \Hb_p)$, and the second term in the RHS of \eqref{lemS1: bound_c} follows because of the bound in (18). Substituting \eqref{lemS1: bound_c} into \eqref{lem2: Sb_bound3} yields

\begin{align}
&\E\bigg[\bigg\|\sum_{p = 1}^{P} \omega_p \nabla_{W}F_p(\Wb_p^{s, t-1}, \Hb_p^{s, t-1}) - \frac{1}{m}\sum_{p \in \Ac^s} \nabla_{W}F_p(\Wb_p^{s, t-1}, \Hb_p^{s,t-1})\bigg\|_F^2\bigg|\Ec^{s-1}\bigg] \notag \\
\leq & \frac{1}{m} \sum_{i = 1}^{P} \omega_i \sum_{p = 1}^{P} \omega_p \bigg(4(L_{W_i}^s)^2\|\wt\Wb^{s, t-1} - \Wb_i^{s, t-1}\|_F^2 + 8\zeta^2 + 4(L_{W_p}^s)^2\|\wt\Wb^{s, t-1} - \Wb_p^{s, t-1}\|_F^2\bigg)  \label{lem2: Sb_bound4}\\
\leq & \frac{8}{m} \zeta^2 + \frac{8}{m} \sum_{p = 1}^{P} \omega_p (L_{W_p})^2 \|\wt\Wb^{s, t-1} - \Wb_p^{s, t-1}\|_F^2. \label{lem2: Sb_bound5}
\end{align}
Then, after further substituting \eqref{lem2: Sa_bound1} and  \eqref{lem2: Sb_bound5} into \eqref{thm3: g_bound}, we have
\begin{align}
& \E\bigg[\bigg\| \nabla_{W} F(\wt\Wb^{s, t-1}, \Hb^{s, t-1}) - \frac{1}{m}\sum_{p \in \Ac^s}\nabla_{W}F_p(\Wb_p^{s, t-1}, \Hb_p^{s, t-1})\bigg\|_F^2\bigg|\Ec^{s-1}\bigg] \notag \\
\leq & 2\sum_{p = 1}^{P}\omega_p(L_{W_p}^s)^2\E[\| \wt\Wb^{s, t-1} -  \Wb_p^{s, t-1}\|_F^2|\Ec^{s-1}]  +  \frac{16}{m}\zeta^2 \notag \\
&~~+  \frac{16}{m}\sum_{p = 1}^{P} \omega_p(L_{W_p}^s)^2\|\wt \Wb^{s, t-1} - \Wb_p^{s, t-1}\|_F^2 \notag \\
= & 2\bigg(1 + \frac{8}{m}\bigg) \sum_{p = 1}^{P}\omega_p(L_{W_p}^s)^2\E[\| \wt\Wb^{s, t-1} -  \Wb_p^{s, t-1}\|_F^2|\Ec^{s-1}] +  \frac{16}{m}\zeta^2. \label{thm3: bound_g}
\end{align}
\hfill $\blacksquare$

\subsection{Proof of Lemma 3}

Firstly, we have
\begin{align}
&\sum_{t = Q_1+1}^{Q^s}\sum_{p = 1}^{P} \omega_p (L_{W_p}^{s})^2 \E[\|\wt\Wb^{s, t-1}- \Wb_p^{s, t-1}\|_F^2|\Ec^{s-1}] \notag \\
\leq & \sum_{t = Q_1+1}^{Q^s}\sum_{p = 1}^{P} \omega_p (L_{W_p}^{s})^2 \bigg(2\E[\|\wt\Wb^{s, t-1}- \ol\Wb^{s, t-1}\|_F^2|\Ec^{s-1}]  + 2\E[\|\ol\Wb^{s, t-1}- \Wb_p^{s, t-1}\|_F^2|\Ec^{s-1}]\bigg) \notag \\
= & 2 \underbrace{\sum_{t = Q_1+1}^{Q^s}\sum_{p = 1}^{P} \omega_p (L_{W_p}^{s})^2\E[\|\wt\Wb^{s, t-1}- \ol\Wb^{s, t-1}\|_F^2|\Ec^{s-1}]}_{\triangleq \rm (S.d)} \notag \\
&~~~~~~~~+ 2\underbrace{\sum_{t = Q_1+1}^{Q^s}\sum_{p = 1}^{P} \omega_p (L_{W_p}^{s})^2\E[ \|\ol\Wb^{s, t-1}- \Wb_p^{s, t-1}\|_F^2|\Ec^{s-1}]}_{\triangleq \rm (S.e)}. \label{thm1: c_bound}
\end{align}

In order to obtain the bound of ${\rm (S.d)}$ and ${\rm (S.e)}$, we need the following lemma which is proved in Section \ref{sec: lem6}.
\begin{Lemma} \label{lem: diff_local}
	For all $t \in \{Q_1\} \cup \Qc_2^s $, we have
	\begin{align}
	&\E[\|\ol\Wb^{s, t} - \Wb_p^{s, t}\|_F^2|\Ec^{s-1}] \notag \\
	\leq&~ \frac{4(t  - Q_1)}{(d^s)^2} \sum_{j=Q_1}^{t-1}\sum_{p=1}^{P} \omega_q(L_{W_p}^s)^2\E[\|\wt\Wb^{s,j} - \Wb_p^{s, j}\|_F^2|\Ec^{s-1}] + \frac{8(t-Q_1)^2}{(d^s)^2}\zeta^2 \notag \\
	&~+ \frac{4(t  - Q_1)}{(d^s)^2} \sum_{j=Q_1}^{t-1} (L_{W_p}^s)^2\E[\|\wt\Wb^{s,j} - \Wb_p^{s, j}\|_F^2|\Ec^{s-1}], \label{lem: diff_local_bound1}\\
	&\E[\|\wt\Wb^{s, t} - \ol\Wb^{s, t}\|_F^2|\Ec^{s-1}] \notag \\
	\leq&~ \frac{3(t - Q_1)}{(d^s)^2}\sum_{j = Q_1}^{t - 1} \sum_{p = 1}^{P} \omega_p (L_{W_p}^s)^2\E[\|\wt\Wb^{s, j}-\Wb_p^{s, j}\|^2|\Ec^{s-1}] + \frac{3(t - Q_1)^2(\zeta^2 + \phi^2)}{(d^s)^2}. \label{lem: diff_local_bound2}
	\end{align}
\end{Lemma}
By applying Lemma \ref{lem: diff_local}, we have 
\begin{align}
{\rm (S.e)} =& \sum_{t = Q_1+1}^{Q^s}\sum_{p = 1}^{P} \omega_p (L_{W_p}^{s})^2 \E[\|\ol\Wb^{s, t-1}- \Wb_p^{s, t-1}\|_F^2|\Ec^{s-1}] \notag \\
\leq &  \sum_{t = Q_1+1}^{Q^s}\sum_{p = 1}^{P} \omega_p (L_{W_p}^{s})^2  \bigg(\frac{4(t -1 - Q_1)}{(d^s)^2} \sum_{j=Q_1}^{t-2}\sum_{q=1}^{P} \omega_q(L_{W_q}^s)^2\E[\|\wt\Wb^{s,j} - \Wb_q^{s, j}\|_F^2|\Ec^{s-1}] \notag \\
&~~~~~~~+ \frac{4(t -1- Q_1)}{(d^s)^2} \sum_{j=Q_1}^{t-2} (L_{W_p}^s)^2\E[\|\wt\Wb^{s,j} - \Wb_p^{s, j}\|_F^2|\Ec^{s-1}] + \frac{8(t-1- Q_1)^2}{(d^s)^2}\zeta^2\bigg) \notag \\
= & \sum_{t = Q_1+1}^{Q^s} \frac{4(t - 1- Q_1) }{(d^s)^2/(L_W^s)^2} \sum_{j=Q_1}^{t-2}\sum_{p=1}^{P} \omega_p(L_{W_p}^s)^2\E[\|\wt\Wb^{s,j} - \Wb_p^{s, j}\|_F^2 |\Ec^{s-1}]\notag \\
&+ \sum_{t = Q_1+1}^{Q^s}\sum_{p = 1}^{P} \omega_p (L_{W_p}^{s})^2\frac{4(t -1- Q_1)}{(d^s)^2/(L_W^s)^2} \sum_{j=Q_1}^{t-2}  \frac{(L_{W_p}^{s})^2}{(L_W^s)^2}\E[\|\wt\Wb^{s,j} - \Wb_p^{s, j}\|_F^2|\Ec^{s-1}] \notag \\
& + \sum_{t = Q_1+1}^{Q^s}\frac{8(t-1-Q_1)^2}{(d^s)^2/(L_W^s)^2}\zeta^2  \label{thm1: e_bound0}  \\
\leq & \sum_{t = Q_1+1}^{Q^s} \frac{4(t - 1- Q_1) }{(d^s)^2/(L_W^s)^2} \sum_{j=Q_1}^{t-2}\sum_{p=1}^{P} \omega_p(L_{W_p}^s)^2\E[\|\wt\Wb^{s,j} - \Wb_p^{s, j}\|_F^2 |\Ec^{s-1}]\notag \\
&+ \sum_{t = Q_1+1}^{Q^s}\sum_{p = 1}^{P} \omega_p (L_{W_p}^{s})^2\frac{4(t -1- Q_1)}{(d^s)^2/(L_W^s)^2}\bigg(\frac{\ol L_W^2}{\underline{L}_W^2}\bigg) \sum_{j=Q_1}^{t-2} \E[\|\wt\Wb^{s,j} - \Wb_p^{s, j}\|_F^2|\Ec^{s-1}] \notag \\
& + \sum_{t = Q_1+1}^{Q^s}\frac{8(t-1-Q_1)^2}{(d^s)^2/(L_W^s)^2}\zeta^2 \label{thm1: e_bound1}  
\end{align}
\begin{align}
= & \sum_{t = Q_1+1}^{Q^s} \frac{4(t - 1- Q_1) }{\gamma_2^2}\bigg(1+ \frac{\ol L_W^2}{\underline{L}_W^2}\bigg) \sum_{j=Q_1}^{t-2}\sum_{p=1}^{P} \omega_p(L_{W_p}^s)^2\E[\|\wt\Wb^{s,j} - \Wb_p^{s, j}\|_F^2|\Ec^{s-1}] \notag \\
& + \sum_{t = Q_1+1}^{Q^s}\frac{8(t-1-Q_1)^2}{\gamma_2^2}\zeta^2 \label{thm1: e_bound2} \\
\leq & \frac{2Q_2^s(Q_2^s - 1)}{\gamma_2^2}(1+ \frac{\ol L_W^2}{\underline{L}_W^2})\sum_{t = Q_1 +1}^{Q^s}\sum_{p=1}^{P} \omega_p(L_{W_p}^s)^2\E[\|\wt\Wb^{s,t-1} - \Wb_p^{s, t-1}\|_F^2|\Ec^{s-1}] \notag \\
&~+\frac{4Q_2^s(Q_2^s - 1)(2Q_2^s - 1)\zeta^2}{3\gamma_2^2}, \label{thm1: e_bound3}
\end{align}
where \eqref{thm1: e_bound0} follows since $(L_W^s)^2 = \sum_{p = 1}^{P}\omega_p(L_{W_p}^s)^2$; \eqref{thm1: e_bound1} follows due to Assumption 2 and $\frac{(L_{W_p}^s)^2}{(L_W^s)^2} \leq \frac{\ol L_W^2}{\underline{L}_W^2}$; \eqref{thm1: e_bound2} holds since $d^s = \gamma_2 L_W^s$, and \eqref{thm1: e_bound3} follows because $\forall a_j > 0$,
\begin{align}
\sum_{t=Q_1 + 1}^{Q^s} (t-1-Q_1)\sum_{j = Q_1}^{t-2}a_j 
\leq \sum_{t=Q_1 + 1}^{Q^s} \frac{Q_2^s(Q_2^s-1)}{2} a_{t-1}, \label{thm1: e_bound4}
\end{align}
and
\begin{align} 
\sum_{t = Q_1 + 1}^{Q^s} (t - 1 - Q_1 )^2 = \frac{Q_2^s(Q_2^s - 1)(2Q_2^s - 1)}{6} \label{eqn: square_sum}.
\end{align}
Similarly, we can bound $\rm (S.d)$ as follows.
\begin{align}
{\rm (S.d)} =&\sum_{t = Q_1+1}^{Q^s}\sum_{p = 1}^{P} \omega_p (L_{W_p}^{s})^2\E[\|\wt\Wb^{s, t-1}- \ol\Wb^{s, t-1}\|_F^2|\Ec^{s-1}] \notag \\
\leq &\sum_{t = Q_1+1}^{Q^s}\sum_{p = 1}^{P} \omega_p (L_{W_p}^{s})^2 \bigg(\frac{3(t -1- Q_1)^2(\zeta^2 + \phi^2)}{(d^s)^2} \notag \\
&~~~~~+\frac{3(t -1- Q_1)}{(d^s)^2}\sum_{j = Q_1}^{t - 2} \sum_{p = 1}^{P} \omega_p (L_{W_p}^s)^2\E[\|\wt\Wb^{s, j}-\Wb_p^{s, j}\|_F^2|\Ec^{s-1}] \bigg) \notag \\
\leq& \sum_{t = Q_1+1}^{Q^s}\frac{3(t -1- Q_1)}{(d^s)^2/(L_W^s)^2}\sum_{j = Q_1}^{t - 2} \sum_{p = 1}^{P} \omega_p (L_{W_p}^s)^2\E[\|\wt\Wb^{s, j}-\Wb_p^{s, j}\|_F^2|\Ec^{s-1}] \notag \\
&~~~~~+\sum_{t = Q_1+1}^{Q^s} \frac{3(t -1- Q_1)^2(\zeta^2 + \phi^2)}{(d^s)^2/( L_W^s)^2} \label{thm1: d_bound0} \\
\leq &~\frac{3Q_2^s(Q_2^s - 1)}{2(d^s)^2/(L_W^s)^2}\sum_{t = Q_1+1}^{Q^s} \sum_{p=1}^{P} \omega_p(L_{W_p}^s)^2\E[\|\wt\Wb^{s,t-1} - \Wb_p^{s, t-1}\|_F^2|\Ec^{s-1}] \notag \\
&~~~~+  \frac{Q_2^s(Q_2^s - 1)(2Q_2^s -1)(\zeta^2 + \phi^2)}{2(d^s)^2/(L_W^s)^2}  \label{thm1: d_bound1} 
\end{align}
\begin{align}
= &~\frac{3Q_2^s(Q_2^s - 1)}{2\gamma_2^2}\sum_{t = Q_1+1}^{Q^s} \sum_{p=1}^{P} \omega_p(L_{W_p}^s)^2\E[\|\wt\Wb^{s,t-1} - \Wb_p^{s, t-1}\|_F^2|\Ec^{s-1}] \notag \\
&~~~~+  \frac{Q_2^s(Q_2^s - 1)(2Q_2^s -1)(\zeta^2 + \phi^2)}{2\gamma_2^2},
\label{thm1: d_bound2}
\end{align}
where \eqref{thm1: d_bound0} follows since $(L_W^s)^2 = \sum_{p = 1}^{P}\omega_p(L_{W_p}^s)^2$; \eqref{thm1: d_bound1} follows due to \eqref{thm1: e_bound4} and \eqref{eqn: square_sum}, and \eqref{thm1: d_bound2} holds since $d^s = \gamma_2 L_W^s$.
Then, substituting \eqref{thm1: d_bound2} and \eqref{thm1: e_bound3} into \eqref{thm1: c_bound} yields 
\begin{align}
	&\sum_{t = Q_1+1}^{Q^s}\sum_{p = 1}^{P} \omega_p (L_{W_p}^{s})^2 \E[\|\wt\Wb^{s, t-1}- \Wb_p^{s, t-1}\|_F^2|\Ec^{s-1}] \notag \\
	\leq&~\frac{Q_2^s(Q_2^s - 1)}{\gamma_2^2}\bigg(7+ 4\frac{(\ol L_W)^2}{(\underline{L}_W)^2}\bigg)\sum_{t = Q_1+1}^{Q^s} \sum_{p=1}^{P} \omega_p(L_{W_p}^s)^2\E[\|\wt\Wb^{s,t-1} - \Wb_p^{s, t-1}\|_F^2|\Ec^{s-1}] \notag \\
	&~~~+ \frac{C_1^s(\frac{11}{3}\zeta^2 + \phi^2)}{\gamma_2^2}, \label{thm1: c_bound1}
\end{align}
where 
\begin{align}
C_1^s  \triangleq Q_2^s(Q_2^s - 1)(2Q_2^s -1). \label{thm1: C_1_def}
\end{align}
Since $\gamma_2 \geq Q_2^s\sqrt{2(7+ 4\ol L_W^2/\underline{L}_W^2)}$, then $\gamma_2^2 > 2Q_2^s(Q_2^s -1)(7+ 4\ol L_W^2/\underline{L}_W^2)$. 
After rearranging \eqref{thm1: c_bound1}, we obtain
\begin{align}
	&\sum_{t = Q_1+1}^{Q^s}\sum_{p = 1}^{P} \omega_p (L_{W_p}^{s})^2 \E[\|\wt\Wb^{s, t-1}- \Wb_p^{s, t-1}\|_F^2|\Ec^{s-1}] \notag \\
	 \leq&~ \frac{Q_2^s(Q_2^s - 1)(2Q_2^s -1)(\frac{11\zeta^2}{3} + \phi^2)}{\gamma_2^2 - Q_2^s(Q_2^s - 1)(7+ 4\ol L_W^2/\underline{L}_W^2)} \notag \\
	\leq &~\frac{2C_1^s(\frac{11\zeta^2}{3} + G^2)}{\gamma_2^2}, \label{thm1: c_bound2}
\end{align}
where \eqref{thm1: c_bound2} follows since $\gamma_2^2 - Q_2^s(Q_2^s -1)(7+ 4\ol L_W^2/\underline{L}_W^2) > \frac{\gamma_2^2}{2}$.
\hfill $\blacksquare$

\subsection{Poof of Lemma \ref{lem: diff_local}}
\label{sec: lem6}
According to the definition of $\ol\Wb^{s, t}$, we have $\forall t \in \Qc_2^s$,
\begin{align}
\ol\Wb^{s, t} &= \frac{1}{m}\sum\limits_{p \in \Ac^s}  \Wb_p^{s, t}\notag  \\
& = \frac{1}{m} \sum_{p \in \Ac^s} \bigg(\Wb^{s} -\frac{1}{d^s} \sum_{j = Q_1}^{t - 1} \nabla_{W}F_p(\Wb_p^{s, j}, \Hb_p^{s, j})\bigg) \label{lem2: W_def0} \\
& = \Wb^{s} - \frac{1}{d^sm}\sum_{j = Q_1}^{t - 1} \sum_{p \in \Ac^s}  \nabla_{W}F_p(\Wb_p^{s, j},\Hb_p^{s, j}), \label{lem2: W_def}
\end{align}
where \eqref{lem2: W_def0} is obtained by applying (13), i.e.,
\begin{align}\label{lem2: W_def5}
\Wb_p^{s,t} = \Wb^{s} - \frac{1}{d^s}\sum_{j = Q_1}^{t-1} \nabla_{W} F_p(\Wb_p^{s, j}, \Hb_p^{s, j}).
\end{align}
As a result, by \eqref{lem2: W_def} and \eqref{lem2: W_def5}, we have
\begin{align}
&\E[\|\ol\Wb^{s, t} - \Wb_p^{s, t}\|_F^2|\Ec^{s-1}] \notag \\ 
= &\E\bigg[\bigg\|\Wb^{s} - \frac{1}{d^sm}\sum_{j = Q_1}^{t - 1} \sum_{p \in \Ac^s} \nabla_{W}F_p(\Wb_p^{s, j}, \Hb_p^{s, j}) \notag \\
&~~~~~~~~~~~~~~~~~~~~~~~~~ - \bigg(\Wb^{s} - \frac{1}{d^s}\sum_{j = Q_1}^{t-1} \nabla_{W} F_p(\Wb_p^{s, j}, \Hb_p^{s, j})\bigg)\bigg\|_F^2\bigg|\Ec^{s-1}\bigg] \notag  \\
= &  \frac{1}{(d^s)^2}\E\bigg[\bigg\|\frac{1}{m}\sum_{j = Q_1}^{t - 1} \sum_{p \in \Ac^s} \nabla_{W}F_p(\Wb_p^{s, j}, \Hb_p^{s, j}) - \sum_{j = Q_1}^{t - 1} \nabla_{W} F_p(\Wb_p^{s, j}, \Hb_p^{s, j})\bigg\|_F^2\bigg|\Ec^{s-1}\bigg] \notag  \\
\leq &\frac{(t-Q_1)}{(d^s)^2}\sum_{j = Q_1}^{t - 1}\E\bigg[\bigg\| \frac{1}{m}\sum_{i \in \Ac^s} \nabla_{W}F_i(\Wb_i^{s, j}, \Hb_i^{s, j}) - \nabla_{W} F_p(\Wb_p^{s, j}, \Hb_p^{s, j})\bigg\|_F^2\bigg|\Ec^{s-1}\bigg] \notag \\
\leq &\frac{(t-Q_1)}{(d^s)^2m}\sum_{j = Q_1}^{t - 1}\E\bigg[\sum_{i \in \Ac^s}\|\nabla_{W}F_i(\Wb_i^{s, j}, \Hb_i^{s, j}) - \nabla_{W} F_p(\Wb_p^{s, j}, \Hb_p^{s, j})\|_F^2\bigg|\Ec^{s-1}\bigg] \notag \\
= &\frac{(t-Q_1)}{(d^s)^2}\sum_{j = Q_1}^{t - 1}\sum_{i =1}^{P} \omega_i \underbrace{\|\nabla_{W}F_i(\Wb_i^{s, j}, \Hb_i^{s, j}) - \nabla_{W} F_p(\Wb_p^{s, j}, \Hb_p^{s, j})\|_F^2}_{\rm \triangleq (S.f)}, \label{lem2: bound_diff1}
\end{align}  
where \eqref{lem2: bound_diff1} follows since $\Ac^s$ is obtained by sampling with replacement. The term ${\rm (S.f)}$ can be bounded by the same procedure to obtain \eqref{lemS1: bound_c}, i.e.,
\begin{align}
{\rm (S.f)} \leq 4(L_{W_i}^s)^2\|\wt\Wb^{s, j} - \Wb_i^{s, j}\|_F^2 + 8\zeta^2 + 4(L_{W_p}^s)^2\|\wt\Wb^{s, j} - \Wb_p^{s, j}\|_F^2. \label{lem6: bound_diff2}
\end{align}
Then, substituting \eqref{lem6: bound_diff2} into \eqref{lem2: bound_diff1} yields
\begin{align}
&\E[\|\ol\Wb^{s, t} - \Wb_p^{s, t}\|_F^2|\Ec^{s-1}] \notag\\
\leq&~\frac{(t-Q_1)}{(d^s)^2}\sum_{j = Q_1}^{t - 1}\sum_{i = 1}^{P} \omega_i \bigg(4(L_{W_i}^s)^2\|\wt\Wb^{s, j} - \Wb_i^{s, j}\|_F^2 + 8\zeta^2 \notag \\
&~~~~~~~~~~~~~~~~~~~~~~~~~~~~~~+ 4(L_{W_p}^s)^2\|\wt\Wb^{s, j} - \Wb_p^{s, j}\|_F^2\bigg) \notag \\
=&~\frac{4(t-Q_1)}{(d^s)^2}\sum_{j = Q_1}^{t - 1}\sum_{i = 1}^{P} \omega_i(L_{W_i}^s)^2\|\wt\Wb^{s, j} - \Wb_i^{s, j}\|_F^2  + \frac{8(t-Q_1)^2}{(d^s)^2}\zeta^2 \notag \\
&~~+ \frac{4(t  - Q_1)}{(d^s)^2} \sum_{j=Q_1}^{t-1} (L_{W_p}^s)^2\|\ol\Wb^{s,j} - \Wb_p^{s, j}\|_F^2.
\end{align}

On the other hand, to prove \eqref{lem: diff_local_bound2}, note that
\begin{align}\label{lem2: opt}
\|\wt\Wb^{s, t} - \ol\Wb^{s, t}\|_F^2\leq &\|\Wb^{s} - \ol\Wb^{s, t}\|_F^2,
\end{align} 
since $\wt \Wb^{s, t} = \Pc_{\Wc}(\ol \Wb^{s, t})$. Then, by \eqref{lem2: opt} and \eqref{lem2: W_def}, we have
\begin{align}
&\E[\|\wt\Wb^{s, t} - \ol\Wb^{s, t}\|_F^2|\Ec^{s-1}] \notag \\
\leq &\E\bigg[ \bigg\|\Wb^{s} - \Wb^{s} + \frac{1}{d^sm}\sum_{j = Q_1}^{t - 1} \sum_{p \in \Ac^s}  \nabla_{W}F_p(\Wb_p^{s, j}, \Hb_p^{s, j})\bigg\|_F^2\bigg|\Ec^{s-1}\bigg] \label{lem2: def}\\
= & \frac{1}{(d^s)^2}\E\bigg[\bigg\|\sum_{j = Q_1}^{t - 1} \frac{1}{m}\sum_{p \in \Ac^s}  \nabla_{W}F_p(\Wb_p^{s, j}, \Hb_p^{s, j})\bigg\|_F^2\bigg|\Ec^{s-1}\bigg] \\
\leq & \frac{(t - Q_1)}{(d^s)^2m} \sum_{j = Q_1}^{t - 1}\E\bigg[ \sum_{p \in \Ac^s}\|\nabla_{W}F_p(\Wb_p^{s, j}, \Hb_p^{s, j})\|_F^2\bigg|\Ec^{s-1}\bigg] \label{lem2: inequality}\\
\leq & \frac{(t - Q_1)}{(d^s)^2}\sum_{j = Q_1}^{t - 1} \sum_{p = 1}^{P} \omega_p \|\nabla_{W}F_p(\Wb_p^{s, j}, \Hb_p^{s, j})\|_F^2 \label{lem2: convex} \\
=& \frac{(t - Q_1)}{(d^s)^2}\sum_{j = Q_1}^{t - 1} \sum_{p = 1}^{P} \omega_p \|\nabla_{W}F_p(\Wb_p^{s, j}, \Hb_p^{s, j}) - \nabla_{W}F_p(\wt\Wb^{s, j}, \Hb_p^{s, j})\notag  \\
&~~ + \nabla_{W}F_p(\wt\Wb^{s, j}, \Hb_p^{s, j})- \nabla_{W}F(\wt\Wb^{s, j}, \Hb^{s, j}) + \nabla_{W}F(\wt\Wb^{s, j}, \Hb^{s, j})\|_F^2 \\
\leq& \frac{(t - Q_1)}{(d^s)^2}\sum_{j = Q_1}^{t - 1} \sum_{p = 1}^{P} \omega_p \bigg(3\|\nabla_{W}F(\wt\Wb^{s, j}, \Hb^{s, j})\|_F^2 \notag \\
&~~~~~~~~~~~~~~~~~~~~~~+ 3\|\nabla_{W}F_p(\Wb_p^{s, j}, \Hb_p^{s, j}) - \nabla_{W}F_p(\wt\Wb^{s, j}, \Hb_p^{s, j})\|^2 \notag \\
&~~~~~~~~~~~~~~~~~~~~~~+ 3\|\nabla_{W}F_p(\wt\Wb^{s, j}, \Hb_p^{s, j}) - \nabla_{W}F(\wt\Wb^{s, j}, \Hb^{s, j})\|^2\bigg) \label{lem2: inequality1} \\
\leq& \frac{3(t - Q_1)}{(d^s)^2}\sum_{j = Q_1}^{t - 1} \sum_{p = 1}^{P} \omega_p (L_{W_p}^s)^2\|\wt\Wb^{s, j}-\Wb_p^{s, j}\|^2  + \frac{3(t - Q_1)^2(\zeta^2 + \phi^2)}{(d^s)^2}, \label{lem1: inequality2} 
\end{align}
where \eqref{lem2: inequality} and \eqref{lem2: inequality1} are obtained by the basic inequality $\|\sum_{i = 1}^{n}\ab_i\|_2^2 \leq n\sum_{i = 1}^{n}\|\ab_i\|_2^2$, \eqref{lem2: convex} is obtained by the Jensen's inequality for the convex function $\|\cdot\|_F^2$, and \eqref{lem1: inequality2} holds due to Assumption 1 and the bound in (18).
\hfill $\blacksquare$

\subsection{Poof of Lemma 4}

We have $\forall t \in \Qc_2^s$, and $\forall p \in \Pc$,
\begin{align}
\Wb_p^{s, t} = \Wb_p^{s, t - 1} - \frac{\nabla_{W} F_p(\Wb_p^{s, t-1}, \Hb_p^{s, t-1})}{d^s}.
\end{align}
Then, due to PCC, we obtain
\begin{align}
\wt \Wb^{s, t} = \Pc_{\Wc}(\ol \Wb^{s, t}) = \Pc_{\Wc}\bigg(\ol \Wb^{s, t-1} - \frac{1}{md^s}\sum_{p \in \Ac^s}\nabla_{W} F_p(\Wb_p^{s, t-1}, \Hb_p^{s, t-1}) \bigg),
\end{align}
where $|\Ac^s| < P$. Thus, we have
	\begin{align}
	&\E[G_{W}(\wt \Wb^{s, t}, \Hb^{s,t})|\Ec^{s - 1}] \notag \\
	=&\E\bigg[(d^s)^2\bigg\|\wt \Wb^{s, t} - \Pc_{\Wc}\bigg(\wt \Wb^{s, t} - \frac{\nabla_{W} F(\wt \Wb^{s,t}, \Hb^{s,t})}{d^s}\bigg)\bigg\|_F^2\bigg|\Ec^{s - 1}\bigg] \notag \\
	= & \E\bigg[(d^s)^2\bigg\|\Pc_{\Wc}\bigg(\ol \Wb^{s, t-1} - \frac{1}{md^s}\sum_{p \in \Ac^s}\nabla_{W} F_p(\Wb_p^{s, t-1}, \Hb_p^{s, t-1})\bigg) \notag \\
	&~~~~~~~~~~~~- \Pc_{\Wc}\bigg(\wt \Wb^{s, t} - \frac{\nabla_{W} F(\wt \Wb^{s,t}, \Hb^{s,t})}{d^s}\bigg)\bigg\|_F^2\bigg|\Ec^{s - 1}\bigg] \notag \\
	\leq &\E\bigg[(d^s)^2\bigg\|\ol \Wb^{s, t-1}- \wt \Wb^{s, t} \notag \\
	&~~~~~~~~~~~~- \frac{1}{d^s}\bigg(\frac{1}{m}\sum_{p \in \Ac^s}\nabla_{W} F_p(\Wb_p^{s, t-1}, \Hb_p^{s, t-1}) - \nabla_{W} F(\wt \Wb^{s,t}, \Hb^{s,t})\bigg)\bigg\|_F^2\bigg|\Ec^{s - 1}\bigg] \\
	\leq &\E\bigg[(d^s)^2\bigg\|\ol \Wb^{s, t-1} - \wt \Wb^{s, t-1} + \wt\Wb^{s, t-1} - \wt \Wb^{s, t} \notag \\
	&~~~~~~~~~~~~- \frac{1}{d^s}\bigg(\frac{1}{m}\sum_{p \in \Ac^s}\nabla_{W} F_p(\Wb_p^{s, t-1}, \Hb_p^{s, t-1}) - \nabla_{W} F(\wt \Wb^{s,t}, \Hb^{s,t})\bigg)\bigg\|_F^2\bigg|\Ec^{s - 1}\bigg] \\
	\leq & 3(d^s)^2\underbrace{\E[\|\ol \Wb^{s, t-1} - \wt \Wb^{s, t-1}\|_F^2|\Ec^{s - 1}]}_{\rm \triangleq (S.g)} + 3(d^s)^2\E[ \|\wt \Wb^{s, t-1} - \wt \Wb^{s, t}\|_F^2|\Ec^{s - 1}] \notag \\
	&~~~~~~~~~~~~+ 3\underbrace{\E\bigg[ \bigg\|\nabla_{W} F(\wt \Wb^{s,t}, \Hb^{s,t}) - \frac{1}{m}\sum_{p \in \Ac^s}\nabla_{W} F_p(\Wb_p^{s, t-1}, \Hb_p^{s, t-1})\bigg\|_F^2\bigg|\Ec^{s - 1}\bigg]}_{\rm \triangleq (S.h)}. \label{lem2: W_stationary}
	\end{align}
	Then, we need to obtain the bounds of $\rm (S.g)$ and $\rm (S.h)$. By applying \eqref{lem: diff_local_bound2} in Lemma \ref{lem: diff_local}, we have
	\begin{align}
	{\rm (S.g)} \leq& \frac{3(t -1- Q_1)}{(d^s)^2}\sum_{j = Q_1}^{t - 2} \sum_{p = 1}^{P}\omega_p(L_{W_p}^s)^2 \E[\|\wt\Wb^{s, j}-\Wb_p^{s, j}\|^2|\Ec^{s - 1}] + \frac{3(t -1- Q_1)^2(\zeta^2 + \phi^2)}{(d^s)^2} \notag \\
	\leq& \frac{3(t - Q_1)}{(d^s)^2}\sum_{j = Q_1}^{t - 1} \sum_{p = 1}^{P}\omega_p(L_{W_p}^s)^2 \E[\|\wt\Wb^{s, j}-\Wb_p^{s, j}\|^2|\Ec^{s - 1}] + \frac{3(t - Q_1)^2(\zeta^2 + \phi^2)}{(d^s)^2}. \label{lem2: bound_e}
	\end{align}
	Moreover, for $\rm (S.h)$, we have
	\begin{align}
	{\rm (S.h)} = &\E\bigg[ \bigg\|\nabla_{W} F(\wt \Wb^{s,t}, \Hb^{s,t}) - \frac{1}{m}\sum_{p \in \Ac^s}\nabla_{W} F_p(\Wb_p^{s, t-1}, \Hb_p^{s, t-1})\bigg\|_F^2\bigg|\Ec^{s - 1}\bigg] \\
	=&\E\bigg[ \bigg\|\nabla_{W} F(\wt \Wb^{s,t}, \Hb^{s,t}) -\nabla_{W} F(\wt \Wb^{s,t-1}, \Hb^{s,t-1}) \notag \\
	&~~~~+ \nabla_{W} F(\wt \Wb^{s,t-1}, \Hb^{s,t-1})-\frac{1}{m}\sum_{p \in \Ac^s}\nabla_{W} F_p(\Wb_p^{s, t-1}, \Hb_p^{s, t-1})\bigg\|_F^2\bigg|\Ec^{s - 1}\bigg] 
	\end{align}
	\begin{align}
	\leq & 2\E[ \|\nabla_{W} F(\wt \Wb^{s,t}, \Hb^{s,t}) -\nabla_{W} F(\wt \Wb^{s,t-1}, \Hb^{s,t-1})\|_F^2|\Ec^{s - 1}] \notag \\
	&~~~~+ 2\E\bigg[\bigg\|\nabla_{W} F(\wt \Wb^{s,t-1}, \Hb^{s,t-1})-\frac{1}{m}\sum_{p \in \Ac^s}\nabla_{W} F_p(\Wb_p^{s, t-1}, \Hb_p^{s, t-1})\bigg\|_F^2\bigg|\Ec^{s - 1}\bigg] \notag\\
	\leq & 2(L_W^s)^2\E[ \|\wt \Wb^{s,t} -\wt \Wb^{s,t-1}\|_F^2|\Ec^{s - 1}] \notag \\
	&~~~~+ 4\bigg(1+ \frac{8}{m}\bigg)\sum_{p=1}^{P} \omega_p(L_{W_p}^s)^2\E[\|\wt \Wb^{s,t-1}-  \Wb_p^{s, t-1}\|_F^2|\Ec^{s - 1}] + \frac{32}{m} \zeta^2, \label{lem2: bound_g}
	\end{align}
	where \eqref{lem2: bound_g} follows by Assumption 2 and \eqref{thm3: bound_g} in Lemma 2.
	Substituting \eqref{lem2: bound_e} and \eqref{lem2: bound_g} into \eqref{lem2: W_stationary} yields
	\begin{align}
	&\E[G_{W}(\wt \Wb^{s, t}, \Hb^{s,t})|\Ec^{s - 1}] \notag \\
	\leq &  9(t - Q_1)\sum_{j = Q_1}^{t - 1} \sum_{p = 1}^{P}\omega_p(L_{W_p}^s)^2 \E[\|\wt\Wb^{s, j}-\Wb_p^{s, j}\|^2|\Ec^{s - 1}] + 9(t - Q_1)^2(\zeta^2 + \phi^2) \notag \\
	&+ 3(d^s)^2\E[ \|\wt \Wb^{s, t-1} - \wt \Wb^{s, t}\|_F^2|\Ec^{s - 1}] +6(L_W^s)^2\E[ \|\wt \Wb^{s,t} -\wt \Wb^{s,t-1}\|_F^2|\Ec^{s - 1}]\notag \\
	&+12(1 + 8/m)\sum_{p = 1}^{P}\omega_p (L_{W_p}^s)^2\|\wt \Wb^{s, t-1} - \Wb_p^{s, t-1}\|_F^2  + \frac{96\zeta^2}{m} \notag \\
	\leq &  9(t - Q_1)\sum_{j = Q_1}^{t - 1} \sum_{p = 1}^{P}\omega_p(L_{W_p}^s)^2 \E[\|\wt\Wb^{s, j}-\Wb_p^{s, j}\|^2|\Ec^{s - 1}] + 9(t - Q_1)^2(\zeta^2 + \phi^2) \notag \\
	&+ 3(\gamma_2^2 + 2)(L_W^s)^2\E[ \|\wt \Wb^{s, t-1} - \wt \Wb^{s, t}\|_F^2|\Ec^{s - 1}] \notag \\
	&+12(1 + 8/m)\sum_{p = 1}^{P}\omega_p (L_{W_p}^s)^2\|\wt \Wb^{s, t-1} - \Wb_p^{s, t-1}\|_F^2  + \frac{96\zeta^2}{m}, \label{lem2: prox_bound1}  
	\end{align}
	where \eqref{lem2: prox_bound1} follows since $d^s = \gamma_2 L_W^s$. Then, summing \eqref{lem2: prox_bound1} up from $t = Q_1$ to $Q^s -1$ yields
	\begin{align}
	&\sum_{t = Q_1}^{Q^s - 1}\E[G_{W}(\wt \Wb^{s, t}, \Hb^{s,t})|\Ec^{s - 1}] \notag \\
	\leq &  9\sum_{t = Q_1}^{Q^s-1} (t - Q_1)\sum_{j = Q_1}^{t - 1} \sum_{p = 1}^{P}\omega_p(L_{W_p}^s)^2 \E[\|\wt\Wb^{s, j}-\Wb_p^{s, j}\|^2|\Ec^{s - 1}] \notag \\
	&~~+ 9\sum_{t = Q_1 }^{Q^s-1}(t - Q_1)^2(\zeta^2 + \phi^2) + \frac{96Q_2^s\zeta^2}{m}   \notag \\
	&~~+3(\gamma_2^2 + 2)(L_W^s)^2\sum_{t = Q_1 }^{Q^s-1}\E[ \|\wt \Wb^{s,t} -\wt \Wb^{s,t-1}\|_F^2|\Ec^{s - 1}] \notag \\
	&~~+ 12(1+ 8/m)\sum_{t = Q_1 }^{Q^s-1}\sum_{p=1}^{P} \omega_p(L_{W_p}^s)^2\E[\|\wt \Wb^{s,t-1}-  \Wb_p^{s, t-1}\|_F^2|\Ec^{s - 1}] 
		\end{align}
	\begin{align}
	\leq &  \frac{9Q_2^s(Q_2^s - 1)}{2}\sum_{t = Q_1}^{Q^s-1} \sum_{p = 1}^{P}\omega_p(L_{W_p}^s)^2 \E[\|\wt\Wb^{s, t}-\Wb_p^{s, t}\|^2|\Ec^{s - 1}] \notag \\
	&~~+ \frac{3Q_2^s(Q_2^s - 1)(2Q_2^s -1)(\zeta^2 + \phi^2)}{2} + \frac{96Q_2^s\zeta^2}{m} \notag \\
	&~~+3(\gamma_2^2 +2)(L_W^s)^2\sum_{t = Q_1+1}^{Q^s}\E[ \|\wt \Wb^{s,t} -\wt \Wb^{s,t-1}\|_F^2|\Ec^{s - 1}] \notag \\
	&~~+ 12(1+8/m)\sum_{t = Q_1 }^{Q^s-1}\sum_{p=1}^{P} \omega_p(L_{W_p}^s)^2\E[\|\wt \Wb^{s,t-1}-  \Wb_p^{s, t-1}\|_F^2|\Ec^{s - 1}]  \label{lem3: prox_bound3} \\ 
	\leq &  3(3Q_2^s(Q_2^s - 1)/2 + 4 + 32/m)\sum_{t = Q_1 + 1}^{Q^s} \sum_{p = 1}^{P}\omega_p(L_{W_p}^s)^2 \E[\|\wt\Wb^{s, t-1}-\Wb_p^{s, t-1}\|^2|\Ec^{s - 1}] \notag \\
	&~~+   3(\gamma_2^2 +2)(L_W^s)^2\sum_{t = Q_1+1}^{Q^s}\E[ \|\wt \Wb^{s,t} -\wt \Wb^{s,t-1}\|_F^2|\Ec^{s - 1}] \notag \\
	&~~+ \frac{3C_1^s(\zeta^2 + \phi^2)}{2} + \frac{96Q_2^s\zeta^2}{m}   \label{lem3: prox_bound4} \\
	\leq &  3(\gamma_2^2 +2)(L_W^s)^2\sum_{t = Q_1 + 1}^{Q^s}\E[ \|\wt \Wb^{s,t} -\wt \Wb^{s,t-1}\|_F^2|\Ec^{s - 1}] \notag \\
	&~~+ \frac{C_2^s(\frac{11}{3}\zeta^2 + \phi^2)}{\gamma_2^2} + \frac{3C_1^s(\zeta^2 + \phi^2)}{2}  + \frac{96Q_2^s\zeta^2}{m}, \label{lem2: bound_prox}
	\end{align}
	where
	\begin{align}
		C_2^s \triangleq 6(3Q_2^s(Q_2^s - 1)/2 + 4 + 32/m)C_1^s,
	\end{align} 
	the first term in the right hand side of \eqref{lem3: prox_bound3} follows because $\forall a_j > 0$,
\begin{align}
\sum_{t=Q_1}^{Q^s-1} (t-Q_1)\sum_{j = Q_1}^{t-1}a_j 
\leq \sum_{t=Q_1}^{Q^s - 1} \frac{Q_2^s(Q_2^s-1)}{2} a_{t}; \label{prox: summation}
\end{align}
	the second term in the RHS of \eqref{lem3: prox_bound3} follows due to \eqref{eqn: square_sum}; the fourth term in the RHS of \eqref{lem3: prox_bound3} follows because $\wt \Wb^{s, Q_1} = \wt \Wb^{s, Q_1 - 1}$;  the first term in the RHS of\eqref{lem3: prox_bound4} follows because $\wt \Wb^{s, Q_1} =  \Wb_p^{s, Q_1}$, and \eqref{lem2: bound_prox} follows by applying Lemma 3 to the first term in the RHS of \eqref{lem3: prox_bound4}. Then, taking expectation over two sides of \eqref{lem2: bound_prox} and summing it up from $s = 1$ to $S$ yields
	\begin{align}
	&\sum_{s = 1}^{S}\sum_{t = Q_1}^{Q^s-1}\E[G_{W}(\wt \Wb^{s, t}, \Hb^{s,t})] \notag \\
	\leq &   3(\gamma_2^2 +2)\ol L_W^2\sum_{s = 1}^{S}\sum_{t = Q_1 + 1}^{Q^s}\E[ \|\wt \Wb^{s,t} -\wt \Wb^{s,t-1}\|_F^2] \notag \\
	&~~+ \frac{(\frac{11}{3}\zeta^2 + \phi^2)\sum_{s = 1}^{S}C_2^s}{\gamma_2^2} + \frac{3(\zeta^2 + \phi^2)\sum_{s = 1}^{S}C_1^s}{2}  + \frac{96\zeta^2}{m} \sum_{s = 1}^{S}Q_2^s. 
	\end{align}
\hfill $\blacksquare$

\section{Poof of Corollary 1}
\noindent \underline{\bf Objective Descent w.r.t. $\Hb$:} Using the same procedure as to obtain (41) in the proof of Theorem 1, we have
{\begin{align}
&F(\wt\Wb^{s,Q_1}, \Hb^{s, Q_1}) - F(\wt\Wb^{s,0}, \Hb^{s, 0})\notag \\ 
\leq& -\frac{\gamma_1 - 1}{2} \sum_{t = 1}^{Q_1}\sum_{p=1}^{P} \omega_p\ol L_H\|\Hb_p^{s, t-1} - \Hb_p^{s, t}\|_F^2.
\label{corly1: descent_H}
\end{align}}
\noindent \underline{\bf Objective Descent w.r.t. $\Wb$:} Using the same procedure as to obtain (45) in the proof of Theorem 1, we have
\begin{align}
&F(\wt\Wb^{s, t}, \Hb^{s, t}) \notag \\
\leq & F(\wt\Wb^{s, t-1}, \Hb^{s, t-1}) - \frac{d^s - L_{W}^s}{2}\|\wt\Wb^{s, t} - \wt\Wb^{s, t-1}\|_F^2 \notag \\ 
&~~+\frac{1}{2d^s}\| \nabla_{W} F(\wt\Wb^{s, t-1}, \Hb^{s, t-1}) - \sum_{p = 1}^{P} \omega_p \nabla_{W}F_p(\Wb_p^{s, t-1}, \Hb_p^{s, t-1})\|_F^2 \\
\leq & F(\wt\Wb^{s, t-1}, \Hb^{s, t-1}) - \frac{d^s - L_{W}^s}{2}\|\wt\Wb^{s, t} - \wt\Wb^{s, t-1}\|_F^2 \notag \\ 
&~~+   \frac{1}{2d^s}\sum_{p = 1}^{P}\omega_p (L_{W_p}^s)^2\| \wt\Wb^{s, t-1} - \Wb_p^{s, t-1}\|_F^2. \label{corly1: descent_W}
\end{align}
Then, summing \eqref{corly1: descent_W} up from $t = Q_1 + 1$ to $Q^s$ yields
\begin{align}
&F(\wt\Wb^{s, Q^s}, \Hb^{s, Q^s}) \notag \\
\leq & F(\wt\Wb^{s, Q_1}, \Hb^{s, Q_1}) - \frac{d^s - L_{W}^s}{2}\sum_{t = Q_1 + 1}^{Q^s}\|\wt\Wb^{s, t} - \wt\Wb^{s, t-1}\|_F^2 \notag \\ 
&~~+   \frac{1}{2d^s}\sum_{t = Q_1 + 1}^{Q^s}\sum_{p = 1}^{P}\omega_p (L_{W_p}^s)^2\| \wt\Wb^{s, t-1} - \Wb_p^{s, t-1}\|_F^2 \\
\leq & F(\wt\Wb^{s, Q_1}, \Hb^{s, Q_1}) - \frac{d^s - L_{W}^s}{2}\sum_{t = Q_1 + 1}^{Q^s}\|\wt\Wb^{s, t} - \wt\Wb^{s, t-1}\|_F^2 \notag \\ 
&~~+ \frac{C_1^s(\frac{11\zeta^2}{3} + \phi^2)}{\gamma_2^3 L_W^s},
\label{corly1: descent_Ws}
\end{align}
where \eqref{corly1: descent_Ws} follows because of Lemma 3 and $d^s  = \gamma_2 L_W^s$. By combing \eqref{corly1: descent_H} and \eqref{corly1: descent_Ws}, we have
{\begin{align}
&\frac{\gamma_1 - 1}{2} \sum_{t = 1}^{Q_1}\sum_{p=1}^{P} \omega_p\ol L_H\|\Hb_p^{s, t-1} - \Hb_p^{s, t}\|_F^2 +\frac{\gamma_2 - 1}{2}\sum_{t = Q_1 + 1}^{Q^s}L_{W}^s\|\wt\Wb^{s, t} - \wt\Wb^{s, t-1}\|_F^2 \notag \\
\leq ~&  F(\wt\Wb^{s,0}, \Hb^{s, 0}) - F(\wt\Wb^{s, Q^s}, \Hb^{s, Q^s})  + \frac{C_1^s(\frac{11\zeta^2}{3} + \phi^2)}{\gamma_2^3 L_W^s}, \label{corly1: descent_WH} 
\end{align}}
which implies that 
\begin{align}
 &\sum_{t = 1}^{Q_1} G_{H}(\wt \Wb^{s, t-1}, \Hb^{s, t-1}) \notag \\
 = ~&\sum_{t = 1}^{Q_1}\sum_{p=1}^{P} \omega_p(c_p^s)^2\|\Hb_p^{s, t-1} - \Hb_p^{s, t}\|_F^2 \notag \\
\leq ~&\frac{\gamma_1^2 \ol L_H}{2(\gamma_1 - 1)} \bigg(F(\wt\Wb^{s,0}, \Hb^{s, 0}) - F(\wt\Wb^{s, Q^s}, \Hb^{s, Q^s})\bigg) + \frac{\gamma_1^2 \ol L_H C_1^s(\frac{11\zeta^2}{3} + G^2)}{2\gamma_2^3(\gamma_1 - 1)L_W^s}. \label{corly1: H_prox1}
\end{align}
Then summing up \eqref{corly1: H_prox1} from $s = 1$ to $S$ yields
\begin{align}
& \sum_{s = 1}^{S}\sum_{t = 1}^{Q_1} G_{H}(\wt \Wb^{s, t-1}, \Hb^{s, t-1}) \notag \\
\leq ~&\frac{\gamma_1^2 \ol L_{H}}{2(\gamma_1 - 1)} \bigg(F(\wt\Wb^{s,0}, \Hb^{s, 0}) - \underline{F} \bigg) + \frac{\gamma_1^2 \ol L_{H}(\frac{11\zeta^2}{3} + \phi^2)\sum_{s = 1}^{S}C_1^s}{2\gamma_2^3(\gamma_1 - 1)\underline{L}_W}. \label{corlly1: H_prox3}
\end{align}
Similarly, we can also have from \eqref{corly1: descent_WH} that
\begin{align}
&\sum_{t = Q_1 + 1}^{Q^s}\|\wt\Wb^{s, t} - \wt\Wb^{s, t-1}\|_F^2 \notag \\
\leq ~&\frac{2 }{(\gamma_2 - 1)L_W^s} \bigg(F(\wt\Wb^{s,0}, \Hb^{s, 0}) - F(\wt\Wb^{s, Q^s}, \Hb^{s, Q^s})\bigg) + \frac{2C_1^s(\frac{11\zeta^2}{3} + \phi^2)}{\gamma_2^3(\gamma_2 - 1)(L_W^s)^2}. \label{corly1: W_prox1}
\end{align}
By summing up \eqref{corly1: W_prox1} from $s = 1$ to $S$, we have
\begin{align}
&\sum_{s = 1}^{S}\sum_{t = Q_1 + 1}^{Q^s}\|\wt\Wb^{s, t} - \wt\Wb^{s, t-1}\|_F^2 \notag \\
\leq ~&\frac{2}{(\gamma_2 - 1)\underline{L}_W} \bigg(F(\wt\Wb^{s,0}, \Hb^{s, 0}) - \underline{F}\bigg) + \frac{2(\frac{11\zeta^2}{3} + \phi^2)\sum_{s = 1}^{S}C_1^s}{\gamma_2^3(\gamma_2 - 1)\underline{L}_W^2}. \label{corly1: W_prox2}
\end{align}
We then proceed with the following lemma which is proved in Section \ref{sec: lem7}.
\begin{Lemma} \label{lem: prox_boundW_full}
	Suppose that $|\Ac^s| = P, ~\forall s$. Then
	\begin{align}
	&\sum_{s = 1}^{S}\sum_{t = Q_1}^{Q^s - 1}G_{W}(\wt \Wb^{s, t}, \Hb^{s, t})\notag \\
	\leq & 3(\gamma_2^2 +2) \ol L_W^2\sum_{s = 1}^{S}\sum_{t = Q_1 + 1}^{Q^s}\|\wt \Wb^{s, t-1} - \wt \Wb^{s, t}\|_F^2 \notag \\
	&~~~+ \frac{(\frac{11\zeta^2}{3} + \phi^2)\sum_{s = 1}^{S}C_3^s}{\gamma_2^2} + \frac{3(\zeta^2 + \phi^2)\sum_{s = 1}^{S}C_1^s}{2},
\end{align}
	where  
	\begin{align}
		C_3^s \triangleq 6(3Q_2^s(Q_2^s - 1)/2 + 2)C_1^s, \label{thm1: C_3_def}
	\end{align} and $C_1^s$ is defined in \eqref{thm1: C_1_def}.
\end{Lemma}

By applying Lemma \ref{lem: prox_boundW_full}, we have
\begin{align}
&\sum_{s = 1}^{S}\sum_{t = Q_1+1}^{Q^s} G_{W}(\wt \Wb^{s, t-1}, \Hb^{s, t-1}) \notag \\
\leq & \frac{(\frac{11\zeta^2}{3} + \phi^2)\sum_{s = 1}^{S}C_3^s}{\gamma_2^2} + \frac{3(\zeta^2 + \phi^2)\sum_{s = 1}^{S}C_1^s}{2} \notag \\
&~+3(\gamma_2^2 +2)\ol L_W^2\bigg[\frac{2}{(\gamma_2 - 1)\underline{L}_W} \bigg(F(\wt\Wb^{s,0}, \Hb^{s, 0}) - \underline{F}\bigg) + \frac{2(\frac{11\zeta^2}{3} + \phi^2)\sum_{s = 1}^{S}C_1^s}{\gamma_2^3(\gamma_2 - 1)\underline{L}_W^2}\bigg] \\
\leq & \frac{6(\gamma_2^2 +2) \ol L_W^2 }{(\gamma_2 - 1)\underline{L}_W} \bigg(F(\wt\Wb^{s,0}, \Hb^{s, 0}) - \underline{F}\bigg) + \frac{6(\gamma_2^2 +2)\ol L_W^2(\frac{11\zeta^2}{3} + \phi^2)\sum_{s = 1}^{S}C_1^s}{\gamma_2^3(\gamma_2 - 1)\underline{L}_W^2} \notag \\
&~+ \frac{(\frac{11\zeta^2}{3} + \phi^2)\sum_{s = 1}^{S}C_3^s}{\gamma_2^2} + \frac{3(\zeta^2 + \phi^2)\sum_{s = 1}^{S}C_1^s}{2}. \label{corlly1: W_prox3}
\end{align}
Combing \eqref{corlly1: H_prox3} and \eqref{corlly1: W_prox3} and then dividing both sides by $T = \sum_{s = 1}^{S} Q_2^s$ yields
\begin{align}
&\frac{1}{T}\bigg[\sum_{s = 1}^{S}\sum_{t = 1}^{Q_1}G_{H}(\wt \Wb^{s, t-1}, \Hb^{s, t-1})   +\sum_{s = 1}^{S}\sum_{t = Q_1 +1}^{Q^s}G_{W}(\wt \Wb^{s,t-1 }, \Hb^{s, t-1})\bigg] \notag\\
\leq & \bigg(\frac{\gamma_1^2 \ol L_{H}}{2(\gamma_1 - 1)} + \frac{6(\gamma_2^2 +2)\ol L_W^2 }{(\gamma_2 - 1) \underline{L}_W}\bigg)\bigg[ \frac{1}{T}\bigg(F(\wt\Wb^{s,0}, \Hb^{s, 0}) - \underline{F} \bigg) + \frac{(\frac{11\zeta^2}{3} + \phi^2)\bigg(\sum_{s = 1}^{S}C_1^s\bigg)}{T\gamma_2^3\underline{L}_W}\bigg] \notag \\
&~~~~~~~+ \frac{(\frac{11\zeta^2}{3} + \phi^2)\bigg(\sum_{s = 1}^{S}C_3^s\bigg)}{T\gamma_2^2} + \frac{3(\zeta^2 + \phi^2)\bigg(\sum_{s = 1}^{S}C_1^s\bigg)}{2T} \\
\leq &   \frac{D}{T}\bigg(F(\wt\Wb^{s,0}, \Hb^{s, 0}) - \underline{F} \bigg) + \frac{1}{T} \bigg [  \frac{3}{2}({\zeta^2} + \phi^2)\bigg(\sum_{s = 1}^{S}C_1^s\bigg) +  \frac{(\frac{11\zeta^2}{3} + \phi^2)}{\gamma_2^2} \bigg( \frac{D \sum_{s = 1}^{S}C_1^s}{\gamma_2\underline{L}_W} + \sum_{s = 1}^{S}C_3^s\bigg) \bigg], \label{corly: conv_rate1}  
\end{align}
where  
\begin{align}
	D \triangleq \frac{\gamma_1^2 \ol L_{H}}{2(\gamma_1 - 1)} + \frac{6(\gamma_2^2 + 1)\ol L_W^2}{(\gamma_2 - 1)\underline{L}_W}.
\end{align}

\subsection{Poof of Lemma \ref{lem: prox_boundW_full}} \label{sec: lem7}

Suppose that $|\Ac^s| = P, ~\forall s$. Since $\wt \Wb^{s, t} = \Pc_{\Wc}\bigg(\sum_{p =1}^{P}\omega_p \Wb_p^{s, t}\bigg)$, we have $\forall t \in \Qc_2^s$,
\begin{align}
\wt \Wb^{s, t} = \Pc_{\Wc}(\ol \Wb^{s, t}) = \Pc_{\Wc}\bigg(\ol \Wb^{s, t-1} - \frac{1}{d^s}\sum_{p = 1}^{P}\omega_p\nabla_{W} F_p(\Wb_p^{s, t-1}, \Hb_p^{s, t-1}) \bigg),
\end{align}
and then 
\begin{align}
&G_{W}(\wt \Wb^{s, t}, \Hb^{s, t}) \notag \\
= &(d^s)^2\bigg\|\wt \Wb^{s, t} - \Pc_{\Wc}\bigg(\wt \Wb^{s, t} - \frac{\nabla_{W} F(\wt \Wb^{s,t}, \Hb^{s,t})}{d^s}\bigg)\bigg\|_F^2 \notag \\
= & (d^s)^2\bigg\|\Pc_{\Wc}\bigg(\ol \Wb^{s, t-1} - \frac{1}{d^s}\sum_{p = 1}^{P}\omega_p\nabla_{W} F_p(\Wb_p^{s, t-1}, \Hb_p^{s, t-1})\bigg) - \Pc_{\Wc}\bigg(\wt \Wb^{s, t} - \frac{\nabla_{W} F(\wt \Wb^{s,t}, \Hb^{s,t})}{d^s}\bigg)\bigg\|_F^2 \notag \\
\leq & (d^s)^2\bigg\| \ol \Wb^{s, t-1}- \wt \Wb^{s, t} - \frac{1}{d^s}\bigg(\sum_{p =1}^{P}\omega_p\nabla_{W} F_p(\Wb_p^{s, t-1}, \Hb_p^{s, t-1}) - \nabla_{W} F(\wt \Wb^{s,t}, \Hb^{s,t})\bigg)\bigg\|_F^2 \label{lemS2: bound1}\\
\leq & (d^s)^2\bigg\|\ol \Wb^{s, t-1}- \wt \Wb^{s, t-1} + \wt \Wb^{s, t-1} - \wt \Wb^{s, t} \notag \\
&~~~~~~~~- \frac{1}{d^s}\bigg(\sum_{p =1}^{P}\omega_p\nabla_{W} F_p(\Wb_p^{s, t-1}, \Hb_p^{s, t-1}) - \nabla_{W} F(\wt \Wb^{s,t}, \Hb^{s,t})\bigg)\|_F^2 \label{lemS2: bound2}\\
\leq & 3 (d^s)^2\underbrace{\|\ol \Wb^{s, t-1} - \wt \Wb^{s, t-1}\|_F^2}_{\triangleq \rm (S.m)} + 3 (d^s)^2 \|\wt \Wb^{s, t-1} - \wt \Wb^{s, t}\|_F^2 \notag \\
&~~+ 3 \underbrace{\bigg\|\sum_{p =1}^{P}\omega_p\nabla_{W} F_p(\Wb_p^{s, t-1}, \Hb_p^{s, t-1}) - \nabla_{W} F(\wt \Wb^{s,t}, \Hb^{s,t})\bigg\|_F^2}_{\triangleq \rm (S.n)} \label{lem7: bound3} 
\end{align}
The term $(\rm S.m)$ can be bounded by applying Lemma \ref{lem: diff_local} as follows.
\begin{align}
{\rm (S.m)} \leq&~ \frac{3(t -1- Q_1)}{(d^s)^2}\sum_{j = Q_1}^{t - 2} \sum_{p = 1}^{P}\omega_p(L_{W_p}^s)^2 \|\wt\Wb^{s, j}-\Wb_p^{s, j}\|^2 + \frac{3(t -1- Q_1)^2(\zeta^2 + \phi^2)}{(d^s)^2} \notag \\
\leq&~ \frac{3(t - Q_1)}{(d^s)^2}\sum_{j = Q_1}^{t - 1} \sum_{p = 1}^{P}\omega_p(L_{W_p}^s)^2 \|\wt\Wb^{s, j}-\Wb_p^{s, j}\|^2 + \frac{3(t - Q_1)^2(\zeta^2 + \phi^2)}{(d^s)^2}. \label{lem7: bound_g}
\end{align}
We can also bound $(\rm S.n)$ by
\begin{align}
{\rm (S.n)} = &~\bigg\|\sum_{p =1}^{P}\omega_p\nabla_{W} F_p(\Wb_p^{s, t-1}, \Hb_p^{s, t-1}) - \nabla_{W} F(\wt \Wb^{s,t}, \Hb^{s,t})\bigg\|_F^2 \notag \\
= &~\bigg\|\sum_{p =1}^{P}\omega_p\bigg(\nabla_{W} F_p(\Wb_p^{s, t-1}, \Hb_p^{s, t-1}) - \nabla_{W} F_p(\wt \Wb^{s,t-1}, \Hb_p^{s,t-1})\bigg) \notag \\
&~~~~+ \nabla_{W} F(\wt \Wb^{s,t-1}, \Hb^{s,t}) - \nabla_{W} F(\wt \Wb^{s,t}, \Hb^{s,t})\bigg\|_F^2 \label{lem7: sh_bound1}\\
\leq &~2\bigg\|\sum_{p =1}^{P}\omega_p\bigg(\nabla_{W} F_p(\Wb_p^{s, t-1}, \Hb_p^{s, t-1}) - \nabla_{W} F_p(\wt \Wb^{s,t-1}, \Hb_p^{s,t-1})\bigg)\bigg\|_F^2 \notag \\
&~~~~+ 2\|\nabla_{W} F(\wt \Wb^{s,t-1}, \Hb^{s,t}) - \nabla_{W} F(\wt \Wb^{s,t}, \Hb^{s,t})\|_F^2 \label{lem7: sh_bound2}
\end{align}
\begin{align}
\leq &~2\sum_{p = 1}^{P} \omega_p (L_{W_p}^s)^2\|\Wb_p^{s, t-1} - \wt \Wb^{s,t-1}\|_F^2 + 2 (L_{W}^s)^2 \|\wt \Wb^{s, t-1} - \wt \Wb^{s, t}\|_F^2, \label{lem7: sh_bound3}
\end{align}
where \eqref{lem7: sh_bound1} follows because $\Hb_p^{s,t} = \Hb_p^{s, t-1}, \forall t \in \Qc_2^s$, and \eqref{lem7: sh_bound3} follows due to the convexity of $\|\cdot\|_2^2$ and Lipschitz continuity of $\nabla_{W} F_p(\cdot, \cdot)$ and $\nabla_{W} F(\cdot, \cdot)$.
Then, substituting \eqref{lem7: bound_g} and \eqref{lem7: sh_bound3} into \eqref{lem7: bound3} yields
\begin{align}
&G_{W}(\wt \Wb^{s, t}, \Hb^{s, t}) \notag \\
\leq &9(t - Q_1)\sum_{j = Q_1}^{t - 1} \sum_{p = 1}^{P}\omega_p(L_{W_p}^s)^2 \|\wt\Wb^{s, j}-\Wb_p^{s, j}\|^2 + 9(t - Q_1)^2(\zeta^2 + \phi^2) \notag \\
&~~+ 3(\gamma_2^2 + 2) (L_W^s)^2\|\wt \Wb^{s, t-1} - \wt \Wb^{s, t}\|_F^2 \notag \\
&~~+ 6\sum_{p = 1}^{P} \omega_p (L_{W_p}^s)^2\|\Wb_p^{s, t-1} - \wt \Wb^{s,t-1}\|_F^2. \label{lem7: bound4}
\end{align}
By summing \eqref{lem7: bound4} up from $t = Q_1$ to $Q_2^s - 1$, we have
\begin{align}
&\sum_{t = Q_1}^{Q^s - 1}G_{W}(\wt \Wb^{s, t}, \Hb^{s, t}) \notag \\
\leq & \sum_{t = Q_1}^{Q^s-1}9(t - Q_1)\sum_{j = Q_1}^{t - 1} \sum_{p = 1}^{P}\omega_p(L_{W_p}^s)^2 \|\wt\Wb^{s, j}-\Wb_p^{s, j}\|^2 + 9\sum_{t = Q_1}^{Q^s - 1}(t- Q_1)^2(\zeta^2 + \phi^2)\notag \\
&~~+3(\gamma_2^2 + 2)(L_W^s)^2\sum_{t = Q_1}^{Q^s-1}\|\wt \Wb^{s, t-1} - \wt \Wb^{s, t}\|_F^2  \notag \\
&~~+6\sum_{t = Q_1}^{Q^s-1}\sum_{p = 1}^{P}\omega_p  (L_{W_p}^s)^2\|\wt \Wb^{s, t-1} - \Wb_p^{s, t-1}\|_F^2 \notag \\
\leq &\frac{9Q_2^s(Q_2^s - 1)}{2}\sum_{t = Q_1}^{Q^s-1} \sum_{p = 1}^{P}\omega_p(L_{W_p}^s)^2 \|\wt\Wb^{s, t}-\Wb_p^{s, t}\|^2 + \frac{3C_1^s(\zeta^2 + \phi^2)}{2} \notag \\
&~~+3(\gamma_2^2 + 2)(L_W^s)^2\sum_{t = Q_1}^{Q^s-1}\|\wt \Wb^{s, t-1} - \wt \Wb^{s, t}\|_F^2\notag \\
&~~+6\sum_{t = Q_1}^{Q^s-1}\sum_{p = 1}^{P}\omega_p  (L_{W_p}^s)^2\|\wt \Wb^{s, t-1} - \Wb_p^{s, t-1}\|_F^2  \label{lemS2: prox_bound1} \\
= & 3(3Q_2^s(Q_2^s - 1)/2 + 2)\sum_{t = Q_1 + 1}^{Q^s} \sum_{p = 1}^{P}\omega_p(L_{W_p}^s)^2 \|\wt\Wb^{s, t-1}-\Wb_p^{s, t-1}\|^2  \notag \\
&~~+3(\gamma_2^2 + 2)(L_W^s)^2\sum_{t = Q_1 + 1}^{Q^s}\|\wt \Wb^{s, t-1} - \wt \Wb^{s, t}\|_F^2 + \frac{3C_1^s(\zeta^2 + \phi^2)}{2}\label{lemS2: prox_bound2} 
\end{align}
\begin{align}
\leq & 3(\gamma_2^2 + 2)(L_W^s)^2\sum_{t = Q_1 + 1}^{Q^s}\|\wt \Wb^{s, t-1} - \wt \Wb^{s, t}\|_F^2 + \frac{C_3^s(\frac{11\zeta^2}{3} + \phi^2)}{\gamma_2^2}+ \frac{3C_1^s(\zeta^2 + \phi^2)}{2}, \label{lemS2: prox_bound3}
\end{align}
where
\begin{align}
C_3^s \triangleq 6(3Q_2^s(Q_2^s - 1)/2 + 2)C_1^s,
\end{align}
the first term in the RHS of \eqref{lemS2: prox_bound1} follows because of \eqref{prox: summation}; the second term in the RHS of \eqref{lemS2: prox_bound1} follows due to \eqref{eqn: square_sum}, and
\eqref{lemS2: prox_bound2} follows because  $\wt \Wb^{s, Q_1} = \wt \Wb^{s, Q_1 - 1}$ and $\wt \Wb^{s, Q_1} = \Wb_p^{s, Q_1}$.
Then, summing up \eqref{lemS2: prox_bound3} from $s = 1$ to $S$ yields
\begin{align}
&\sum_{s = 1}^{S}\sum_{t = Q_1}^{Q^s - 1}G_{W}(\wt \Wb^{s, t}, \Hb^{s, t})\notag \\
\leq & 3(\gamma_2^2 +2) (\ol L_W)^2\sum_{s = 1}^{S}\sum_{t = Q_1 + 1}^{Q^s}\|\wt \Wb^{s, t-1} - \wt \Wb^{s, t}\|_F^2 \notag \\
&~~~+ \frac{(\frac{11\zeta^2}{3} + \phi^2)\sum_{s = 1}^{S}C_3^s}{\gamma_2^2} + \frac{3(\zeta^2 + \phi^2)\sum_{s = 1}^{S}C_1^s}{2}.
\end{align}
\hfill $\blacksquare$

\section{Lemma 7 and its proof}

\begin{Lemma} \label{lem: bound_cs}
	For any $s$, if $Q_2^s = \lfloor \frac{\hat{Q}}{s} \rfloor + 1$, the following equalities hold.
	{\small \begin{align}
		\sum_{s = 1}^{S} C_1^s = \mathcal{O}(\hat{Q}^3), ~\sum_{s = 1}^{S} C_2^s = \mathcal{O}( \hat{Q}^5), ~ \sum_{s = 1}^{S} C_3^s = \mathcal{O}( \hat{Q}^5).
		\end{align}}
\end{Lemma}

\noindent {\bf Poof:} Firstly, we have
\begin{align}
\sum_{s = 1}^{\hat{Q}} \frac{\hat{Q}}{s} &=\hat{Q}\bigg( 1 + \sum_{s = 2}^{\hat{Q}} \frac{1}{s} \bigg) \notag \\
&\leq\hat{Q}\bigg( 1 + \int_{1}^{\hat{Q}} \frac{1}{s} ds\bigg) \notag \\
&= \hat{Q}+ \hat{Q}\ln\hat{Q},
\end{align}
and for any $n > 1$,
\begin{align}
\sum_{s = 1}^{\hat{Q}} \bigg(\frac{\hat{Q}}{s}\bigg)^n &= \hat{Q}^n \bigg( 1 + \sum_{s = 2}^{\hat{Q}} \frac{1}{s^n} \bigg)\notag \\
&\leq \hat{Q}^n \bigg( 1 + \int_{1}^{\hat{Q}}  \frac{1}{s^n}ds \bigg)\notag \\
&=\hat{Q}^n \bigg( 1 - \bigg(\frac{1}{(n-1)\hat{Q}^{n-1}} - \frac{1}{n - 1}\bigg) \bigg)\notag \\
&= \hat{Q}^n \bigg(\frac{n}{n - 1} - \frac{1}{(n-1)\hat{Q}^{n-1}}\bigg) \notag \\
&= \frac{\hat{Q}^nn}{n - 1} - \frac{\hat{Q}}{(n-1)}.
\end{align}
Then, since $Q_2^s = \lfloor {\hat{Q}}/{s} \rfloor +1, \hat{Q} > 1$, we have 
\begin{align}
\sum_{s = 1}^{S} C_1^s =& \sum_{s = 1}^{S} Q_2^s(Q_2^s - 1)(2Q_2^s - 1) \notag \\
= & \sum_{s = 1}^{S} \bigg (\bigg\lfloor \frac{\hat{Q}}{s} \bigg\rfloor + 1\bigg)\bigg\lfloor \frac{\hat{Q}}{s} \bigg\rfloor\bigg(2\bigg\lfloor \frac{\hat{Q}}{s} \bigg\rfloor + 1\bigg) \notag \\
\leq  & \sum_{s = 1}^{\hat{Q}} \bigg(\frac{\hat{Q}}{s} + 1\bigg) \bigg(\frac{\hat{Q}}{s}\bigg)\bigg(2 \frac{\hat{Q}}{s}  + 1\bigg) \notag \\
= &2 \sum_{s = 1}^{\hat{Q}} \bigg(\frac{\hat{Q}}{s}\bigg)^3 + 3 \sum_{s = 1}^{\hat{Q}} \bigg(\frac{\hat{Q}}{s}\bigg)^2 + \sum_{s = 1}^{\hat{Q}} \frac{\hat{Q}}{s} \notag  \\
\leq  &2 \bigg(\frac{3\hat{Q}^3}{2} - \frac{\hat{Q}}{2}\bigg) + 3 \bigg(2\hat{Q}^2 - \hat{Q}\bigg) + \hat{Q} + \hat{Q} \ln \hat{Q} \\
= & 3\hat{Q}^3 + 6\hat{Q}^2 - 3\hat{Q} + \hat{Q}\ln \hat{Q} \\
= & \mathcal{O}(\hat{Q}^3).
\end{align}
For $C_2^s$, we have
\begin{align}
\sum_{s = 1}^{S} C_2^s =& \sum_{s = 1}^{S} 6(3Q_2^s(Q_2^s - 1)/2 + 4 + 32/m)C_1^s \notag \\
\leq & 9 \sum_{s = 1}^{S}(Q_2^s)^2(Q_2^s - 1)^2(2Q_2^s - 1) + 216\sum_{s = 1}^{S} C_1^s \\
= & 9 \sum_{s = 1}^{S}\bigg(\bigg\lfloor \frac{\hat{Q}}{s} \bigg\rfloor + 1\bigg)^2\bigg(\bigg\lfloor \frac{\hat{Q}}{s} \bigg\rfloor\bigg)^2\bigg(2\bigg\lfloor \frac{\hat{Q}}{s} \bigg\rfloor + 1\bigg) + 216 \sum_{s = 1}^{S} C_1^s 
\end{align}
\begin{align}
\leq & 9 \sum_{s = 1}^{\hat{Q}}\bigg(\frac{\hat{Q}}{s} + 1\bigg)^2\bigg( \frac{\hat{Q}}{s} \bigg)^2\bigg(2 \frac{\hat{Q}}{s}  + 1\bigg) + 216 \sum_{s = 1}^{S} C_1^s \label{c2_bound}\\
= &\mathcal{O}( \hat{Q}^5).
\end{align}
Similarly, we have for $C_3^s$ that 
\begin{align}
\sum_{s = 1}^{S} C_3^s \leq &\sum_{s = 1}^{S} 6(3Q_2^s(Q_2^s - 1)/2 + 2)C_1^s \notag \\
\leq & 9 \sum_{s = 1}^{S}(Q_2^s)^2(Q_2^s - 1)^2(2Q_2^s - 1) + 12\sum_{s = 1}^{S} C_1^s \\
\leq& 18 \bigg(\frac{5\hat{Q}^5}{4} - \frac{\hat{Q}}{4}\bigg) + 45 \bigg(\frac{4\hat{Q}^4}{3} - \frac{\hat{Q}}{3}\bigg) + 36 \bigg(\frac{3\hat{Q}^3}{2} - \frac{\hat{Q}}{2}\bigg) \notag \\
&~~~~+ 9 \bigg(2\hat{Q}^2 - \hat{Q}\bigg)+ \mathcal{O}(\hat{Q}^3) \label{c3_bound}\\
= & \mathcal{O}( \hat{Q}^5),
\end{align}
where \eqref{c3_bound} follows by using the same procedure as in  \eqref{c2_bound}.

\hfill $\blacksquare$

\section{Proof of Theorem 2}
	Firstly, we define $\Qc_1 \triangleq \{1, \ldots, Q_1\}, \Qc_2 \triangleq \{Q_1 + 1, \ldots, Q\}$, and $\Pc = \{1, \ldots, P\}$. 
	From Algorithm 2, firstly note that 
		\begin{align}
		&\Wb^{s, Q} = \Wb^{s+1, 0}=\Wb^{s+1}, ~ \Wb^{s, t} = \Wb^{s, t-1},~t \in \Qc_1,  \notag\\
		&\Hb^{s, Q} = \Hb^{s+1, 0},~ \Hb^{s, t} = \Hb^{s, t-1},~t \in \Qc_2.\label{smeqn: H iterates}
		\end{align}
	Secondly, under partial client participation, the local updates of $\Hb_p^{s,t}$ are
\begin{align}
		\Hb_p^{s, t}\! =\left\{\!
		\begin{array}{ll}
		\Pc_{\Hc_p} \big\{\Hb_p^{s,t-1}\! -\! \frac{\nabla_{H_p}F_p(\Wb^{s,0}, \Hb_p^{s,t-1})} {c_p^{s}}\big\},~&{\rm if}~p\in \Ac^s, \\
		\Hb_p^{s,t-1},~&{\rm otherwise},
		\end{array}\right. \notag
	\end{align}
		for $t \in \Qc_1$, where only clients in $\Ac^s$ perform PGD. 
We also define $\Ec^{s-1}$ by the same way as that in the proof of Theorem 1, and denote $\mathbb{I}_{\Ac^s}^p$ as the indicator function which is one if the event $p\in \Ac^s$ is true and zero otherwise.}

\noindent \underline{\bf Objective Descent w.r.t. $\Hb$:}
Let us consider the descent of the objective function with respect to the update of $\Hb$ when $\Ec^{s-1}$ is given. Specifically, we have the following chain 

		\begin{align}
		&\E[F(\Wb^{s, 0}, \Hb^{s, Q_1})|\Ec^{s-1}] - \E[F(\Wb^{s, 0}, \Hb^{s, 0})|\Ec^{s-1}] \notag \\
		=& \E\bigg[\sum_{p \in \Ac^s} \omega_p \bigg(  F_p(\Wb^{s, 0}, \wt\Hb_p^{s, Q_1}) -F_p(\Wb^{s, 0}, \Hb_p^{s, 0})\bigg)|\Ec^{s-1}\bigg] \label{thm2: expect_descent_H1 2} \\
		=& \E\bigg[\sum_{p = 1}^{P} \mathbb{I}_{\Ac^s}^p\omega_p \bigg( F_p(\Wb^{s, 0}, \wt\Hb_p^{s, Q_1}) - F_p(\Wb^{s, 0}, \Hb_p^{s, 0}) \bigg)|\Ec^{s-1}\bigg] \notag 
		\end{align}
		\begin{align}
		=& \sum_{p = 1}^{P} \E[\mathbb{I}_{\Ac^s}^p|\Ec^{s-1}]~\omega_p \bigg( F_p(\Wb^{s, 0}, \wt\Hb_p^{s, Q_1}) - F_p(\Wb^{s, 0}, \Hb_p^{s, 0}) \bigg) \label{thm2: expect_descent_H1 3} \\
		=&\frac{m}{P}\sum_{p = 1}^{P} \omega_p \bigg( F_p(\Wb^{s, 0}, \wt\Hb_p^{s, Q_1}) - F_p(\Wb^{s, 0}, \Hb_p^{s, 0})\bigg) \label{thm2: expect_descent_H1 4} \\
		=&\frac{m}{P}\sum_{p = 1}^{P} \omega_p \sum_{t = 1}^{Q_1}\bigg( F_p(\Wb^{s, 0}, \wt\Hb_p^{s, t}) - F_p(\Wb^{s, 0}, \wt\Hb_p^{s, t-1})\bigg) \label{thm2: expect_descent_H1 4_1} \\
		\leq&-\frac{m(\gamma - 1)}{2P} \bigg(\sum_{t = 1}^{Q_1}  \sum_{p = 1}^{P} \omega_p L_{H_p}^s \|\wt\Hb_p^{s, t-1} - \wt\Hb_p^{s, t}\|_F^2\bigg), 
		\label{thm2: expect_descent_H1} 
		\end{align}
	where \eqref{thm2: expect_descent_H1 2} is due to partial client participation (PCP); \eqref{thm2: expect_descent_H1 3} holds because $F_p(\Wb^{s, 0}, \Hb_p^{s, 0}) - F_p(\Wb^{s, 0}, \wt\Hb_p^{s, Q_1})$ is deterministic given $\Ec^{s-1}$; 
	\eqref{thm2: expect_descent_H1 4} is true since $\E[\mathbb{I}_{\Ac^s}^p|\Ec^{s-1}]=m/P$ when uniform sampling without replacement is employed;  
	\eqref{thm2: expect_descent_H1} follows (40) according to \cite[Lemma 3.2]{PALM_2014} and $c_p^s = \frac{\gamma}{2}L_{H_p}^s$.
	
\noindent \underline{\bf Objective Descent w.r.t. $\Wb$:}
By applying \cite[Lemma 3.2]{PALM_2014} to the update of $\Wb$ in (32) with $d^s=\frac{\gamma}{2} L_W^s$ and noting from \eqref{smeqn: H iterates} that 
$\Wb^{s, Q_1}=\Wb^{s, 0}$, $\Wb^{s, Q}=\Wb^{s+1, 0}$ and $\Hb^{s, Q_1}=\Hb^{s+1, 0}$, we immediately obtain
 \begin{align}
	&\E[F(\Wb^{s+1, 0}, \Hb^{s+1, 0})|\Ec^{s-1}] - \E[F(\Wb^{s, 0}, \Hb^{s, Q_1})|\Ec^{s-1}] \notag \\
	=&\sum_{t = Q_1 + 1}^{Q} \E[F(\Wb^{s, t}, \Hb^{s, Q_1}) - F(\Wb^{s, t-1}, \Hb^{s, Q_1})|\Ec^{s-1}] \notag \\
	\leq& -\frac{\gamma - 1}{2} \sum_{t = Q_1 + 1}^{Q} L_{W}^s \E[ \|\Wb^{s, t} - \Wb^{s, t-1}\|_F^2|\Ec^{s-1}]. \label{thm2: descent_W}
	\end{align}
\noindent \underline{\bf Derivation of the Main Result:} After combing \eqref{thm2: expect_descent_H1} and \eqref{thm2: descent_W}, we have
	\vspace{-0.18cm}
 \begin{align}
		&\frac{m(\gamma - 1)}{2P}  \sum_{t = 1}^{Q_1} \sum_{p = 1}^{P}\omega_pL_{H_p}^s \E[\|\wt \Hb_p^{s, t} - \wt \Hb_p^{s, t-1}\|_F^2|\Ec^{s-1}] \notag\\
		&~~~~+\frac{\gamma - 1}{2} \sum_{t = Q_1 + 1}^{Q} L_{W}^s \E[\|\Wb^{s, t} - \Wb^{s, t-1}\|_F^2|\Ec^{s-1}] \notag \\
		\leq& \E[F(\Wb^{s, 0}, \Hb^{s, 0})]- \E[F(\Wb^{s+1, 0}, \Hb^{s + 1, 0})]. \label{thm2: expect_descent}
		\end{align}
	Since $c_p^s = \frac{\gamma L_{H_p}^s}{2}$, and $L_{H_1}^s = \ldots = L_{H_P}^s$, we have from \eqref{thm2: expect_descent} that
		\begin{align}
		&\sum_{t = 1}^{Q_1} \sum_{p = 1}^{P}\omega_p (c_p^s)^2 \E[\|\wt \Hb_p^{s, t} - \wt \Hb_p^{s, t-1}\|_F^2|\Ec^{s-1}] \notag \\
		\leq & \frac{P\gamma^2 L_{H_p}^s}{2m(\gamma - 1)} \bigg(\E[F(\Wb^{s, 0}, \Hb^{s, 0})- F(\Wb^{s+1, 0}, \Hb^{s + 1, 0})|\Ec^{s-1}]\bigg).\label{thm2: expect_descent_H}
		\end{align}
	By taking expectation over two sides of \eqref{thm2: expect_descent_H} and summing it up from $s = 1$ to $S$, we have
    \begin{align} 
		&\sum_{s = 1}^{S}\sum_{t = 1}^{Q_1} \sum_{p = 1}^{P}\omega_p \E[(c_p^s)^2 \|\wt \Hb_p^{s, t} - \wt \Hb_p^{s, t-1}\|_F^2] \notag \\
		\leq & \frac{P\gamma^2 \ol L_{H}}{2m(\gamma - 1)} \bigg(F(\Wb^{1, 0}, \Hb^{1, 0})- \E[F(\Wb^{S+1, 0}, \Hb^{S + 1, 0})]\bigg) \notag \\
		\leq & \frac{P\gamma^2 \ol L_{H}}{2m(\gamma - 1)} \bigg(F(\Wb^{1, 0}, \Hb^{1, 0})- \underline{F}\bigg), \label{thm2: expect_descent_H2}
		\end{align}
	where \eqref{thm2: expect_descent_H2} follows because of Assumption 1 and 2. Then, we have
    \begin{align} 
		&\sum_{s = 1}^{S} \sum_{t = 1}^{Q_1}\E[G_{H}(\Wb^{s, t - 1}, \wt \Hb^{s, t-1})] \notag \\
		\leq & \frac{P\gamma^2 \ol L_{H}}{2m(\gamma - 1)} \bigg(F(\Wb^{1, 0}, \Hb^{1, 0})- \underline{F}\bigg), \label{thm2: expect_descent_Hprox}
		\end{align} 
	Similarly, by using the same procedure as to obtain \eqref{thm2: expect_descent_Hprox}, we have 
	\begin{align}
		&\sum_{s = 1}^{S}\sum_{t = Q_1 + 1}^{Q} \E[G_{W}(\Wb^{s, t-1}, \Hb^{s, t-1})] \notag \\
		\leq & \frac{\gamma^2 \ol L_W}{2(\gamma - 1)}\bigg(F(\Wb^{1, 0}, \Hb^{1, 0})- \underline{F}\bigg)\label{thm2: expect_descent_Wprox}
		\end{align}
	Combing \eqref{thm2: expect_descent_Hprox} and \eqref{thm2: expect_descent_Wprox} and then dividing two sides by $T = SQ$ yields
	\begin{align}
		&\frac{1}{T}\bigg[\sum_{s = 1}^{S}  \sum_{t = 1}^{Q_1}\E[G_{H}(\Wb^{s, t - 1}, \wt \Hb^{s, t-1})]\notag \\
		&~~~~~~~~~+\sum_{s = 1}^{S}\sum_{t = Q_1 + 1}^{Q} \E[G_{W}(\Wb^{s, t-1}, \Hb^{s, t-1})]\bigg] \notag \\
		\leq & \frac{1}{T}\bigg(\frac{P\gamma^2 \ol L_{H}}{2m(\gamma - 1)} + \frac{\gamma^2 \ol L_W}{2(\gamma - 1)}\bigg) \bigg(F(\Wb^{1, 0}, \Hb^{1, 0})- \underline{F}\bigg)
		\end{align}
		 \hfill $\blacksquare$

\newpage 
\begin{center}
	\LARGE
	\bf Supplementary Materials: Figures
	
\end{center}

\begin{figure} [H]
	\centering
	\subfigure[\scriptsize TCGA, \textbf{Case 2}, $m=100$]{
		\includegraphics[width=5.3cm]{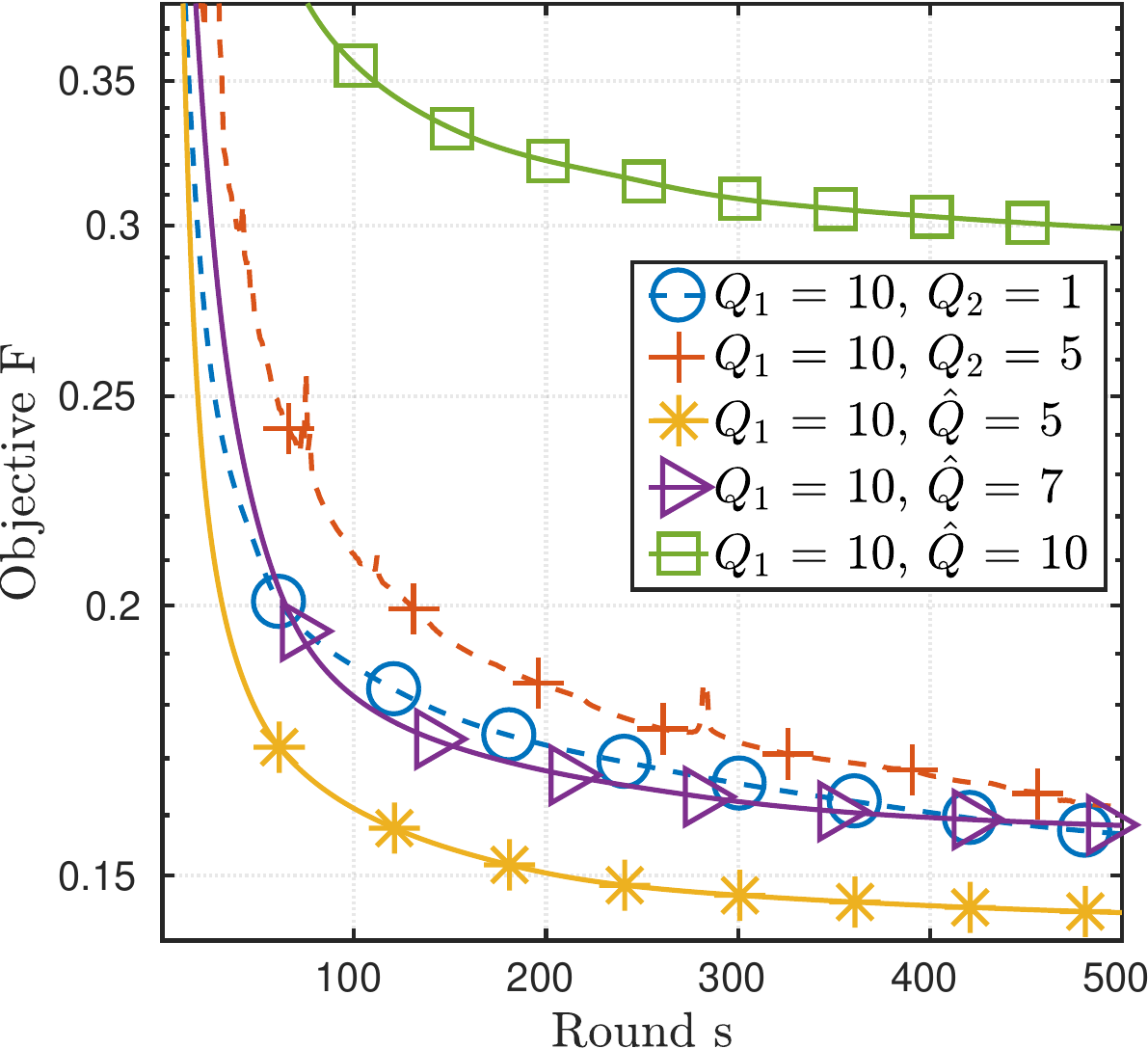} 
	}\vspace{-0.01cm}
	\subfigure[\scriptsize syn, \textbf{Case 1}]{
		\includegraphics[width=5.5cm]{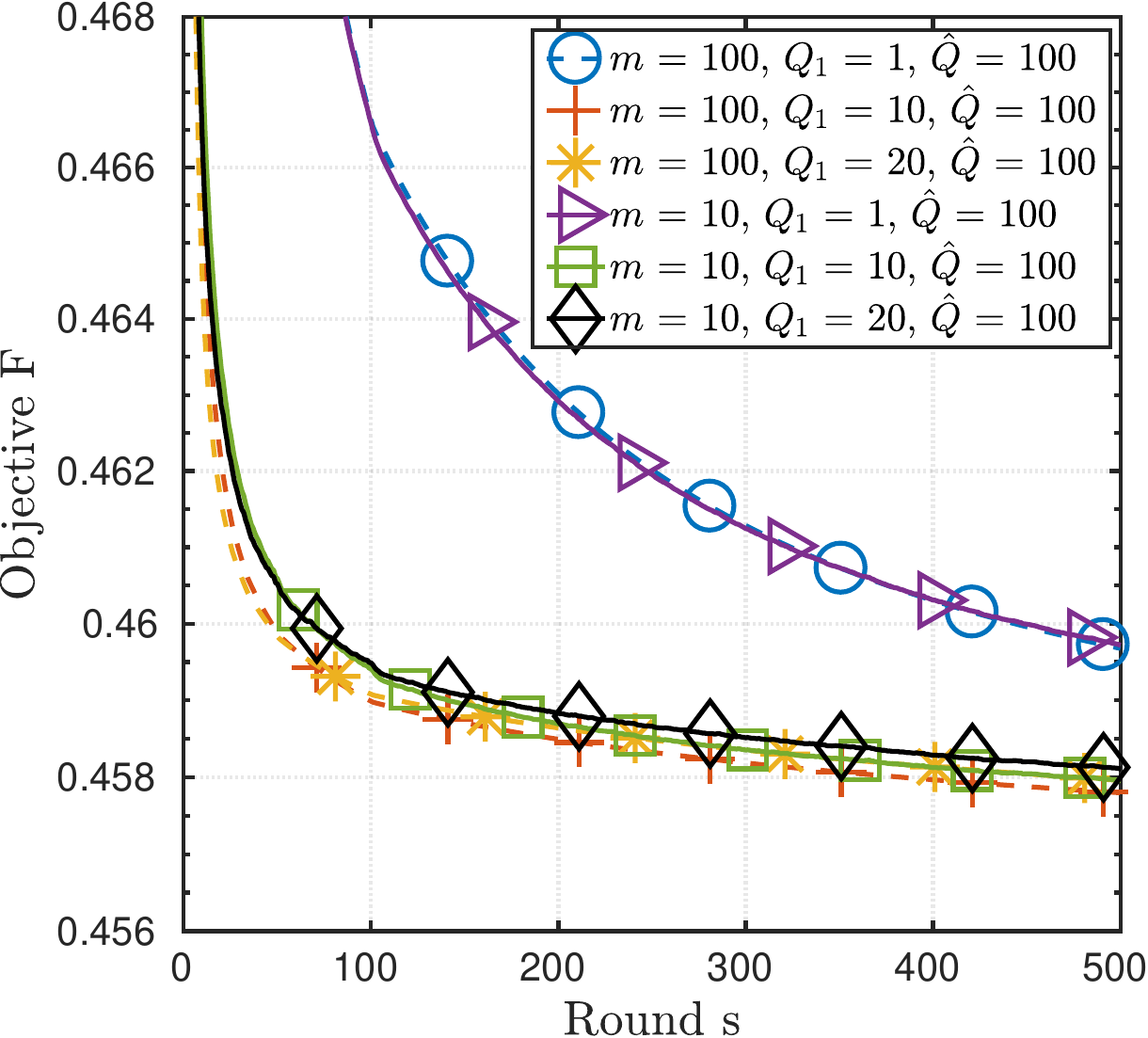} 
	}\vspace{-0.1cm}
	\caption{Convergence curve versus number of rounds of FedMAvg with different values of $Q_1$ and $\hat{Q}$.}
	\vspace{-0.0cm}
\end{figure}

\begin{figure} [H]
	\centering
	\subfigure[\scriptsize  FedMAvg, TDT2, $F$ v.s. round $s$]{
		\includegraphics[width=5.5cm]{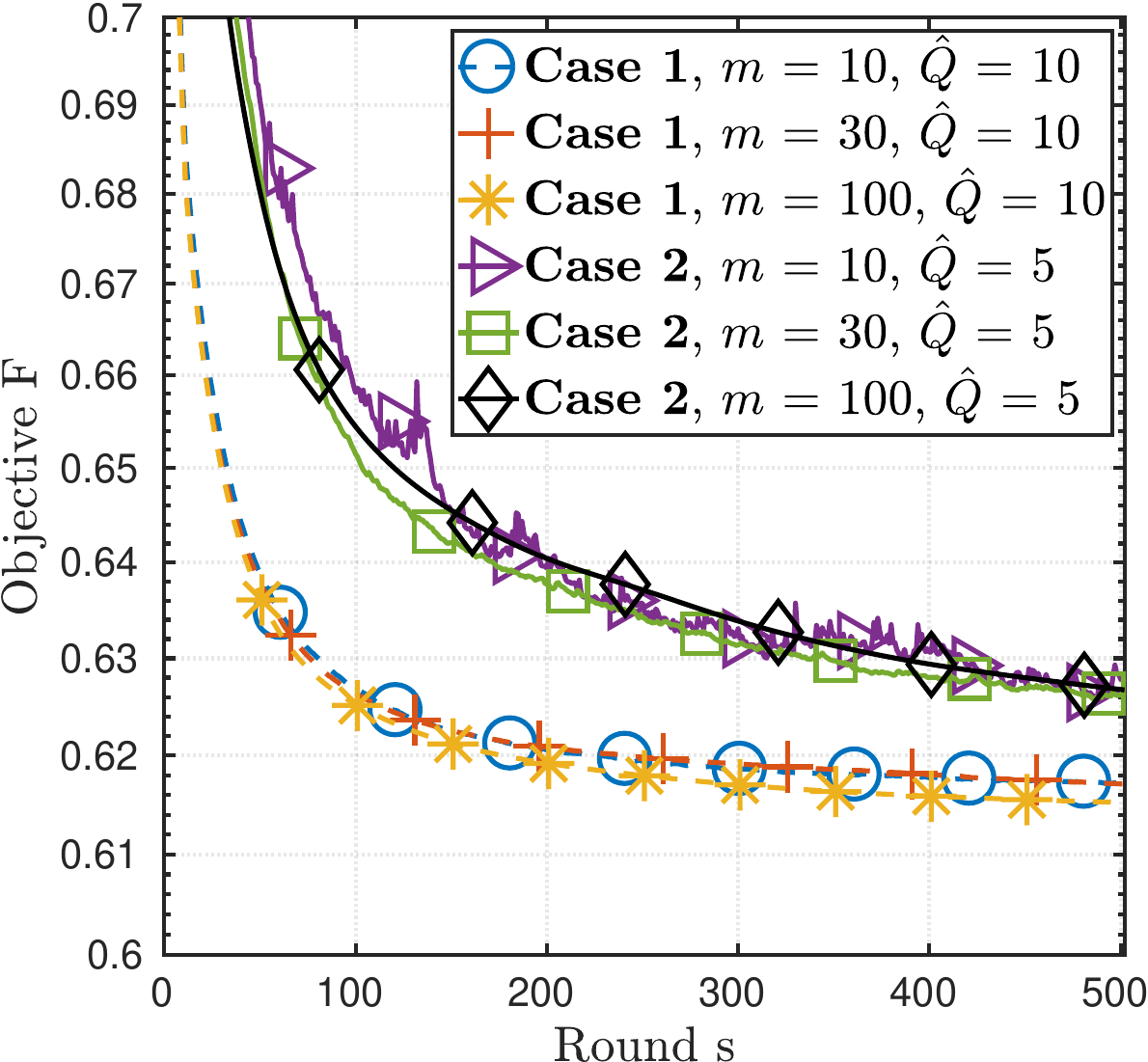} 
	} 
	\subfigure[\scriptsize  FedMGS, TDT2, $F$ v.s. round $s$]{
		\includegraphics[width=5.5cm]{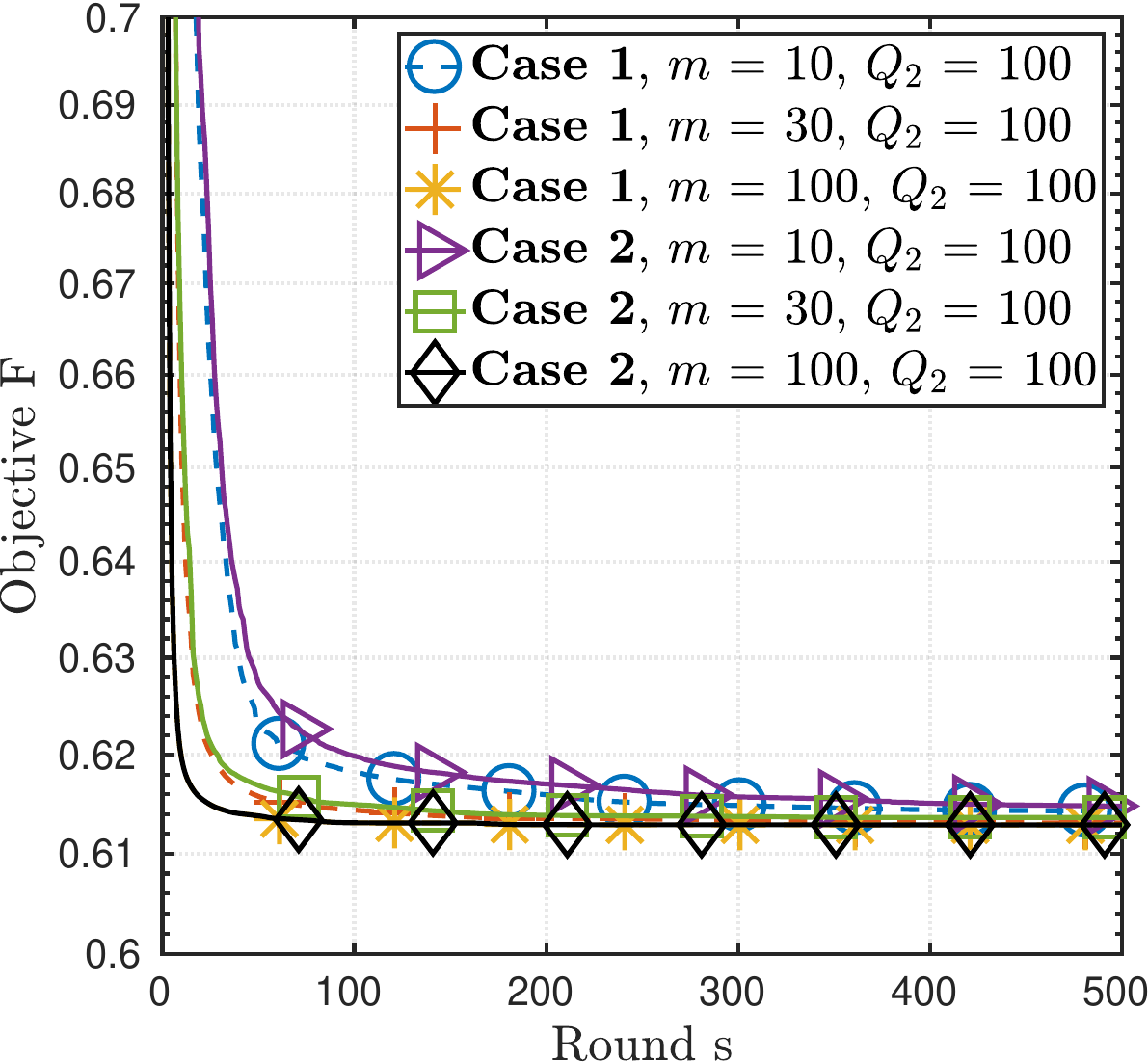} 
	}
	\subfigure[\scriptsize FedMAvg, MNIST, $F$ v.s. round $s$]{
		\includegraphics[width=5.5cm]{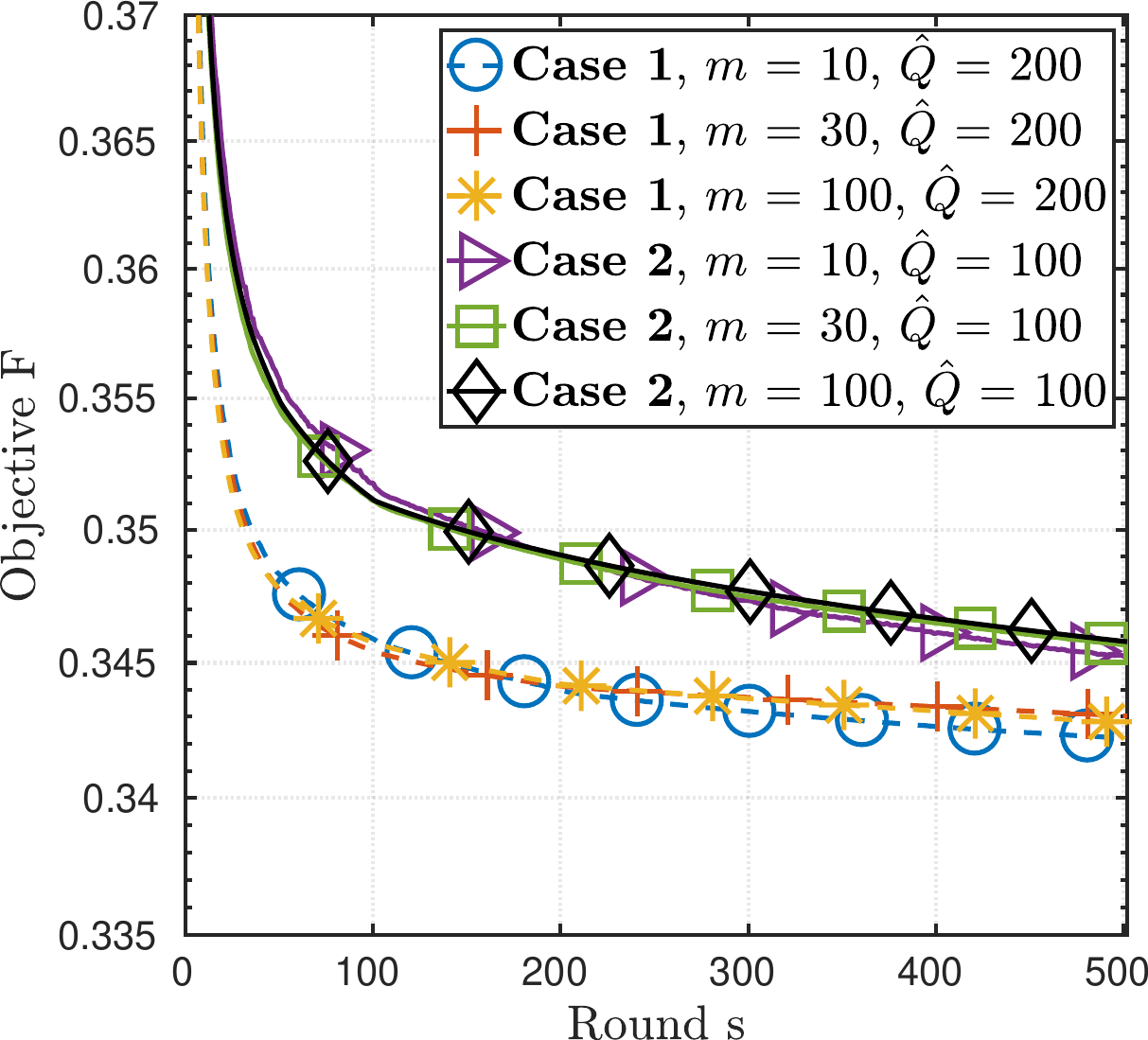} 
	} 
	\subfigure[\scriptsize FedMGS, MNIST, $F$ v.s. round $s$]{
		\includegraphics[width=5.5cm]{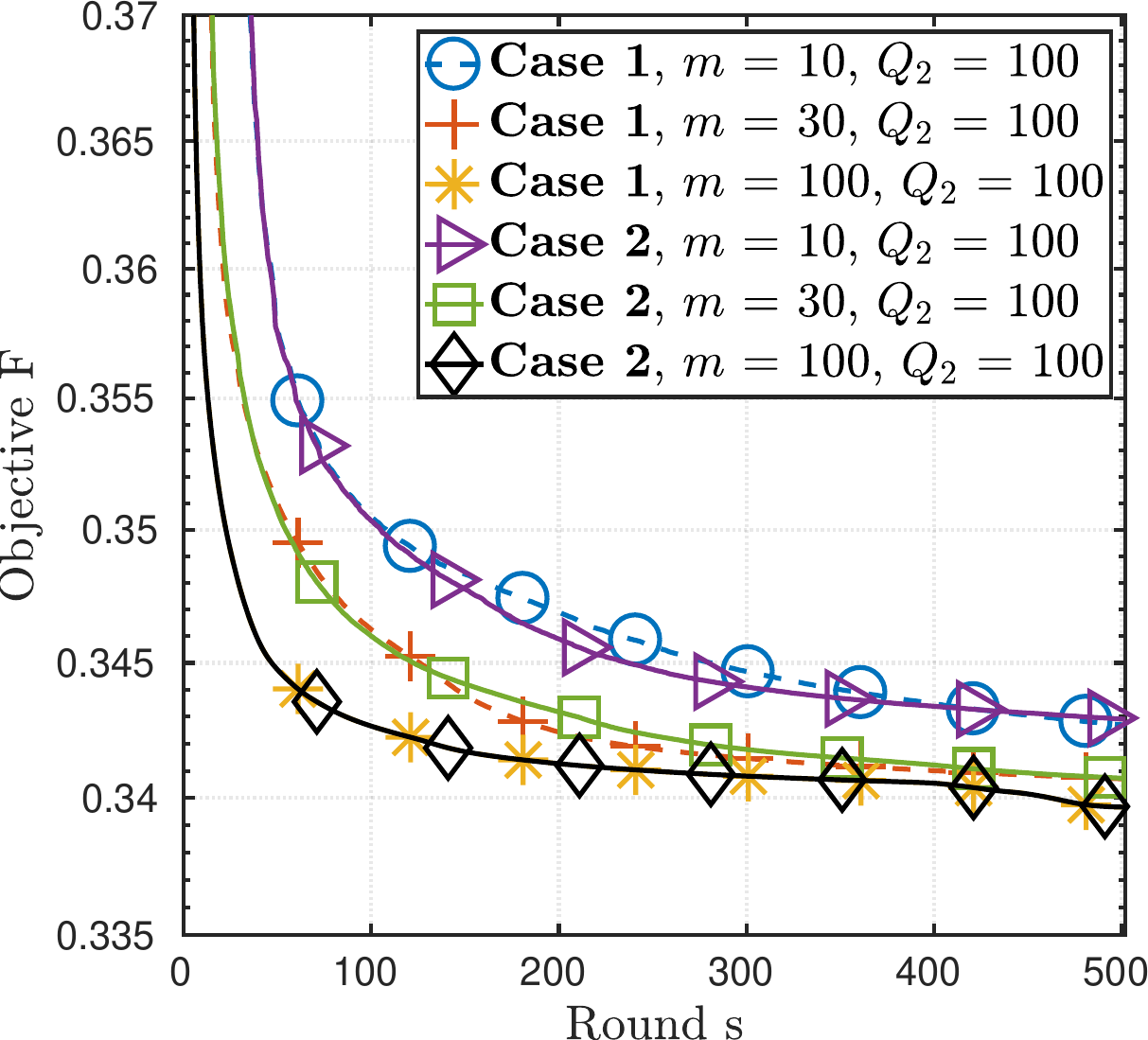} 
	} 
	\caption{Convergence curve versus number of rounds of FedMAvg and FedMGS on the TDT2 and MNIST dataset. It is set that $Q_1 = 10$ for both FedMAvg and FedMGS.}
		\label{subfig:comparison}
	\vspace{-0.0cm}
\end{figure}

\begin{figure} [H]
	\centering
	\subfigure[\scriptsize  TDT2, $F$ v.s. round $s$]{
		\includegraphics[width=5.5cm]{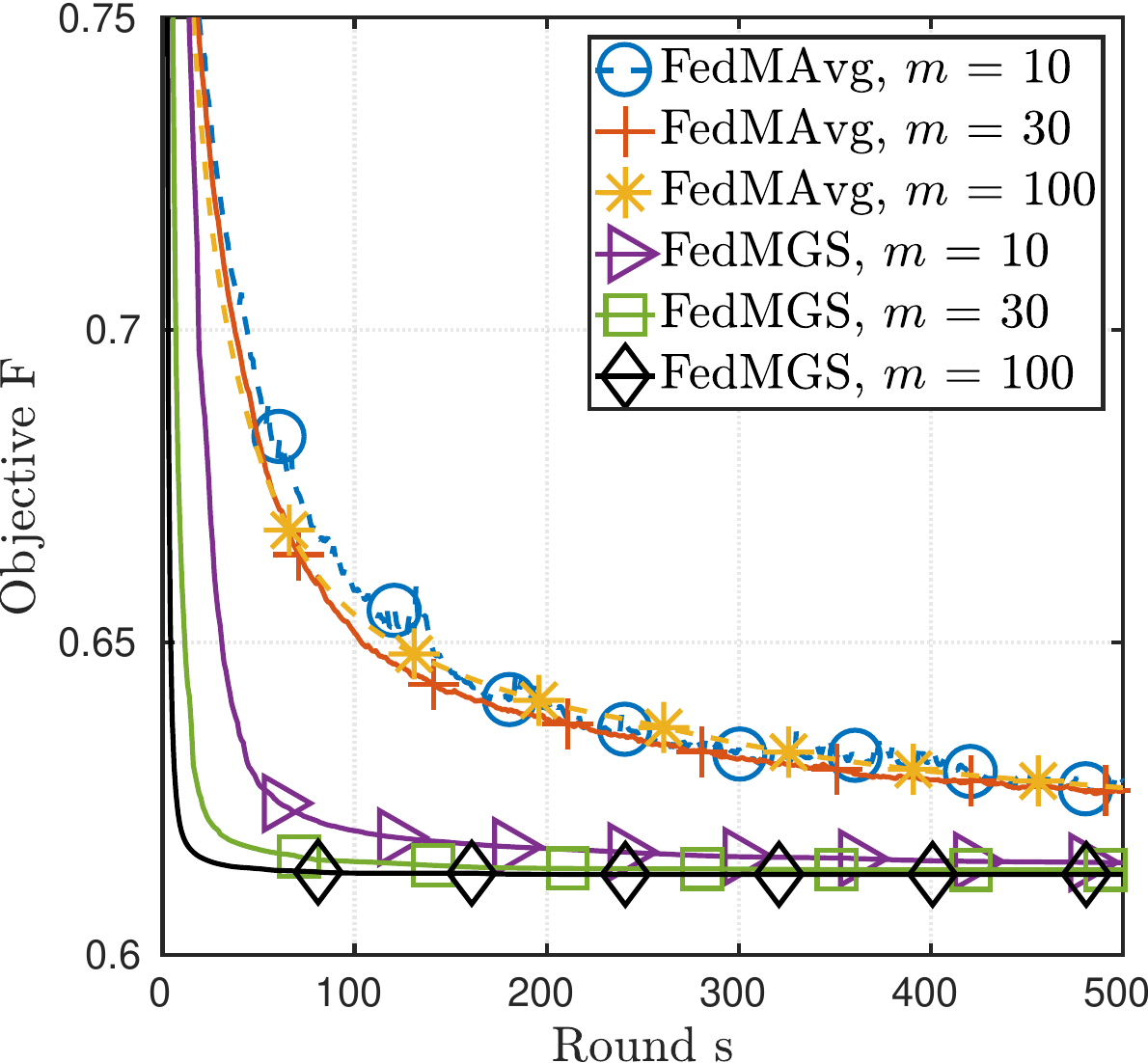} 
	} \vspace{-0.2cm}
	\subfigure[\scriptsize   TDT2, $F$ v.s. com. cost]{
		\includegraphics[width=5.5cm]{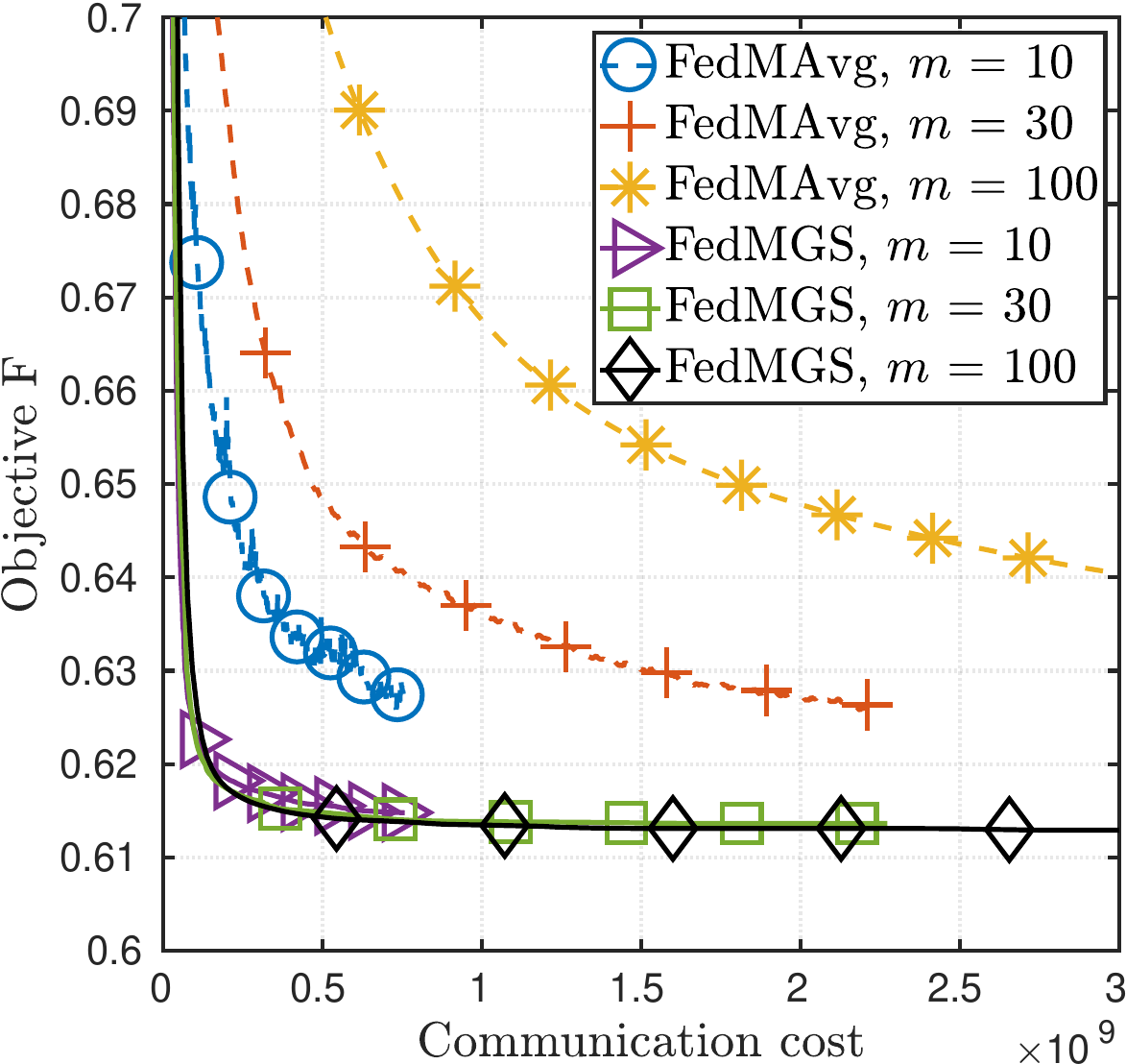} 
	}
	\subfigure[\scriptsize  MNIST, $F$ v.s. round $s$]{
		\includegraphics[width=5.5cm]{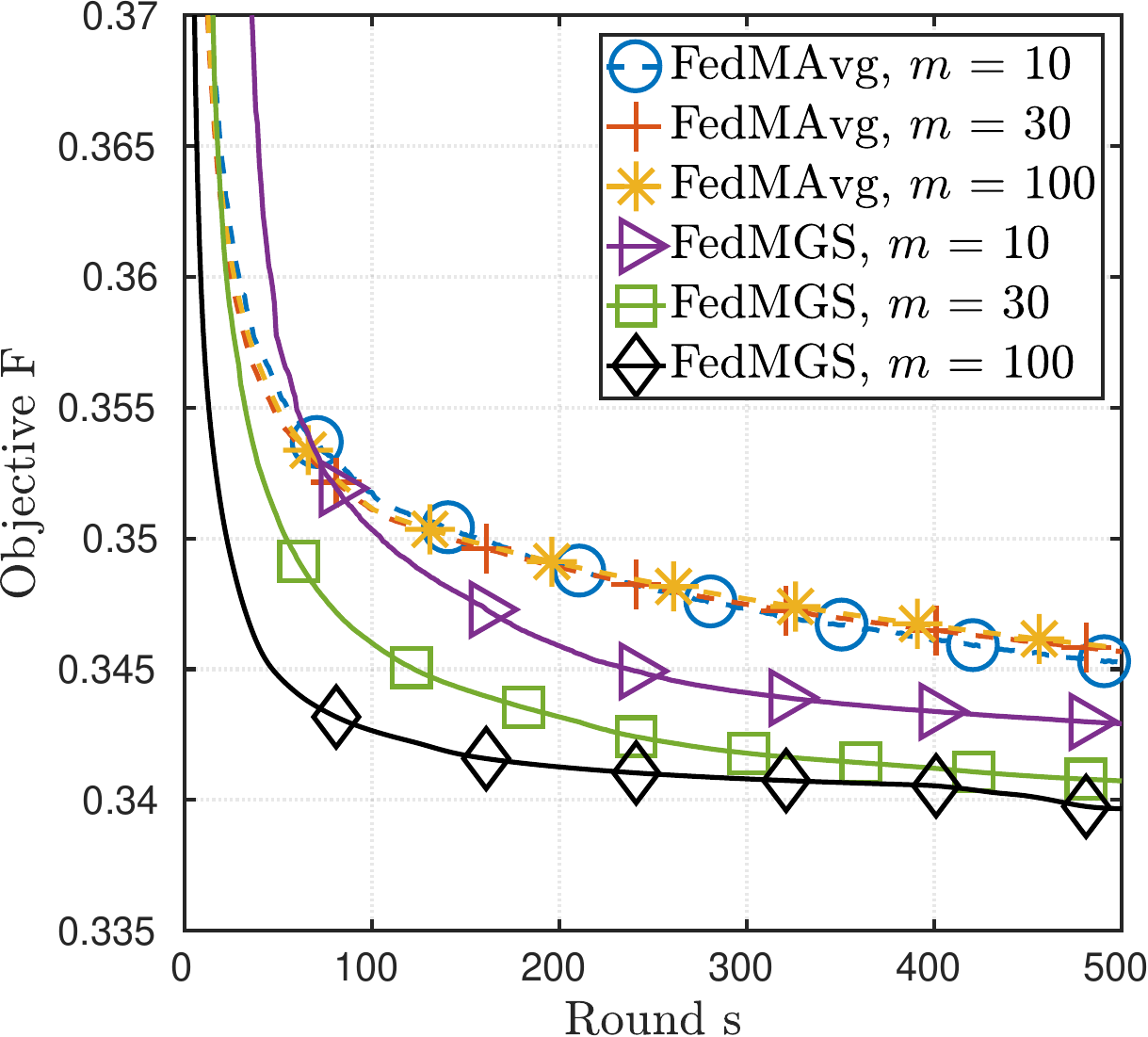} 
	} 
	\subfigure[\scriptsize  MNIST, $F$ v.s. com. cost]{
		\includegraphics[width=5.5cm]{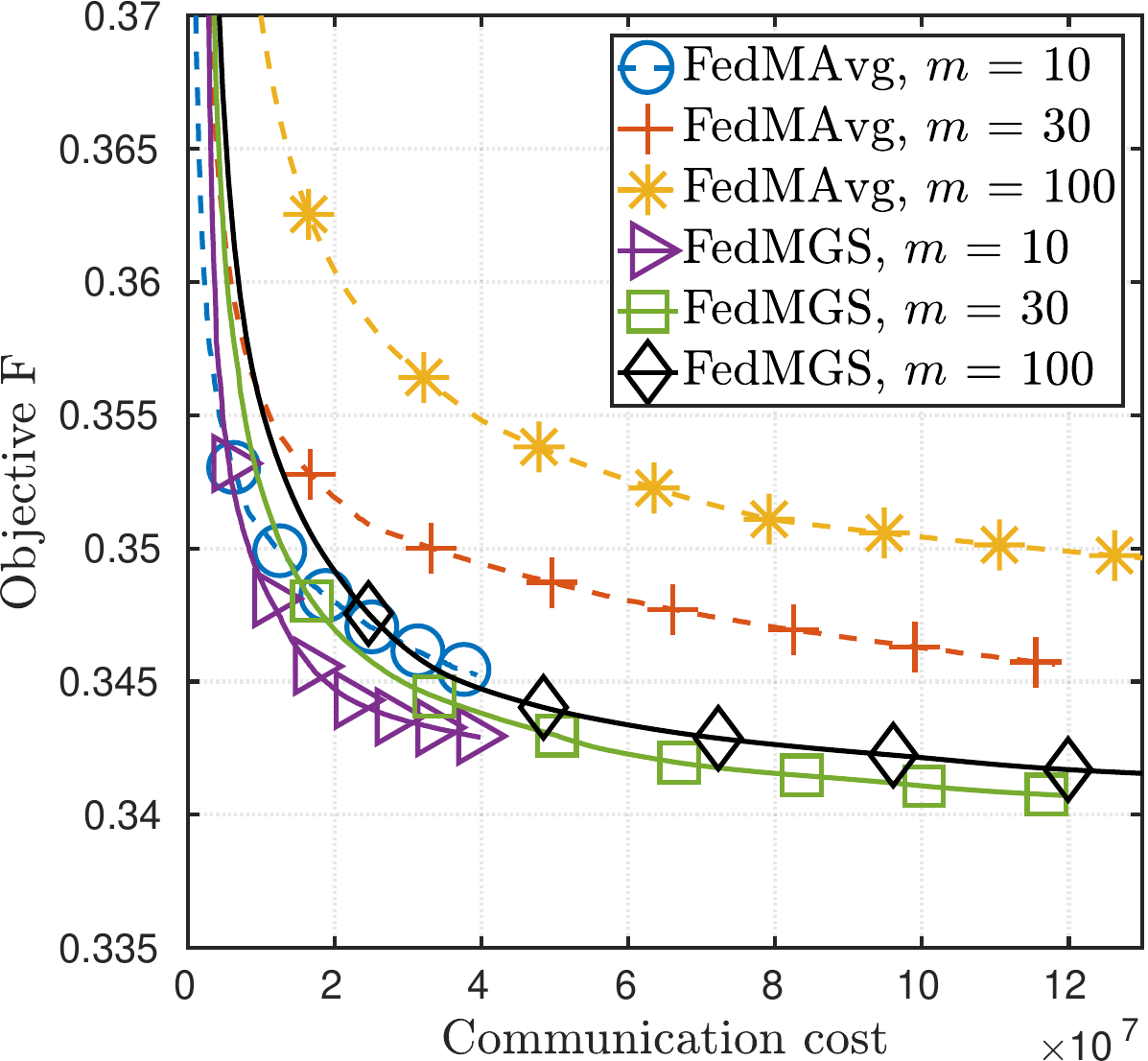} 
	} 
	\vspace{-0.3cm}
	\caption{Convergence curve versus number of rounds/communication cost of FedMAvg and FedMGS under non-i.i.d data.}
	\label{subfig: comparison}
	\vspace{-0.0cm}
\end{figure}

\begin{figure} [H]
	\centering
	\subfigure[\scriptsize TDT2]{
		\includegraphics[width=5.5cm]{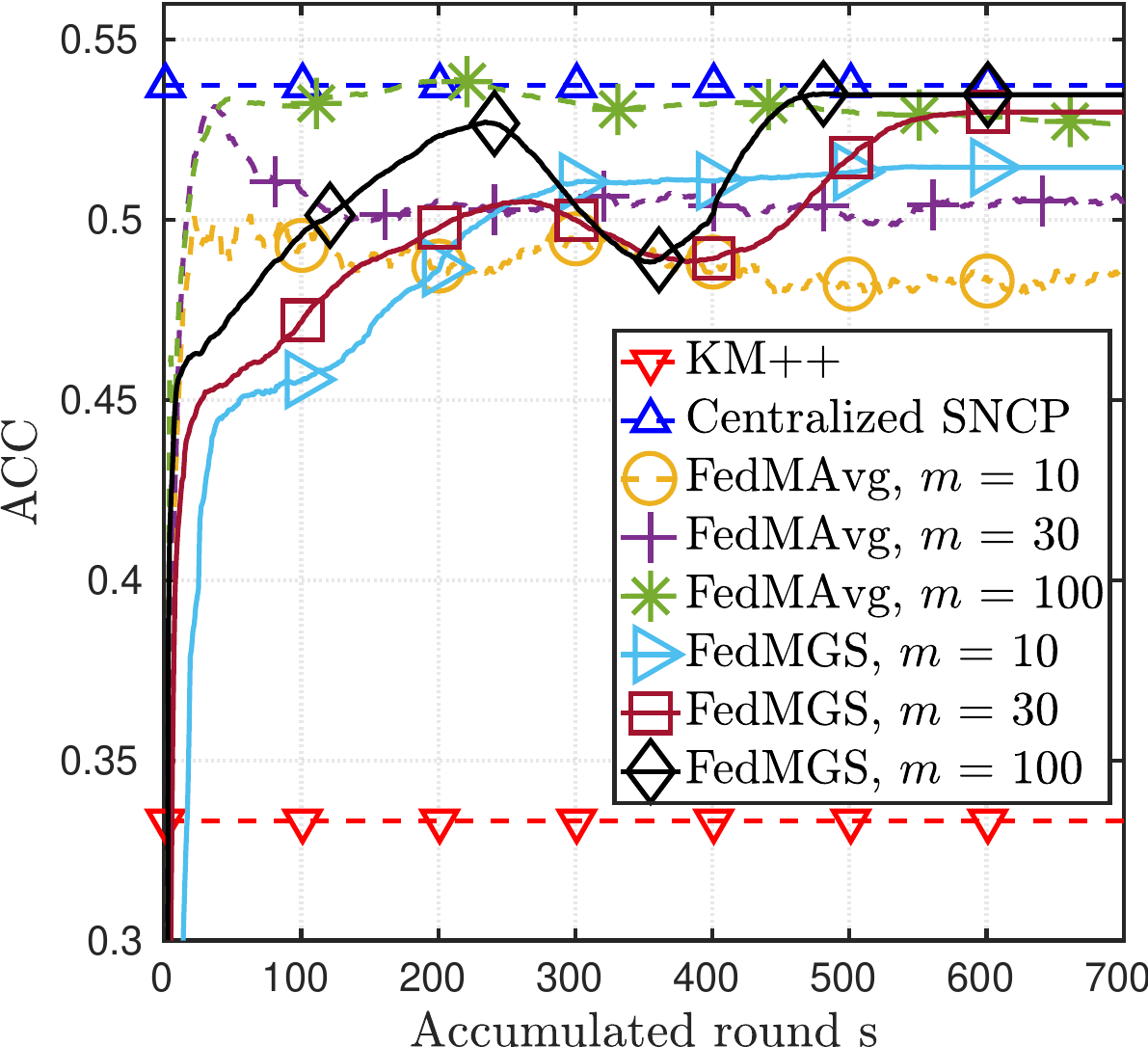}
	}
	\subfigure[\scriptsize TDT2]{
		\includegraphics[width=5.5cm]{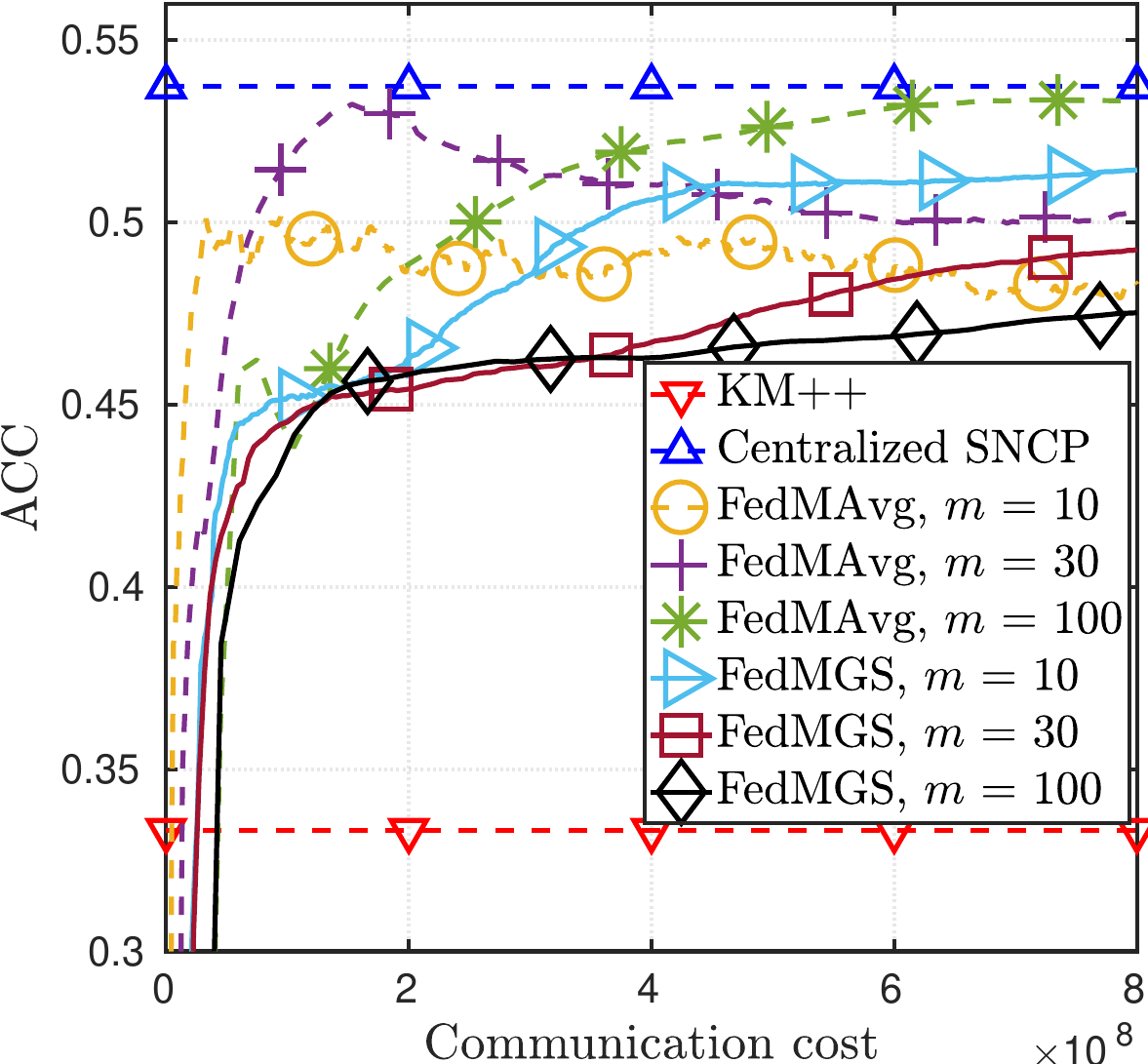}
	}
	\subfigure[\scriptsize MNIST]{
		\includegraphics[width=5.5cm]{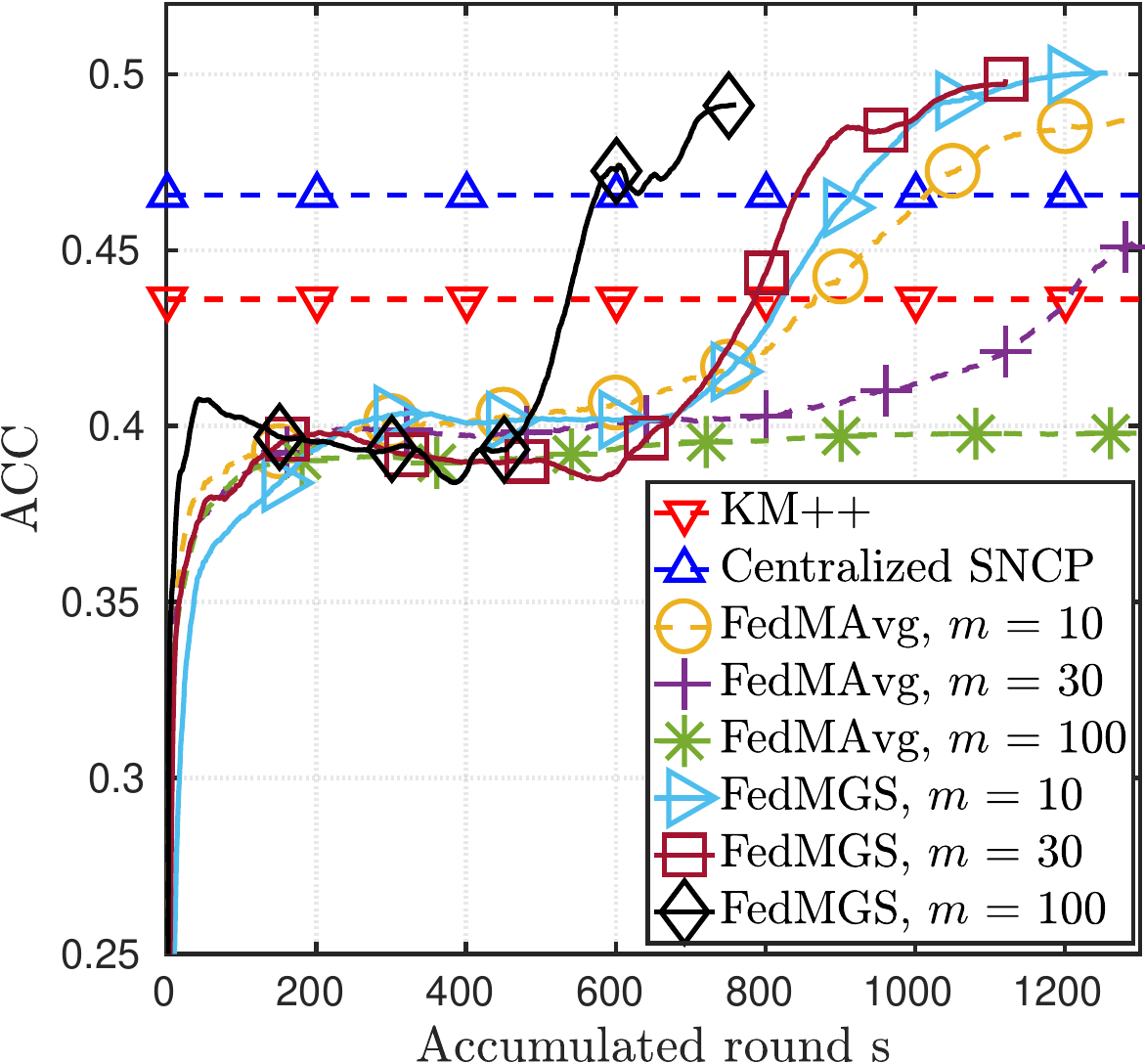}
	}
	\subfigure[\scriptsize MNIST]{
		\includegraphics[width=5.5cm]{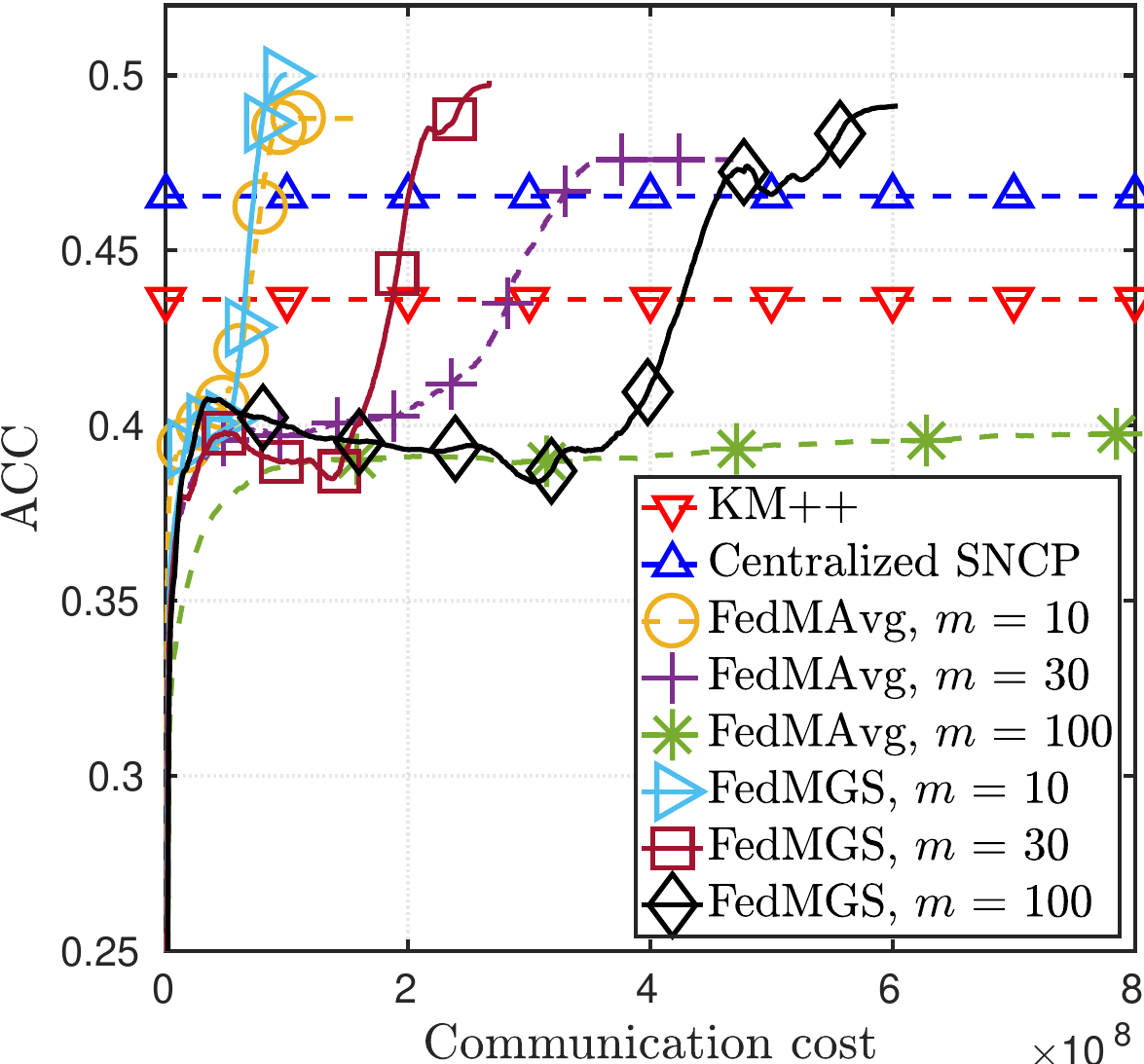}
	}
	\caption{Clustering accuracy versus number of accumulated rounds/communication cost of FedMAvg and FedMGS for the TDT2 and MNIST datasets.}
	\label{subfig:accuracy}
\end{figure}


\footnotesize

\bibliography{refs20,refs10}